\documentclass[11pt]{article}

\PassOptionsToPackage{table,dvipsnames}{xcolor} 
\usepackage[final]{acl}

\usepackage{times}
\usepackage{latexsym}
\usepackage[T1]{fontenc}
\usepackage[utf8]{inputenc}
\usepackage{microtype}
\usepackage{inconsolata}
\usepackage{graphicx}
\graphicspath{{figures/}}

\usepackage{amsmath,amssymb,mathtools}
\usepackage{bm}
\usepackage{booktabs}
\usepackage{multirow}
\usepackage{makecell}
\usepackage{array}
\usepackage{tabularx}
\usepackage{subcaption}
\usepackage{enumitem}
\usepackage{xspace}
\usepackage{pifont}
\usepackage{colortbl}   

\usepackage[most]{tcolorbox}
\usepackage[capitalize,nameinlink,noabbrev]{cleveref} 

\definecolor{cOurs}{HTML}{0072B2}     
\definecolor{cCmpA}{HTML}{E69F00}     
\definecolor{cCmpB}{HTML}{009E73}     
\definecolor{cInterv}{HTML}{D55E00}   
\definecolor{cJudge}{HTML}{CC79A7}    
\definecolor{cCmpC}{HTML}{56B4E9}     
\definecolor{cBase}{HTML}{8C8C8C}     
\definecolor{cInk}{HTML}{1A1A1A}      
\definecolor{cUp}{HTML}{0072B2}       
\definecolor{cDown}{HTML}{D55E00}     
\definecolor{cNeutral}{HTML}{8C8C8C}  
\definecolor{cRowHi}{HTML}{EAF2F8}    
\definecolor{cBoxBg}{HTML}{F2F2F2}    
\definecolor{cBoxFrame}{HTML}{A3A3A3} 
\definecolor{cMarkBg}{HTML}{FFF6E5}   

\newcommand{\vs}{\textit{vs.}\xspace}

\newcounter{hypothesis}
\newenvironment{hypothesis}[1][]{%
  \refstepcounter{hypothesis}\par\smallskip\noindent
  \textbf{H\thehypothesis}\if\relax\detokenize{#1}\relax\else\ (\textit{#1})\fi.\ \ignorespaces}%
  {\par\smallskip}
\crefformat{hypothesis}{#2H#1#3}
\Crefformat{hypothesis}{#2H#1#3}
\crefmultiformat{hypothesis}{#2H#1#3}{ and~#2H#1#3}{, #2H#1#3}{, and~#2H#1#3}
\Crefmultiformat{hypothesis}{#2H#1#3}{ and~#2H#1#3}{, #2H#1#3}{, and~#2H#1#3}
\crefrangeformat{hypothesis}{#3H#1#4--#5H#2#6}
\Crefrangeformat{hypothesis}{#3H#1#4--#5H#2#6}
\newcommand{\hyp}[1]{\textbf{H#1}\xspace}

\newtcolorbox{claimbox}{enhanced,colback=cBoxBg,colframe=cBoxFrame,boxrule=0.5pt,
  arc=1.5pt,left=5pt,right=5pt,top=3pt,bottom=3pt,fontupper=\itshape}

\newcommand{\circled}[1]{\ding{\numexpr181+#1\relax}}

\newcommand{\std}[1]{{\scriptsize\,$\pm$\,#1}}

\newcommand{\sig}[1]{\textsuperscript{#1}}
\newcommand{\hup}{\,$\uparrow$}
\newcommand{\hdown}{\,$\downarrow$}

\newcommand{\hirow}{\rowcolor{cRowHi}} 
\newcommand{\tabstyle}{\small\setlength{\tabcolsep}{4pt}\renewcommand{\arraystretch}{1.0}}

\newcommand{\resdef}[2]{\expandafter\def\csname res@#1\endcsname{#2}}
\newcommand{\res}[1]{\ifcsname res@#1\endcsname\csname res@#1\endcsname\else\textbf{\textcolor{red}{??\detokenize{#1}??}}\fi} 

\newif\ifdraft
\drafttrue

\newcommand{\tbd}{\ifdraft\textcolor{gray}{--.-}\else\errmessage{Placeholder table cell (tbd) left in final build}\fi}

\newcommand{\E}{\mathbb{E}}

\newcommand{\CR}[1]{\mathrm{CR}_{\mathrm{#1}}}          
\newcommand{\AIdx}[1]{\mathrm{AI}_{\mathrm{#1}}}        
\newcommand{\dF}[1]{\Delta F_{\mathrm{#1}}}             
\newcommand{\Dens}[1]{D_{\mathrm{#1}}}                  
\newcommand{\Split}{S}                                  
\newcommand{\MB}[1]{\mathrm{MB}_{\mathrm{#1}}}          
\newcommand{\COMP}[1]{\mathrm{COMP}_{\mathrm{#1}}}      
\newcommand{\stage}[1]{\textsc{#1}}                     
\newcommand{\vty}[1]{\textsc{\MakeLowercase{#1}}}       
\newcommand{\opp}{\mathrm{opp}}                          
\newcommand{\real}{\mathrm{real}}                        

\newcommand{\rsig}[1]{\ifcsname res@#1_sig\endcsname\csname res@#1_sig\endcsname\fi}
\newcommand{\resT}[1]{\ifcsname res@#1\endcsname\csname res@#1\endcsname\else\tbd\fi}
\newcommand{\rpm}[1]{\resT{#1}\rsig{#1}\ifcsname res@#1_std\endcsname\std{\resT{#1_std}}\fi}  

\newcommand{\rci}[1]{[\resT{#1_lo}, \resT{#1_hi}]}
\newcommand{\rhl}[1]{\ifcsname res@#1_hl\endcsname\csname res@#1_hl\endcsname{\resT{#1}}\else\resT{#1}\fi}
\newcommand{\rcell}[1]{\rhl{#1}\rsig{#1}\ifcsname res@#1_std\endcsname\std{\resT{#1_std}}\fi}

\usepackage{titletoc}
\usepackage{placeins} 
\resdef{design.n_lineages}{10}  
\resdef{design.n_open_lineages}{7}  
\resdef{design.n_samples_per_seed}{2}  
\resdef{design.n_sampling_seeds}{3}  
\resdef{design.n_seeds}{287}  
\resdef{design.n_varieties}{6}  
\resdef{design.temperature}{0.7}  
\resdef{design.top_p}{0.95}  
\resdef{e1.l1.base.aae.cr}{0.000}  
\resdef{e1.l1.base.aae.cr_std}{0.002}  
\resdef{e1.l1.base.bre.cr_rate}{0.166}  
\resdef{e1.l1.base.bre.cr_rate_std}{0.008}  
\resdef{e1.l1.base.inf.cr}{0.068}  
\resdef{e1.l1.base.inf.cr_std}{0.006}  
\resdef{e1.l1.base.ngp.cr_rate}{0.012}  
\resdef{e1.l1.base.ngp.cr_rate_std}{0.002}  
\resdef{e1.l1.base.rho}{0.778}  
\resdef{e1.l1.base.sge.cr_rate}{0.001}  
\resdef{e1.l1.base.sge.cr_rate_std}{0.001}  
\resdef{e1.l1.base.split_rate}{0.165}  
\resdef{e1.l1.dpo.aae.cr}{0.000}  
\resdef{e1.l1.dpo.aae.cr_std}{0.001}  
\resdef{e1.l1.dpo.bre.cr_rate}{0.059}  
\resdef{e1.l1.dpo.bre.cr_rate_std}{0.005}  
\resdef{e1.l1.dpo.inf.cr}{0.011}  
\resdef{e1.l1.dpo.inf.cr_std}{0.005}  
\resdef{e1.l1.dpo.ngp.cr_rate}{0.005}  
\resdef{e1.l1.dpo.ngp.cr_rate_std}{0.002}  
\resdef{e1.l1.dpo.rho}{0.845}  
\resdef{e1.l1.dpo.sge.cr_rate}{0.001}  
\resdef{e1.l1.dpo.sge.cr_rate_std}{0.001}  
\resdef{e1.l1.dpo.split_rate}{0.030}  
\resdef{e1.l1.instruct.aae.cr}{0.000}  
\resdef{e1.l1.instruct.aae.cr_std}{0.001}  
\resdef{e1.l1.instruct.bre.cr_rate}{0.068}  
\resdef{e1.l1.instruct.bre.cr_rate_std}{0.011}  
\resdef{e1.l1.instruct.inf.cr}{0.008}  
\resdef{e1.l1.instruct.inf.cr_std}{0.005}  
\resdef{e1.l1.instruct.ngp.cr_rate}{0.005}  
\resdef{e1.l1.instruct.ngp.cr_rate_std}{0.000}  
\resdef{e1.l1.instruct.rho}{0.845}  
\resdef{e1.l1.instruct.sge.cr_rate}{0.002}  
\resdef{e1.l1.instruct.sge.cr_rate_std}{0.001}  
\resdef{e1.l1.instruct.split_rate}{0.034}  
\resdef{e1.l1.rlvr.aae.cr}{0.000}  
\resdef{e1.l1.rlvr.aae.cr_std}{0.001}  
\resdef{e1.l1.rlvr.bre.cr_rate}{0.066}  
\resdef{e1.l1.rlvr.bre.cr_rate_std}{0.007}  
\resdef{e1.l1.rlvr.inf.cr}{0.008}  
\resdef{e1.l1.rlvr.inf.cr_std}{0.004}  
\resdef{e1.l1.rlvr.ngp.cr_rate}{0.006}  
\resdef{e1.l1.rlvr.ngp.cr_rate_std}{0.001}  
\resdef{e1.l1.rlvr.rho}{0.845}  
\resdef{e1.l1.rlvr.sge.cr_rate}{0.003}  
\resdef{e1.l1.rlvr.sge.cr_rate_std}{0.001}  
\resdef{e1.l1.rlvr.split_rate}{0.031}  
\resdef{e1.l1.sft.aae.cr}{0.000}  
\resdef{e1.l1.sft.aae.cr_std}{0.001}  
\resdef{e1.l1.sft.bre.cr_rate}{0.078}  
\resdef{e1.l1.sft.bre.cr_rate_std}{0.007}  
\resdef{e1.l1.sft.inf.cr}{0.045}  
\resdef{e1.l1.sft.inf.cr_std}{0.002}  
\resdef{e1.l1.sft.ngp.cr_rate}{0.009}  
\resdef{e1.l1.sft.ngp.cr_rate_std}{0.003}  
\resdef{e1.l1.sft.rho}{0.845}  
\resdef{e1.l1.sft.sge.cr_rate}{0.001}  
\resdef{e1.l1.sft.sge.cr_rate_std}{0.001}  
\resdef{e1.l1.sft.split_rate}{0.076}  
\resdef{e1.l2.base.aae.cr}{0.001}  
\resdef{e1.l2.base.aae.cr_std}{0.001}  
\resdef{e1.l2.base.bre.cr_rate}{0.171}  
\resdef{e1.l2.base.bre.cr_rate_hi}{0.207}  
\resdef{e1.l2.base.bre.cr_rate_lo}{0.137}  
\resdef{e1.l2.base.bre.cr_rate_std}{0.005}  
\resdef{e1.l2.base.inf.cr}{0.023}  
\resdef{e1.l2.base.inf.cr_std}{0.003}  
\resdef{e1.l2.base.ngp.cr_rate}{0.012}  
\resdef{e1.l2.base.ngp.cr_rate_std}{0.004}  
\resdef{e1.l2.base.rho}{0.778}  
\resdef{e1.l2.base.sge.cr_rate}{0.004}  
\resdef{e1.l2.base.sge.cr_rate_std}{0.002}  
\resdef{e1.l2.base.split}{0.230}  
\resdef{e1.l2.base.split_rate}{0.109}  
\resdef{e1.l2.base_f1.aae.cr}{0.017}  
\resdef{e1.l2.base_f1.bre.cr_rate}{0.174}  
\resdef{e1.l2.base_f1.inf.cr}{0.413}  
\resdef{e1.l2.base_f1.sge.cr_rate}{0.055}  
\resdef{e1.l2.base_raw.aae.cr}{0.022}  
\resdef{e1.l2.base_raw.aae.cr_hi}{0.031}  
\resdef{e1.l2.base_raw.aae.cr_lo}{0.013}  
\resdef{e1.l2.base_raw.bre.cr_rate}{0.136}  
\resdef{e1.l2.base_raw.sge.cr_lex}{0.051}  
\resdef{e1.l2.base_raw.sge.cr_lex_hi}{0.072}  
\resdef{e1.l2.base_raw.sge.cr_lex_lo}{0.031}  
\resdef{e1.l2.base_raw.sge.cr_rate}{0.051}  
\resdef{e1.l2.dpo.aae.cr}{0.001}  
\resdef{e1.l2.dpo.aae.cr_std}{0.001}  
\resdef{e1.l2.dpo.aae.cr_tot}{0.001}  
\resdef{e1.l2.dpo.aae.d_cond}{0.001}  
\resdef{e1.l2.dpo.aae.d_sae}{0.000}  
\resdef{e1.l2.dpo.bre.cr_rate}{0.058}  
\resdef{e1.l2.dpo.bre.cr_rate_hi}{0.076}  
\resdef{e1.l2.dpo.bre.cr_rate_lo}{0.040}  
\resdef{e1.l2.dpo.bre.cr_rate_std}{0.001}  
\resdef{e1.l2.dpo.bre.cr_rate_tot}{0.066}  
\resdef{e1.l2.dpo.bre.retention}{0.293}  
\resdef{e1.l2.dpo.for.cr}{0.141}  
\resdef{e1.l2.dpo.inf.cr}{0.009}  
\resdef{e1.l2.dpo.inf.cr_hi}{0.019}  
\resdef{e1.l2.dpo.inf.cr_lo}{\ensuremath{-}0.001}  
\resdef{e1.l2.dpo.inf.cr_std}{0.006}  
\resdef{e1.l2.dpo.ngp.cr_rate}{0.003}  
\resdef{e1.l2.dpo.ngp.cr_rate_std}{0.001}  
\resdef{e1.l2.dpo.rho}{0.845}  
\resdef{e1.l2.dpo.sge.cr_rate}{0.002}  
\resdef{e1.l2.dpo.sge.cr_rate_std}{0.001}  
\resdef{e1.l2.dpo.sge.cr_rate_tot}{0.002}  
\resdef{e1.l2.dpo.split}{0.087}  
\resdef{e1.l2.dpo.split_rate}{0.030}  
\resdef{e1.l2.instruct.aae.cr}{0.000}  
\resdef{e1.l2.instruct.aae.cr_std}{0.001}  
\resdef{e1.l2.instruct.bre.cr_rate}{0.060}  
\resdef{e1.l2.instruct.bre.cr_rate_std}{0.005}  
\resdef{e1.l2.instruct.inf.cr}{0.007}  
\resdef{e1.l2.instruct.inf.cr_std}{0.005}  
\resdef{e1.l2.instruct.ngp.cr_rate}{0.003}  
\resdef{e1.l2.instruct.ngp.cr_rate_std}{0.000}  
\resdef{e1.l2.instruct.rho}{0.845}  
\resdef{e1.l2.instruct.sge.cr_rate}{0.002}  
\resdef{e1.l2.instruct.sge.cr_rate_std}{0.000}  
\resdef{e1.l2.instruct.split_rate}{0.029}  
\resdef{e1.l2.sft.aae.cr}{0.001}  
\resdef{e1.l2.sft.aae.cr_std}{0.001}  
\resdef{e1.l2.sft.bre.cr_rate}{0.120}  
\resdef{e1.l2.sft.bre.cr_rate_std}{0.001}  
\resdef{e1.l2.sft.inf.cr}{0.039}  
\resdef{e1.l2.sft.inf.cr_std}{0.002}  
\resdef{e1.l2.sft.ngp.cr_rate}{0.006}  
\resdef{e1.l2.sft.ngp.cr_rate_std}{0.001}  
\resdef{e1.l2.sft.rho}{0.845}  
\resdef{e1.l2.sft.sge.cr_rate}{0.004}  
\resdef{e1.l2.sft.sge.cr_rate_std}{0.002}  
\resdef{e1.l2.sft.split}{0.141}  
\resdef{e1.l2.sft.split_rate}{0.093}  
\resdef{e1.l3.base.aae.cr}{0.002}  
\resdef{e1.l3.base.aae.cr_std}{0.005}  
\resdef{e1.l3.base.bre.cr_rate}{0.129}  
\resdef{e1.l3.base.bre.cr_rate_std}{0.015}  
\resdef{e1.l3.base.inf.cr}{0.059}  
\resdef{e1.l3.base.inf.cr_std}{0.008}  
\resdef{e1.l3.base.ngp.cr_rate}{0.009}  
\resdef{e1.l3.base.ngp.cr_rate_std}{0.001}  
\resdef{e1.l3.base.rho}{0.845}  
\resdef{e1.l3.base.sge.cr_rate}{0.003}  
\resdef{e1.l3.base.sge.cr_rate_std}{0.002}  
\resdef{e1.l3.base.split_rate}{0.113}  
\resdef{e1.l3.dpo.aae.cr}{0.000}  
\resdef{e1.l3.dpo.aae.cr_std}{0.001}  
\resdef{e1.l3.dpo.bre.cr_rate}{0.099}  
\resdef{e1.l3.dpo.bre.cr_rate_std}{0.003}  
\resdef{e1.l3.dpo.inf.cr}{0.020}  
\resdef{e1.l3.dpo.inf.cr_std}{0.012}  
\resdef{e1.l3.dpo.ngp.cr_rate}{0.005}  
\resdef{e1.l3.dpo.ngp.cr_rate_std}{0.001}  
\resdef{e1.l3.dpo.rho}{0.845}  
\resdef{e1.l3.dpo.sge.cr_rate}{0.002}  
\resdef{e1.l3.dpo.sge.cr_rate_std}{0.001}  
\resdef{e1.l3.dpo.split_rate}{0.063}  
\resdef{e1.l3.instruct.aae.cr}{\ensuremath{-}0.001}  
\resdef{e1.l3.instruct.aae.cr_std}{0.002}  
\resdef{e1.l3.instruct.bre.cr_rate}{0.095}  
\resdef{e1.l3.instruct.bre.cr_rate_std}{0.006}  
\resdef{e1.l3.instruct.inf.cr}{0.017}  
\resdef{e1.l3.instruct.inf.cr_std}{0.005}  
\resdef{e1.l3.instruct.ngp.cr_rate}{0.004}  
\resdef{e1.l3.instruct.ngp.cr_rate_std}{0.001}  
\resdef{e1.l3.instruct.rho}{0.845}  
\resdef{e1.l3.instruct.sge.cr_rate}{0.001}  
\resdef{e1.l3.instruct.sge.cr_rate_std}{0.001}  
\resdef{e1.l3.instruct.split_rate}{0.057}  
\resdef{e1.l3.sft.aae.cr}{0.001}  
\resdef{e1.l3.sft.aae.cr_std}{0.002}  
\resdef{e1.l3.sft.bre.cr_rate}{0.121}  
\resdef{e1.l3.sft.bre.cr_rate_std}{0.009}  
\resdef{e1.l3.sft.inf.cr}{0.049}  
\resdef{e1.l3.sft.inf.cr_std}{0.014}  
\resdef{e1.l3.sft.ngp.cr_rate}{0.009}  
\resdef{e1.l3.sft.ngp.cr_rate_std}{0.002}  
\resdef{e1.l3.sft.rho}{0.845}  
\resdef{e1.l3.sft.sge.cr_rate}{0.001}  
\resdef{e1.l3.sft.sge.cr_rate_std}{0.000}  
\resdef{e1.l3.sft.split_rate}{0.073}  
\resdef{e1.l4.dpo.aae.cr}{\ensuremath{-}0.001}  
\resdef{e1.l4.dpo.aae.cr_std}{0.002}  
\resdef{e1.l4.dpo.bre.cr_rate}{0.085}  
\resdef{e1.l4.dpo.bre.cr_rate_std}{0.003}  
\resdef{e1.l4.dpo.inf.cr}{0.027}  
\resdef{e1.l4.dpo.inf.cr_std}{0.008}  
\resdef{e1.l4.dpo.ngp.cr_rate}{0.004}  
\resdef{e1.l4.dpo.ngp.cr_rate_std}{0.001}  
\resdef{e1.l4.dpo.rho}{0.845}  
\resdef{e1.l4.dpo.sge.cr_rate}{0.005}  
\resdef{e1.l4.dpo.sge.cr_rate_std}{0.000}  
\resdef{e1.l4.dpo.split_rate}{0.059}  
\resdef{e1.l4.rlvr.aae.cr}{0.000}  
\resdef{e1.l4.rlvr.aae.cr_std}{0.002}  
\resdef{e1.l4.rlvr.bre.cr_rate}{0.094}  
\resdef{e1.l4.rlvr.bre.cr_rate_std}{0.002}  
\resdef{e1.l4.rlvr.inf.cr}{0.022}  
\resdef{e1.l4.rlvr.inf.cr_std}{0.006}  
\resdef{e1.l4.rlvr.ngp.cr_rate}{0.005}  
\resdef{e1.l4.rlvr.ngp.cr_rate_std}{0.001}  
\resdef{e1.l4.rlvr.rho}{0.845}  
\resdef{e1.l4.rlvr.sge.cr_rate}{0.005}  
\resdef{e1.l4.rlvr.sge.cr_rate_std}{0.000}  
\resdef{e1.l4.rlvr.split_rate}{0.055}  
\resdef{e1.l4.sft.aae.cr}{\ensuremath{-}0.001}  
\resdef{e1.l4.sft.aae.cr_std}{0.003}  
\resdef{e1.l4.sft.bre.cr_rate}{0.130}  
\resdef{e1.l4.sft.bre.cr_rate_std}{0.006}  
\resdef{e1.l4.sft.inf.cr}{0.080}  
\resdef{e1.l4.sft.inf.cr_std}{0.004}  
\resdef{e1.l4.sft.ngp.cr_rate}{0.046}  
\resdef{e1.l4.sft.ngp.cr_rate_std}{0.007}  
\resdef{e1.l4.sft.rho}{0.845}  
\resdef{e1.l4.sft.sge.cr_rate}{0.020}  
\resdef{e1.l4.sft.sge.cr_rate_std}{0.011}  
\resdef{e1.l4.sft.split_rate}{0.110}  
\resdef{e1.olmo2_13b.base.aae.cr}{0.000}  
\resdef{e1.olmo2_13b.base.aae.cr_std}{0.000}  
\resdef{e1.olmo2_13b.base.bre.cr_rate}{0.179}  
\resdef{e1.olmo2_13b.base.bre.cr_rate_std}{0.010}  
\resdef{e1.olmo2_13b.base.inf.cr}{0.131}  
\resdef{e1.olmo2_13b.base.inf.cr_std}{0.014}  
\resdef{e1.olmo2_13b.base.ngp.cr_rate}{0.071}  
\resdef{e1.olmo2_13b.base.ngp.cr_rate_std}{0.004}  
\resdef{e1.olmo2_13b.base.rho}{0.845}  
\resdef{e1.olmo2_13b.base.sge.cr_rate}{0.017}  
\resdef{e1.olmo2_13b.base.sge.cr_rate_std}{0.009}  
\resdef{e1.olmo2_13b.base.split_rate}{0.154}  
\resdef{e1.olmo2_13b.dpo.aae.cr}{0.001}  
\resdef{e1.olmo2_13b.dpo.aae.cr_std}{0.001}  
\resdef{e1.olmo2_13b.dpo.bre.cr_rate}{0.045}  
\resdef{e1.olmo2_13b.dpo.bre.cr_rate_std}{0.005}  
\resdef{e1.olmo2_13b.dpo.inf.cr}{0.002}  
\resdef{e1.olmo2_13b.dpo.inf.cr_std}{0.003}  
\resdef{e1.olmo2_13b.dpo.ngp.cr_rate}{0.015}  
\resdef{e1.olmo2_13b.dpo.ngp.cr_rate_std}{0.000}  
\resdef{e1.olmo2_13b.dpo.rho}{0.845}  
\resdef{e1.olmo2_13b.dpo.sge.cr_rate}{0.001}  
\resdef{e1.olmo2_13b.dpo.sge.cr_rate_std}{0.001}  
\resdef{e1.olmo2_13b.dpo.split_rate}{0.019}  
\resdef{e1.olmo2_13b.instruct.aae.cr}{0.002}  
\resdef{e1.olmo2_13b.instruct.aae.cr_std}{0.002}  
\resdef{e1.olmo2_13b.instruct.bre.cr_rate}{0.053}  
\resdef{e1.olmo2_13b.instruct.bre.cr_rate_std}{0.004}  
\resdef{e1.olmo2_13b.instruct.inf.cr}{0.007}  
\resdef{e1.olmo2_13b.instruct.inf.cr_std}{0.010}  
\resdef{e1.olmo2_13b.instruct.ngp.cr_rate}{0.012}  
\resdef{e1.olmo2_13b.instruct.ngp.cr_rate_std}{0.003}  
\resdef{e1.olmo2_13b.instruct.rho}{0.845}  
\resdef{e1.olmo2_13b.instruct.sge.cr_rate}{0.001}  
\resdef{e1.olmo2_13b.instruct.sge.cr_rate_std}{0.000}  
\resdef{e1.olmo2_13b.instruct.split_rate}{0.025}  
\resdef{e1.olmo2_13b.sft.aae.cr}{0.004}  
\resdef{e1.olmo2_13b.sft.aae.cr_std}{0.001}  
\resdef{e1.olmo2_13b.sft.bre.cr_rate}{0.090}  
\resdef{e1.olmo2_13b.sft.bre.cr_rate_std}{0.004}  
\resdef{e1.olmo2_13b.sft.inf.cr}{0.040}  
\resdef{e1.olmo2_13b.sft.inf.cr_std}{0.007}  
\resdef{e1.olmo2_13b.sft.ngp.cr_rate}{0.033}  
\resdef{e1.olmo2_13b.sft.ngp.cr_rate_std}{0.013}  
\resdef{e1.olmo2_13b.sft.rho}{0.845}  
\resdef{e1.olmo2_13b.sft.sge.cr_rate}{0.006}  
\resdef{e1.olmo2_13b.sft.sge.cr_rate_std}{0.005}  
\resdef{e1.olmo2_13b.sft.split_rate}{0.067}  
\resdef{e1.olmo2_32b.base.aae.cr}{0.005}  
\resdef{e1.olmo2_32b.base.aae.cr_std}{0.001}  
\resdef{e1.olmo2_32b.base.bre.cr_rate}{0.196}  
\resdef{e1.olmo2_32b.base.bre.cr_rate_std}{0.018}  
\resdef{e1.olmo2_32b.base.inf.cr}{0.062}  
\resdef{e1.olmo2_32b.base.inf.cr_std}{0.002}  
\resdef{e1.olmo2_32b.base.ngp.cr_rate}{0.076}  
\resdef{e1.olmo2_32b.base.ngp.cr_rate_std}{0.021}  
\resdef{e1.olmo2_32b.base.rho}{0.778}  
\resdef{e1.olmo2_32b.base.sge.cr_rate}{0.015}  
\resdef{e1.olmo2_32b.base.sge.cr_rate_std}{0.002}  
\resdef{e1.olmo2_32b.base.split_rate}{0.125}  
\resdef{e1.olmo2_32b.dpo.aae.cr}{0.001}  
\resdef{e1.olmo2_32b.dpo.aae.cr_std}{0.001}  
\resdef{e1.olmo2_32b.dpo.bre.cr_rate}{0.091}  
\resdef{e1.olmo2_32b.dpo.bre.cr_rate_std}{0.008}  
\resdef{e1.olmo2_32b.dpo.inf.cr}{0.020}  
\resdef{e1.olmo2_32b.dpo.inf.cr_std}{0.003}  
\resdef{e1.olmo2_32b.dpo.ngp.cr_rate}{0.015}  
\resdef{e1.olmo2_32b.dpo.ngp.cr_rate_std}{0.002}  
\resdef{e1.olmo2_32b.dpo.rho}{0.845}  
\resdef{e1.olmo2_32b.dpo.sge.cr_rate}{0.012}  
\resdef{e1.olmo2_32b.dpo.sge.cr_rate_std}{0.001}  
\resdef{e1.olmo2_32b.dpo.split_rate}{0.051}  
\resdef{e1.olmo2_32b.instruct.aae.cr}{0.002}  
\resdef{e1.olmo2_32b.instruct.aae.cr_std}{0.004}  
\resdef{e1.olmo2_32b.instruct.bre.cr_rate}{0.096}  
\resdef{e1.olmo2_32b.instruct.bre.cr_rate_std}{0.011}  
\resdef{e1.olmo2_32b.instruct.inf.cr}{0.017}  
\resdef{e1.olmo2_32b.instruct.inf.cr_std}{0.007}  
\resdef{e1.olmo2_32b.instruct.ngp.cr_rate}{0.020}  
\resdef{e1.olmo2_32b.instruct.ngp.cr_rate_std}{0.003}  
\resdef{e1.olmo2_32b.instruct.rho}{0.845}  
\resdef{e1.olmo2_32b.instruct.sge.cr_rate}{0.024}  
\resdef{e1.olmo2_32b.instruct.sge.cr_rate_std}{0.003}  
\resdef{e1.olmo2_32b.instruct.split_rate}{0.044}  
\resdef{e1.olmo2_32b.sft.aae.cr}{0.000}  
\resdef{e1.olmo2_32b.sft.aae.cr_std}{0.002}  
\resdef{e1.olmo2_32b.sft.bre.cr_rate}{0.126}  
\resdef{e1.olmo2_32b.sft.bre.cr_rate_std}{0.009}  
\resdef{e1.olmo2_32b.sft.inf.cr}{0.045}  
\resdef{e1.olmo2_32b.sft.inf.cr_std}{0.009}  
\resdef{e1.olmo2_32b.sft.ngp.cr_rate}{0.031}  
\resdef{e1.olmo2_32b.sft.ngp.cr_rate_std}{0.001}  
\resdef{e1.olmo2_32b.sft.rho}{0.845}  
\resdef{e1.olmo2_32b.sft.sge.cr_rate}{0.007}  
\resdef{e1.olmo2_32b.sft.sge.cr_rate_std}{0.003}  
\resdef{e1.olmo2_32b.sft.split_rate}{0.100}  
\resdef{e1.qwen25_72b.base.aae.cr}{\ensuremath{-}0.001}  
\resdef{e1.qwen25_72b.base.aae.cr_std}{0.001}  
\resdef{e1.qwen25_72b.base.bre.cr_rate}{0.087}  
\resdef{e1.qwen25_72b.base.bre.cr_rate_std}{0.005}  
\resdef{e1.qwen25_72b.base.inf.cr}{0.025}  
\resdef{e1.qwen25_72b.base.inf.cr_std}{0.015}  
\resdef{e1.qwen25_72b.base.ngp.cr_rate}{0.004}  
\resdef{e1.qwen25_72b.base.ngp.cr_rate_std}{0.001}  
\resdef{e1.qwen25_72b.base.rho}{0.845}  
\resdef{e1.qwen25_72b.base.sge.cr_rate}{0.003}  
\resdef{e1.qwen25_72b.base.sge.cr_rate_std}{0.002}  
\resdef{e1.qwen25_72b.base.split_rate}{0.058}  
\resdef{e1.qwen25_72b.instruct.aae.cr}{0.000}  
\resdef{e1.qwen25_72b.instruct.aae.cr_std}{0.001}  
\resdef{e1.qwen25_72b.instruct.bre.cr_rate}{0.045}  
\resdef{e1.qwen25_72b.instruct.bre.cr_rate_std}{0.005}  
\resdef{e1.qwen25_72b.instruct.inf.cr}{0.020}  
\resdef{e1.qwen25_72b.instruct.inf.cr_std}{0.005}  
\resdef{e1.qwen25_72b.instruct.ngp.cr_rate}{0.003}  
\resdef{e1.qwen25_72b.instruct.ngp.cr_rate_std}{0.000}  
\resdef{e1.qwen25_72b.instruct.rho}{0.845}  
\resdef{e1.qwen25_72b.instruct.sge.cr_rate}{0.006}  
\resdef{e1.qwen25_72b.instruct.sge.cr_rate_std}{0.001}  
\resdef{e1.qwen25_72b.instruct.split_rate}{0.034}  
\resdef{e1.olmo31_32b.base.aae.cr}{0.002}  
\resdef{e1.olmo31_32b.base.bre.cr_rate}{0.150}  
\resdef{e1.olmo31_32b.base.inf.cr}{0.106}  
\resdef{e1.olmo31_32b.base.ngp.cr_rate}{0.031}  
\resdef{e1.olmo31_32b.base.sge.cr_rate}{0.007}  
\resdef{e1.olmo31_32b.dpo.aae.cr}{\ensuremath{-}0.001}  
\resdef{e1.olmo31_32b.dpo.bre.cr_rate}{0.078}  
\resdef{e1.olmo31_32b.dpo.inf.cr}{0.018}  
\resdef{e1.olmo31_32b.dpo.ngp.cr_rate}{0.030}  
\resdef{e1.olmo31_32b.dpo.sge.cr_rate}{0.011}  
\resdef{e1.olmo31_32b.rlvr.aae.cr}{0.003}  
\resdef{e1.olmo31_32b.rlvr.bre.cr_rate}{0.081}  
\resdef{e1.olmo31_32b.rlvr.inf.cr}{0.008}  
\resdef{e1.olmo31_32b.rlvr.ngp.cr_rate}{0.027}  
\resdef{e1.olmo31_32b.rlvr.sge.cr_rate}{0.013}  
\resdef{e1.olmo31_32b.sft.aae.cr}{0.002}  
\resdef{e1.olmo31_32b.sft.bre.cr_rate}{0.075}  
\resdef{e1.olmo31_32b.sft.inf.cr}{0.019}  
\resdef{e1.olmo31_32b.sft.ngp.cr_rate}{0.028}  
\resdef{e1.olmo31_32b.sft.sge.cr_rate}{0.008}  
\resdef{e1.tulu2_13b.dpo.aae.cr}{0.000}  
\resdef{e1.tulu2_13b.dpo.bre.cr_rate}{0.045}  
\resdef{e1.tulu2_13b.dpo.inf.cr}{0.023}  
\resdef{e1.tulu2_13b.dpo.ngp.cr_rate}{0.007}  
\resdef{e1.tulu2_13b.dpo.sge.cr_rate}{0.004}  
\resdef{e1.tulu2_13b.sft.aae.cr}{0.000}  
\resdef{e1.tulu2_13b.sft.bre.cr_rate}{0.073}  
\resdef{e1.tulu2_13b.sft.inf.cr}{0.028}  
\resdef{e1.tulu2_13b.sft.ngp.cr_rate}{0.006}  
\resdef{e1.tulu2_13b.sft.sge.cr_rate}{0.003}  
\resdef{e1.tulu2_7b.dpo.aae.cr}{0.000}  
\resdef{e1.tulu2_7b.dpo.bre.cr_rate}{0.051}  
\resdef{e1.tulu2_7b.dpo.inf.cr}{0.024}  
\resdef{e1.tulu2_7b.dpo.ngp.cr_rate}{0.003}  
\resdef{e1.tulu2_7b.dpo.sge.cr_rate}{0.005}  
\resdef{e1.tulu2_7b.sft.aae.cr}{0.001}  
\resdef{e1.tulu2_7b.sft.bre.cr_rate}{0.064}  
\resdef{e1.tulu2_7b.sft.inf.cr}{0.036}  
\resdef{e1.tulu2_7b.sft.ngp.cr_rate}{0.005}  
\resdef{e1.tulu2_7b.sft.sge.cr_rate}{0.004}  
\resdef{e1.zephyr_7b.base.aae.cr}{0.000}  
\resdef{e1.zephyr_7b.base.bre.cr_rate}{0.193}  
\resdef{e1.zephyr_7b.base.inf.cr}{0.422}  
\resdef{e1.zephyr_7b.base.ngp.cr_rate}{0.024}  
\resdef{e1.zephyr_7b.base.sge.cr_rate}{0.021}  
\resdef{e1.zephyr_7b.dpo.aae.cr}{0.000}  
\resdef{e1.zephyr_7b.dpo.bre.cr_rate}{0.066}  
\resdef{e1.zephyr_7b.dpo.inf.cr}{0.013}  
\resdef{e1.zephyr_7b.dpo.ngp.cr_rate}{0.004}  
\resdef{e1.zephyr_7b.dpo.sge.cr_rate}{0.006}  
\resdef{e1.zephyr_7b.sft.aae.cr}{0.000}  
\resdef{e1.zephyr_7b.sft.bre.cr_rate}{0.125}  
\resdef{e1.zephyr_7b.sft.inf.cr}{0.032}  
\resdef{e1.zephyr_7b.sft.ngp.cr_rate}{0.007}  
\resdef{e1.zephyr_7b.sft.sge.cr_rate}{0.004}  
\resdef{e1.olmo31_32b.base.rho}{0.845}  
\resdef{e1.olmo31_32b.base.split_rate}{0.085}  
\resdef{e1.olmo31_32b.dpo.rho}{0.845}  
\resdef{e1.olmo31_32b.dpo.split_rate}{0.044}  
\resdef{e1.olmo31_32b.rlvr.rho}{0.845}  
\resdef{e1.olmo31_32b.rlvr.split_rate}{0.040}  
\resdef{e1.olmo31_32b.sft.rho}{0.845}  
\resdef{e1.olmo31_32b.sft.split_rate}{0.040}  
\resdef{e1.tulu2_13b.dpo.rho}{0.778}  
\resdef{e1.tulu2_13b.dpo.split_rate}{0.037}  
\resdef{e1.tulu2_13b.sft.rho}{0.778}  
\resdef{e1.tulu2_13b.sft.split_rate}{0.047}  
\resdef{e1.tulu2_7b.dpo.rho}{0.845}  
\resdef{e1.tulu2_7b.dpo.split_rate}{0.042}  
\resdef{e1.tulu2_7b.sft.rho}{0.778}  
\resdef{e1.tulu2_7b.sft.split_rate}{0.053}  
\resdef{e1.zephyr_7b.base.rho}{0.778}  
\resdef{e1.zephyr_7b.base.split_rate}{0.251}  
\resdef{e1.zephyr_7b.dpo.rho}{0.845}  
\resdef{e1.zephyr_7b.dpo.split_rate}{0.046}  
\resdef{e1.zephyr_7b.sft.rho}{0.845}  
\resdef{e1.zephyr_7b.sft.split_rate}{0.073}  
\resdef{e1b.dsllm_7b.base.aae.cr}{0.010}  
\resdef{e1b.dsllm_7b.base.bre.cr_rate}{0.026}  
\resdef{e1b.dsllm_7b.base.inf.cr}{0.114}  
\resdef{e1b.dsllm_7b.base.ngp.cr_rate}{0.001}  
\resdef{e1b.dsllm_7b.base.sge.cr_rate}{0.024}  
\resdef{e1b.dsllm_7b.chat.aae.cr}{0.000}  
\resdef{e1b.dsllm_7b.chat.bre.cr_rate}{0.034}  
\resdef{e1b.dsllm_7b.chat.inf.cr}{0.019}  
\resdef{e1b.dsllm_7b.chat.ngp.cr_rate}{0.000}  
\resdef{e1b.dsllm_7b.chat.sge.cr_rate}{0.000}  
\resdef{e1b.dsv2lite.base.aae.cr}{0.007}  
\resdef{e1b.dsv2lite.base.bre.cr_rate}{0.220}  
\resdef{e1b.dsv2lite.base.inf.cr}{0.385}  
\resdef{e1b.dsv2lite.base.ngp.cr_rate}{0.028}  
\resdef{e1b.dsv2lite.base.sge.cr_rate}{0.032}  
\resdef{e1b.dsv2lite.chat.aae.cr}{0.001}  
\resdef{e1b.dsv2lite.chat.bre.cr_rate}{0.070}  
\resdef{e1b.dsv2lite.chat.inf.cr}{0.012}  
\resdef{e1b.dsv2lite.chat.ngp.cr_rate}{0.010}  
\resdef{e1b.dsv2lite.chat.sge.cr_rate}{0.003}  
\resdef{e1b.falcon3_7b.base.aae.cr}{0.000}  
\resdef{e1b.falcon3_7b.base.bre.cr_rate}{0.056}  
\resdef{e1b.falcon3_7b.base.inf.cr}{0.034}  
\resdef{e1b.falcon3_7b.base.ngp.cr_rate}{0.009}  
\resdef{e1b.falcon3_7b.base.sge.cr_rate}{0.003}  
\resdef{e1b.falcon3_7b.instruct.aae.cr}{0.000}  
\resdef{e1b.falcon3_7b.instruct.bre.cr_rate}{0.081}  
\resdef{e1b.falcon3_7b.instruct.inf.cr}{0.018}  
\resdef{e1b.falcon3_7b.instruct.ngp.cr_rate}{0.005}  
\resdef{e1b.falcon3_7b.instruct.sge.cr_rate}{0.003}  
\resdef{e1b.granite33_8b.base.aae.cr}{0.092}  
\resdef{e1b.granite33_8b.base.bre.cr_rate}{0.076}  
\resdef{e1b.granite33_8b.base.inf.cr}{0.238}  
\resdef{e1b.granite33_8b.base.ngp.cr_rate}{0.055}  
\resdef{e1b.granite33_8b.base.sge.cr_rate}{0.136}  
\resdef{e1b.granite33_8b.instruct.aae.cr}{0.001}  
\resdef{e1b.granite33_8b.instruct.bre.cr_rate}{0.064}  
\resdef{e1b.granite33_8b.instruct.inf.cr}{0.017}  
\resdef{e1b.granite33_8b.instruct.ngp.cr_rate}{0.004}  
\resdef{e1b.granite33_8b.instruct.sge.cr_rate}{0.003}  
\resdef{e1b.mimo_7b.base.aae.cr}{0.001}  
\resdef{e1b.mimo_7b.base.bre.cr_rate}{0.127}  
\resdef{e1b.mimo_7b.base.inf.cr}{0.276}  
\resdef{e1b.mimo_7b.base.ngp.cr_rate}{0.046}  
\resdef{e1b.mimo_7b.base.sge.cr_rate}{0.009}  
\resdef{e1b.mimo_7b.rl.aae.cr}{\ensuremath{-}0.001}  
\resdef{e1b.mimo_7b.rl.bre.cr_rate}{0.039}  
\resdef{e1b.mimo_7b.rl.inf.cr}{\ensuremath{-}0.008}  
\resdef{e1b.mimo_7b.rl.ngp.cr_rate}{0.021}  
\resdef{e1b.mimo_7b.rl.sge.cr_rate}{0.001}  
\resdef{e1b.mimo_7b.sft.aae.cr}{0.000}  
\resdef{e1b.mimo_7b.sft.bre.cr_rate}{0.033}  
\resdef{e1b.mimo_7b.sft.inf.cr}{\ensuremath{-}0.009}  
\resdef{e1b.mimo_7b.sft.ngp.cr_rate}{0.018}  
\resdef{e1b.mimo_7b.sft.sge.cr_rate}{0.002}  
\resdef{e1b.mistral_24b.base.aae.cr}{0.006}  
\resdef{e1b.mistral_24b.base.bre.cr_rate}{0.203}  
\resdef{e1b.mistral_24b.base.inf.cr}{0.208}  
\resdef{e1b.mistral_24b.base.ngp.cr_rate}{0.032}  
\resdef{e1b.mistral_24b.base.sge.cr_rate}{0.010}  
\resdef{e1b.mistral_24b.instruct.aae.cr}{0.001}  
\resdef{e1b.mistral_24b.instruct.bre.cr_rate}{0.081}  
\resdef{e1b.mistral_24b.instruct.inf.cr}{0.028}  
\resdef{e1b.mistral_24b.instruct.ngp.cr_rate}{0.004}  
\resdef{e1b.mistral_24b.instruct.sge.cr_rate}{0.004}  
\resdef{e1b.mistral_7b.base.aae.cr}{0.002}  
\resdef{e1b.mistral_7b.base.bre.cr_rate}{0.222}  
\resdef{e1b.mistral_7b.base.inf.cr}{0.408}  
\resdef{e1b.mistral_7b.base.ngp.cr_rate}{0.027}  
\resdef{e1b.mistral_7b.base.sge.cr_rate}{0.046}  
\resdef{e1b.mistral_7b.instruct.aae.cr}{0.000}  
\resdef{e1b.mistral_7b.instruct.bre.cr_rate}{0.033}  
\resdef{e1b.mistral_7b.instruct.inf.cr}{0.012}  
\resdef{e1b.mistral_7b.instruct.ngp.cr_rate}{0.001}  
\resdef{e1b.mistral_7b.instruct.sge.cr_rate}{0.003}  
\resdef{e1b.qwen3_8b.base.aae.cr}{0.004}  
\resdef{e1b.qwen3_8b.base.bre.cr_rate}{0.124}  
\resdef{e1b.qwen3_8b.base.inf.cr}{0.097}  
\resdef{e1b.qwen3_8b.base.ngp.cr_rate}{0.009}  
\resdef{e1b.qwen3_8b.base.sge.cr_rate}{0.004}  
\resdef{e1b.qwen3_8b.instruct.aae.cr}{0.000}  
\resdef{e1b.qwen3_8b.instruct.bre.cr_rate}{0.061}  
\resdef{e1b.qwen3_8b.instruct.inf.cr}{0.036}  
\resdef{e1b.qwen3_8b.instruct.ngp.cr_rate}{0.009}  
\resdef{e1b.qwen3_8b.instruct.sge.cr_rate}{0.009}  
\resdef{e1b.smollm3_3b.apo.aae.cr}{0.002}  
\resdef{e1b.smollm3_3b.apo.bre.cr_rate}{0.079}  
\resdef{e1b.smollm3_3b.apo.inf.cr}{0.015}  
\resdef{e1b.smollm3_3b.apo.ngp.cr_rate}{0.015}  
\resdef{e1b.smollm3_3b.apo.sge.cr_rate}{0.002}  
\resdef{e1b.smollm3_3b.base.aae.cr}{0.000}  
\resdef{e1b.smollm3_3b.base.bre.cr_rate}{0.135}  
\resdef{e1b.smollm3_3b.base.inf.cr}{0.066}  
\resdef{e1b.smollm3_3b.base.ngp.cr_rate}{0.013}  
\resdef{e1b.smollm3_3b.base.sge.cr_rate}{0.002}  
\resdef{e1b.smollm3_3b.mid.aae.cr}{0.000}  
\resdef{e1b.smollm3_3b.mid.bre.cr_rate}{0.033}  
\resdef{e1b.smollm3_3b.mid.inf.cr}{\ensuremath{-}0.002}  
\resdef{e1b.smollm3_3b.mid.ngp.cr_rate}{0.015}  
\resdef{e1b.smollm3_3b.mid.sge.cr_rate}{0.003}  
\resdef{e1b.smollm3_3b.rlvr.aae.cr}{0.001}  
\resdef{e1b.smollm3_3b.rlvr.bre.cr_rate}{0.071}  
\resdef{e1b.smollm3_3b.rlvr.inf.cr}{0.014}  
\resdef{e1b.smollm3_3b.rlvr.ngp.cr_rate}{0.014}  
\resdef{e1b.smollm3_3b.rlvr.sge.cr_rate}{0.001}  
\resdef{e1b.smollm3_3b.sft.aae.cr}{0.000}  
\resdef{e1b.smollm3_3b.sft.bre.cr_rate}{0.088}  
\resdef{e1b.smollm3_3b.sft.inf.cr}{0.010}  
\resdef{e1b.smollm3_3b.sft.ngp.cr_rate}{0.020}  
\resdef{e1b.smollm3_3b.sft.sge.cr_rate}{0.002}  
\resdef{e1b.yi15_9b.base.aae.cr}{0.002}  
\resdef{e1b.yi15_9b.base.bre.cr_rate}{0.045}  
\resdef{e1b.yi15_9b.base.inf.cr}{0.051}  
\resdef{e1b.yi15_9b.base.ngp.cr_rate}{0.029}  
\resdef{e1b.yi15_9b.base.sge.cr_rate}{0.013}  
\resdef{e1b.yi15_9b.chat.aae.cr}{0.000}  
\resdef{e1b.yi15_9b.chat.bre.cr_rate}{0.058}  
\resdef{e1b.yi15_9b.chat.inf.cr}{0.030}  
\resdef{e1b.yi15_9b.chat.ngp.cr_rate}{0.008}  
\resdef{e1b.yi15_9b.chat.sge.cr_rate}{0.001}  
\resdef{e1b.dsllm_7b.base.rho}{0.778}  
\resdef{e1b.dsllm_7b.base.split_rate}{0.036}  
\resdef{e1b.dsllm_7b.chat.rho}{0.778}  
\resdef{e1b.dsllm_7b.chat.split_rate}{0.038}  
\resdef{e1b.dsv2lite.base.rho}{0.778}  
\resdef{e1b.dsv2lite.base.split_rate}{0.291}  
\resdef{e1b.dsv2lite.chat.rho}{0.845}  
\resdef{e1b.dsv2lite.chat.split_rate}{0.043}  
\resdef{e1b.falcon3_7b.base.rho}{0.845}  
\resdef{e1b.falcon3_7b.base.split_rate}{0.055}  
\resdef{e1b.falcon3_7b.instruct.rho}{0.845}  
\resdef{e1b.falcon3_7b.instruct.split_rate}{0.052}  
\resdef{e1b.granite33_8b.base.rho}{0.034}  
\resdef{e1b.granite33_8b.base.split_rate}{0.268}  
\resdef{e1b.granite33_8b.instruct.rho}{0.845}  
\resdef{e1b.granite33_8b.instruct.split_rate}{0.042}  
\resdef{e1b.mimo_7b.base.rho}{0.778}  
\resdef{e1b.mimo_7b.base.split_rate}{0.112}  
\resdef{e1b.mimo_7b.rl.rho}{0.845}  
\resdef{e1b.mimo_7b.rl.split_rate}{0.003}  
\resdef{e1b.mimo_7b.sft.rho}{0.845}  
\resdef{e1b.mimo_7b.sft.split_rate}{\ensuremath{-}0.001}  
\resdef{e1b.mistral_24b.base.rho}{0.778}  
\resdef{e1b.mistral_24b.base.split_rate}{0.196}  
\resdef{e1b.mistral_24b.instruct.rho}{0.845}  
\resdef{e1b.mistral_24b.instruct.split_rate}{0.052}  
\resdef{e1b.mistral_7b.base.rho}{0.778}  
\resdef{e1b.mistral_7b.base.split_rate}{0.364}  
\resdef{e1b.mistral_7b.instruct.rho}{0.778}  
\resdef{e1b.mistral_7b.instruct.split_rate}{0.027}  
\resdef{e1b.qwen3_8b.base.rho}{0.845}  
\resdef{e1b.qwen3_8b.base.split_rate}{0.098}  
\resdef{e1b.qwen3_8b.instruct.rho}{0.845}  
\resdef{e1b.qwen3_8b.instruct.split_rate}{0.055}  
\resdef{e1b.smollm3_3b.apo.rho}{0.845}  
\resdef{e1b.smollm3_3b.apo.split_rate}{0.044}  
\resdef{e1b.smollm3_3b.base.rho}{0.845}  
\resdef{e1b.smollm3_3b.base.split_rate}{0.118}  
\resdef{e1b.smollm3_3b.mid.rho}{0.845}  
\resdef{e1b.smollm3_3b.mid.split_rate}{0.005}  
\resdef{e1b.smollm3_3b.rlvr.rho}{0.778}  
\resdef{e1b.smollm3_3b.rlvr.split_rate}{0.042}  
\resdef{e1b.smollm3_3b.sft.rho}{0.778}  
\resdef{e1b.smollm3_3b.sft.split_rate}{0.052}  
\resdef{e1b.yi15_9b.base.rho}{0.778}  
\resdef{e1b.yi15_9b.base.split_rate}{0.097}  
\resdef{e1b.yi15_9b.chat.rho}{0.845}  
\resdef{e1b.yi15_9b.chat.split_rate}{0.037}  
\resdef{e1n.qwen35_4b.base.aae.cr}{0.000}  
\resdef{e1n.qwen35_4b.base.bre.cr_rate}{0.113}  
\resdef{e1n.qwen35_4b.base.inf.cr}{0.048}  
\resdef{e1n.qwen35_4b.base.ngp.cr_rate}{0.021}  
\resdef{e1n.qwen35_4b.base.sge.cr_rate}{0.008}  
\resdef{e1n.qwen35_4b.instruct.aae.cr}{\ensuremath{-}0.001}  
\resdef{e1n.qwen35_4b.instruct.bre.cr_rate}{0.055}  
\resdef{e1n.qwen35_4b.instruct.inf.cr}{0.012}  
\resdef{e1n.qwen35_4b.instruct.ngp.cr_rate}{0.029}  
\resdef{e1n.qwen35_4b.instruct.sge.cr_rate}{0.010}  
\resdef{e1n.qwen35_9b.base.aae.cr}{0.001}  
\resdef{e1n.qwen35_9b.base.bre.cr_rate}{0.076}  
\resdef{e1n.qwen35_9b.base.inf.cr}{0.043}  
\resdef{e1n.qwen35_9b.base.ngp.cr_rate}{0.023}  
\resdef{e1n.qwen35_9b.base.sge.cr_rate}{0.006}  
\resdef{e1n.qwen35_9b.instruct.aae.cr}{\ensuremath{-}0.001}  
\resdef{e1n.qwen35_9b.instruct.bre.cr_rate}{0.047}  
\resdef{e1n.qwen35_9b.instruct.inf.cr}{0.007}  
\resdef{e1n.qwen35_9b.instruct.ngp.cr_rate}{0.042}  
\resdef{e1n.qwen35_9b.instruct.sge.cr_rate}{0.014}  
\resdef{e1n.qwen35_4b.base.rho}{0.778}  
\resdef{e1n.qwen35_4b.base.split_rate}{0.084}  
\resdef{e1n.qwen35_4b.instruct.rho}{0.845}  
\resdef{e1n.qwen35_4b.instruct.split_rate}{0.029}  
\resdef{e1n.qwen35_9b.base.rho}{0.778}  
\resdef{e1n.qwen35_9b.base.split_rate}{0.060}  
\resdef{e1n.qwen35_9b.instruct.rho}{0.845}  
\resdef{e1n.qwen35_9b.instruct.split_rate}{0.022}  
\resdef{e1n.qwen38_27b.chat.aae.cr}{0.006}  
\resdef{e1n.qwen38_27b.chat.bre.cr_rate}{0.050}  
\resdef{e1n.qwen38_27b.chat.inf.cr}{0.033}  
\resdef{e1n.qwen38_27b.chat.ngp.cr_rate}{0.061}  
\resdef{e1n.qwen38_27b.chat.sge.cr_rate}{0.010}  
\resdef{e1n.qwen38_27b.chat.rho}{0.439}  
\resdef{e1n.qwen38_27b.chat.split_rate}{0.042}  
\resdef{e1o.openpangu_embedded_7b_v11.chat.aae.cr}{0.002}  
\resdef{e1o.openpangu_embedded_7b_v11.chat.bre.cr_rate}{0.096}  
\resdef{e1o.openpangu_embedded_7b_v11.chat.inf.cr}{0.018}  
\resdef{e1o.openpangu_embedded_7b_v11.chat.ngp.cr_rate}{0.018}  
\resdef{e1o.openpangu_embedded_7b_v11.chat.sge.cr_rate}{0.015}  
\resdef{e1o.openpangu_embedded_7b_v11.chat_enprompt.aae.cr}{0.000}  
\resdef{e1o.openpangu_embedded_7b_v11.chat_enprompt.bre.cr_rate}{0.115}  
\resdef{e1o.openpangu_embedded_7b_v11.chat_enprompt.inf.cr}{0.023}  
\resdef{e1o.openpangu_embedded_7b_v11.chat_enprompt.ngp.cr_rate}{0.016}  
\resdef{e1o.openpangu_embedded_7b_v11.chat_enprompt.sge.cr_rate}{0.002}  
\resdef{e1o.openpangu_embedded_7b_v11.chat_nosys.aae.cr}{0.004}  
\resdef{e1o.openpangu_embedded_7b_v11.chat_nosys.bre.cr_rate}{0.060}  
\resdef{e1o.openpangu_embedded_7b_v11.chat_nosys.inf.cr}{0.036}  
\resdef{e1o.openpangu_embedded_7b_v11.chat_nosys.ngp.cr_rate}{0.030}  
\resdef{e1o.openpangu_embedded_7b_v11.chat_nosys.sge.cr_rate}{0.008}  
\resdef{e1o.openpangu_embedded_7b_v11.chat.rho}{0.845}  
\resdef{e1o.openpangu_embedded_7b_v11.chat.split_rate}{0.052}  
\resdef{e1o.openpangu_embedded_7b_v11.chat_enprompt.rho}{0.845}  
\resdef{e1o.openpangu_embedded_7b_v11.chat_enprompt.split_rate}{0.067}  
\resdef{e1o.openpangu_embedded_7b_v11.chat_nosys.rho}{0.845}  
\resdef{e1o.openpangu_embedded_7b_v11.chat_nosys.split_rate}{0.052}  
\resdef{e1d.dsv4pro.deployed.aae.cr}{0.009}  
\resdef{e1d.dsv4pro.deployed.bre.cr_rate}{0.190}  
\resdef{e1d.dsv4pro.deployed.inf.cr}{0.037}  
\resdef{e1d.dsv4pro.deployed.ngp.cr_rate}{0.385}  
\resdef{e1d.dsv4pro.deployed.sge.cr_rate}{0.078}  
\resdef{e1d.gemini25flash.deployed.aae.cr}{0.002}  
\resdef{e1d.gemini25flash.deployed.bre.cr_rate}{0.089}  
\resdef{e1d.gemini25flash.deployed.inf.cr}{0.013}  
\resdef{e1d.gemini25flash.deployed.ngp.cr_rate}{0.189}  
\resdef{e1d.gemini25flash.deployed.sge.cr_rate}{0.070}  
\resdef{e1d.gpt53chat.deployed.aae.cr}{0.006}  
\resdef{e1d.gpt53chat.deployed.bre.cr_rate}{0.187}  
\resdef{e1d.gpt53chat.deployed.inf.cr}{0.053}  
\resdef{e1d.gpt53chat.deployed.ngp.cr_rate}{0.116}  
\resdef{e1d.gpt53chat.deployed.sge.cr_rate}{0.024}  
\resdef{e1d.kimik25.deployed.aae.cr}{0.007}  
\resdef{e1d.kimik25.deployed.bre.cr_rate}{0.191}  
\resdef{e1d.kimik25.deployed.inf.cr}{0.012}  
\resdef{e1d.kimik25.deployed.ngp.cr_rate}{0.502}  
\resdef{e1d.kimik25.deployed.sge.cr_rate}{0.119}  
\resdef{e1d.sonnet46.deployed.aae.cr}{\ensuremath{-}0.003}  
\resdef{e1d.sonnet46.deployed.bre.cr_rate}{0.207}  
\resdef{e1d.sonnet46.deployed.inf.cr}{0.022}  
\resdef{e1d.sonnet46.deployed.ngp.cr_rate}{0.148}  
\resdef{e1d.sonnet46.deployed.sge.cr_rate}{0.020}  
\resdef{e1d.dsv4pro.deployed.rho}{0.439}  
\resdef{e1d.dsv4pro.deployed.split_rate}{0.006}  
\resdef{e1d.gemini25flash.deployed.rho}{0.270}  
\resdef{e1d.gemini25flash.deployed.split_rate}{\ensuremath{-}0.013}  
\resdef{e1d.gpt53chat.deployed.rho}{0.845}  
\resdef{e1d.gpt53chat.deployed.split_rate}{0.085}  
\resdef{e1d.kimik25.deployed.rho}{0.439}  
\resdef{e1d.kimik25.deployed.split_rate}{\ensuremath{-}0.044}  
\resdef{e1d.sonnet46.deployed.rho}{0.845}  
\resdef{e1d.sonnet46.deployed.split_rate}{0.101}  
\resdef{comp.l2.dpo.aae.comp}{0.620}  
\resdef{comp.l2.dpo.aae.default_ratio}{0.009}  
\resdef{comp.l2.dpo.sge.default_ratio}{0.002}  
\resdef{comp.l3.dpo.aae.default_ratio}{0.015}  
\resdef{comp.l3.dpo.sge.default_ratio}{0.006}  
\resdef{comp.l4.dpo.aae.default_ratio}{0.009}  
\resdef{comp.l4.dpo.sge.default_ratio}{0.007}  
\resdef{g2r.l2.dpo.aae.dF_v_minus_dF_inf}{\ensuremath{-}0.015}  
\resdef{g2r.l2.dpo.aae.hi}{\ensuremath{-}0.008}  
\resdef{g2r.l2.dpo.aae.lo}{\ensuremath{-}0.023}  
\resdef{au.l2.dpo.auth_gro.aae.cr}{0.009}  
\resdef{e4.l1.dpo_vs_sft.aae.mb}{\ensuremath{-}0.009}  
\resdef{e4.l1.dpo_vs_sft.aae.mb_hi}{\ensuremath{-}0.006}  
\resdef{e4.l1.dpo_vs_sft.aae.mb_lo}{\ensuremath{-}0.012}  
\resdef{e4.l1.dpo_vs_sft.aus.mb}{0.005}  
\resdef{e4.l1.dpo_vs_sft.aus.mb_hi}{0.009}  
\resdef{e4.l1.dpo_vs_sft.aus.mb_lo}{0.002}  
\resdef{e4.l1.dpo_vs_sft.bre.mb}{0.004}  
\resdef{e4.l1.dpo_vs_sft.bre.mb_hi}{0.007}  
\resdef{e4.l1.dpo_vs_sft.bre.mb_lo}{0.001}  
\resdef{e4.l1.dpo_vs_sft.inf.mb}{\ensuremath{-}0.007}  
\resdef{e4.l1.dpo_vs_sft.inf.mb_hi}{\ensuremath{-}0.004}  
\resdef{e4.l1.dpo_vs_sft.inf.mb_lo}{\ensuremath{-}0.010}  
\resdef{e4.l1.dpo_vs_sft.sge.mb}{0.005}  
\resdef{e4.l1.dpo_vs_sft.sge.mb_hi}{0.011}  
\resdef{e4.l1.dpo_vs_sft.sge.mb_lo}{\ensuremath{-}0.001}  
\resdef{e4.l1.sft_vs_base.aae.mb}{\ensuremath{-}0.017}  
\resdef{e4.l1.sft_vs_base.aae.mb_hi}{\ensuremath{-}0.013}  
\resdef{e4.l1.sft_vs_base.aae.mb_lo}{\ensuremath{-}0.022}  
\resdef{e4.l1.sft_vs_base.aus.mb}{0.010}  
\resdef{e4.l1.sft_vs_base.aus.mb_hi}{0.014}  
\resdef{e4.l1.sft_vs_base.aus.mb_lo}{0.005}  
\resdef{e4.l1.sft_vs_base.bre.mb}{0.010}  
\resdef{e4.l1.sft_vs_base.bre.mb_hi}{0.014}  
\resdef{e4.l1.sft_vs_base.bre.mb_lo}{0.005}  
\resdef{e4.l1.sft_vs_base.inf.mb}{\ensuremath{-}0.032}  
\resdef{e4.l1.sft_vs_base.inf.mb_hi}{\ensuremath{-}0.028}  
\resdef{e4.l1.sft_vs_base.inf.mb_lo}{\ensuremath{-}0.036}  
\resdef{e4.l1.sft_vs_base.sge.mb}{\ensuremath{-}0.032}  
\resdef{e4.l1.sft_vs_base.sge.mb_hi}{\ensuremath{-}0.025}  
\resdef{e4.l1.sft_vs_base.sge.mb_lo}{\ensuremath{-}0.040}  
\resdef{e4.l2.dpo_vs_sft.aae.mb}{\ensuremath{-}0.018}  
\resdef{e4.l2.dpo_vs_sft.aae.mb_hi}{\ensuremath{-}0.012}  
\resdef{e4.l2.dpo_vs_sft.aae.mb_lo}{\ensuremath{-}0.024}  
\resdef{e4.l2.dpo_vs_sft.aus.mb}{0.001}  
\resdef{e4.l2.dpo_vs_sft.aus.mb_hi}{0.008}  
\resdef{e4.l2.dpo_vs_sft.aus.mb_lo}{\ensuremath{-}0.005}  
\resdef{e4.l2.dpo_vs_sft.bre.mb}{0.002}  
\resdef{e4.l2.dpo_vs_sft.bre.mb_hi}{0.009}  
\resdef{e4.l2.dpo_vs_sft.bre.mb_lo}{\ensuremath{-}0.005}  
\resdef{e4.l2.dpo_vs_sft.inf.mb}{\ensuremath{-}0.004}  
\resdef{e4.l2.dpo_vs_sft.inf.mb_hi}{0.002}  
\resdef{e4.l2.dpo_vs_sft.inf.mb_lo}{\ensuremath{-}0.010}  
\resdef{e4.l2.dpo_vs_sft.sge.mb}{\ensuremath{-}0.007}  
\resdef{e4.l2.dpo_vs_sft.sge.mb_hi}{0.001}  
\resdef{e4.l2.dpo_vs_sft.sge.mb_lo}{\ensuremath{-}0.017}  
\resdef{e4.l2.instruct_vs_dpo.aae.mb}{\ensuremath{-}0.002}  
\resdef{e4.l2.instruct_vs_dpo.aae.mb_hi}{\ensuremath{-}0.001}  
\resdef{e4.l2.instruct_vs_dpo.aae.mb_lo}{\ensuremath{-}0.003}  
\resdef{e4.l2.instruct_vs_dpo.aus.mb}{0.002}  
\resdef{e4.l2.instruct_vs_dpo.aus.mb_hi}{0.003}  
\resdef{e4.l2.instruct_vs_dpo.aus.mb_lo}{0.001}  
\resdef{e4.l2.instruct_vs_dpo.bre.mb}{0.001}  
\resdef{e4.l2.instruct_vs_dpo.bre.mb_hi}{0.002}  
\resdef{e4.l2.instruct_vs_dpo.bre.mb_lo}{0.001}  
\resdef{e4.l2.instruct_vs_dpo.inf.mb}{\ensuremath{-}0.001}  
\resdef{e4.l2.instruct_vs_dpo.inf.mb_hi}{0.000}  
\resdef{e4.l2.instruct_vs_dpo.inf.mb_lo}{\ensuremath{-}0.002}  
\resdef{e4.l2.instruct_vs_dpo.sge.mb}{0.001}  
\resdef{e4.l2.instruct_vs_dpo.sge.mb_hi}{0.002}  
\resdef{e4.l2.instruct_vs_dpo.sge.mb_lo}{\ensuremath{-}0.001}  
\resdef{e4.l2.sft_vs_base.aae.mb}{\ensuremath{-}0.015}  
\resdef{e4.l2.sft_vs_base.aae.mb_hi}{\ensuremath{-}0.012}  
\resdef{e4.l2.sft_vs_base.aae.mb_lo}{\ensuremath{-}0.019}  
\resdef{e4.l2.sft_vs_base.aus.mb}{0.015}  
\resdef{e4.l2.sft_vs_base.aus.mb_hi}{0.021}  
\resdef{e4.l2.sft_vs_base.aus.mb_lo}{0.009}  
\resdef{e4.l2.sft_vs_base.bre.mb}{0.011}  
\resdef{e4.l2.sft_vs_base.bre.mb_hi}{0.018}  
\resdef{e4.l2.sft_vs_base.bre.mb_lo}{0.005}  
\resdef{e4.l2.sft_vs_base.inf.mb}{\ensuremath{-}0.037}  
\resdef{e4.l2.sft_vs_base.inf.mb_hi}{\ensuremath{-}0.032}  
\resdef{e4.l2.sft_vs_base.inf.mb_lo}{\ensuremath{-}0.043}  
\resdef{e4.l2.sft_vs_base.sge.mb}{\ensuremath{-}0.021}  
\resdef{e4.l2.sft_vs_base.sge.mb_hi}{\ensuremath{-}0.014}  
\resdef{e4.l2.sft_vs_base.sge.mb_lo}{\ensuremath{-}0.028}  
\resdef{e4.l3.dpo_vs_sft.aae.mb}{\ensuremath{-}0.026}  
\resdef{e4.l3.dpo_vs_sft.aae.mb_hi}{\ensuremath{-}0.018}  
\resdef{e4.l3.dpo_vs_sft.aae.mb_lo}{\ensuremath{-}0.033}  
\resdef{e4.l3.dpo_vs_sft.aus.mb}{\ensuremath{-}0.013}  
\resdef{e4.l3.dpo_vs_sft.aus.mb_hi}{\ensuremath{-}0.007}  
\resdef{e4.l3.dpo_vs_sft.aus.mb_lo}{\ensuremath{-}0.020}  
\resdef{e4.l3.dpo_vs_sft.bre.mb}{\ensuremath{-}0.014}  
\resdef{e4.l3.dpo_vs_sft.bre.mb_hi}{\ensuremath{-}0.008}  
\resdef{e4.l3.dpo_vs_sft.bre.mb_lo}{\ensuremath{-}0.021}  
\resdef{e4.l3.dpo_vs_sft.inf.mb}{\ensuremath{-}0.028}  
\resdef{e4.l3.dpo_vs_sft.inf.mb_hi}{\ensuremath{-}0.019}  
\resdef{e4.l3.dpo_vs_sft.inf.mb_lo}{\ensuremath{-}0.037}  
\resdef{e4.l3.dpo_vs_sft.sge.mb}{\ensuremath{-}0.025}  
\resdef{e4.l3.dpo_vs_sft.sge.mb_hi}{\ensuremath{-}0.012}  
\resdef{e4.l3.dpo_vs_sft.sge.mb_lo}{\ensuremath{-}0.037}  
\resdef{e4.l4.dpo_vs_sft.aae.mb}{\ensuremath{-}0.018}  
\resdef{e4.l4.dpo_vs_sft.aae.mb_hi}{\ensuremath{-}0.012}  
\resdef{e4.l4.dpo_vs_sft.aae.mb_lo}{\ensuremath{-}0.024}  
\resdef{e4.l4.dpo_vs_sft.aus.mb}{0.010}  
\resdef{e4.l4.dpo_vs_sft.aus.mb_hi}{0.015}  
\resdef{e4.l4.dpo_vs_sft.aus.mb_lo}{0.004}  
\resdef{e4.l4.dpo_vs_sft.bre.mb}{0.009}  
\resdef{e4.l4.dpo_vs_sft.bre.mb_hi}{0.015}  
\resdef{e4.l4.dpo_vs_sft.bre.mb_lo}{0.003}  
\resdef{e4.l4.dpo_vs_sft.inf.mb}{\ensuremath{-}0.007}  
\resdef{e4.l4.dpo_vs_sft.inf.mb_hi}{\ensuremath{-}0.001}  
\resdef{e4.l4.dpo_vs_sft.inf.mb_lo}{\ensuremath{-}0.013}  
\resdef{e4.l4.dpo_vs_sft.sge.mb}{\ensuremath{-}0.023}  
\resdef{e4.l4.dpo_vs_sft.sge.mb_hi}{\ensuremath{-}0.012}  
\resdef{e4.l4.dpo_vs_sft.sge.mb_lo}{\ensuremath{-}0.035}  
\resdef{e4.olmo2_13b.dpo_vs_sft.aae.mb}{\ensuremath{-}0.021}  
\resdef{e4.olmo2_13b.dpo_vs_sft.aae.mb_hi}{\ensuremath{-}0.016}  
\resdef{e4.olmo2_13b.dpo_vs_sft.aae.mb_lo}{\ensuremath{-}0.026}  
\resdef{e4.olmo2_13b.dpo_vs_sft.aus.mb}{0.000}  
\resdef{e4.olmo2_13b.dpo_vs_sft.aus.mb_hi}{0.006}  
\resdef{e4.olmo2_13b.dpo_vs_sft.aus.mb_lo}{\ensuremath{-}0.005}  
\resdef{e4.olmo2_13b.dpo_vs_sft.bre.mb}{\ensuremath{-}0.002}  
\resdef{e4.olmo2_13b.dpo_vs_sft.bre.mb_hi}{0.002}  
\resdef{e4.olmo2_13b.dpo_vs_sft.bre.mb_lo}{\ensuremath{-}0.007}  
\resdef{e4.olmo2_13b.dpo_vs_sft.inf.mb}{\ensuremath{-}0.013}  
\resdef{e4.olmo2_13b.dpo_vs_sft.inf.mb_hi}{\ensuremath{-}0.006}  
\resdef{e4.olmo2_13b.dpo_vs_sft.inf.mb_lo}{\ensuremath{-}0.018}  
\resdef{e4.olmo2_13b.dpo_vs_sft.sge.mb}{\ensuremath{-}0.015}  
\resdef{e4.olmo2_13b.dpo_vs_sft.sge.mb_hi}{\ensuremath{-}0.008}  
\resdef{e4.olmo2_13b.dpo_vs_sft.sge.mb_lo}{\ensuremath{-}0.023}  
\resdef{e4.olmo2_32b.dpo_vs_sft.aae.mb}{0.020}  
\resdef{e4.olmo2_32b.dpo_vs_sft.aae.mb_hi}{0.025}  
\resdef{e4.olmo2_32b.dpo_vs_sft.aae.mb_lo}{0.015}  
\resdef{e4.olmo2_32b.dpo_vs_sft.aus.mb}{0.017}  
\resdef{e4.olmo2_32b.dpo_vs_sft.aus.mb_hi}{0.023}  
\resdef{e4.olmo2_32b.dpo_vs_sft.aus.mb_lo}{0.012}  
\resdef{e4.olmo2_32b.dpo_vs_sft.bre.mb}{0.017}  
\resdef{e4.olmo2_32b.dpo_vs_sft.bre.mb_hi}{0.024}  
\resdef{e4.olmo2_32b.dpo_vs_sft.bre.mb_lo}{0.012}  
\resdef{e4.olmo2_32b.dpo_vs_sft.inf.mb}{0.012}  
\resdef{e4.olmo2_32b.dpo_vs_sft.inf.mb_hi}{0.016}  
\resdef{e4.olmo2_32b.dpo_vs_sft.inf.mb_lo}{0.007}  
\resdef{e4.olmo2_32b.dpo_vs_sft.sge.mb}{0.071}  
\resdef{e4.olmo2_32b.dpo_vs_sft.sge.mb_hi}{0.085}  
\resdef{e4.olmo2_32b.dpo_vs_sft.sge.mb_lo}{0.057}  
\resdef{e4b.mistral_7b.instruct_vs_base.aae.mb}{\ensuremath{-}0.006}  
\resdef{e4b.mistral_7b.instruct_vs_base.aae.mb_std}{0.026}  
\resdef{e4b.mistral_7b.instruct_vs_base.aus.mb}{0.017}  
\resdef{e4b.mistral_7b.instruct_vs_base.aus.mb_std}{0.036}  
\resdef{e4b.mistral_7b.instruct_vs_base.bre.mb}{0.018}  
\resdef{e4b.mistral_7b.instruct_vs_base.bre.mb_std}{0.039}  
\resdef{e4b.mistral_7b.instruct_vs_base.inf.mb}{\ensuremath{-}0.024}  
\resdef{e4b.mistral_7b.instruct_vs_base.inf.mb_std}{0.035}  
\resdef{e4b.mistral_7b.instruct_vs_base.sge.mb}{0.009}  
\resdef{e4b.mistral_7b.instruct_vs_base.sge.mb_std}{0.057}  
\resdef{e4b.qwen3_8b.instruct_vs_base.aae.mb}{0.015}  
\resdef{e4b.qwen3_8b.instruct_vs_base.aae.mb_std}{0.034}  
\resdef{e4b.qwen3_8b.instruct_vs_base.aus.mb}{0.029}  
\resdef{e4b.qwen3_8b.instruct_vs_base.aus.mb_std}{0.042}  
\resdef{e4b.qwen3_8b.instruct_vs_base.bre.mb}{0.030}  
\resdef{e4b.qwen3_8b.instruct_vs_base.bre.mb_std}{0.041}  
\resdef{e4b.qwen3_8b.instruct_vs_base.inf.mb}{\ensuremath{-}0.010}  
\resdef{e4b.qwen3_8b.instruct_vs_base.inf.mb_std}{0.032}  
\resdef{e4b.qwen3_8b.instruct_vs_base.sge.mb}{0.005}  
\resdef{e4b.qwen3_8b.instruct_vs_base.sge.mb_std}{0.059}  
\resdef{e4i.dpo_vs_sft.adjusted.coef}{0.020}  
\resdef{e4i.dpo_vs_sft.adjusted.hi}{0.027}  
\resdef{e4i.dpo_vs_sft.adjusted.lo}{0.014}  
\resdef{e4i.dpo_vs_sft.per_lineage.olmo2_32b.raw}{\ensuremath{-}0.031}  
\resdef{e4i.dpo_vs_sft.per_lineage.olmo31_32b.raw}{0.030}  
\resdef{e4i.dpo_vs_sft.raw_hi}{0.020}  
\resdef{e4i.dpo_vs_sft.raw_lo}{0.000}  
\resdef{e4i.dpo_vs_sft.raw_mean_over_lineages}{0.011}  
\resdef{e4i.sft_vs_base.per_lineage.l1.raw}{0.035}  
\resdef{e4i.sft_vs_base.per_lineage.l2.raw}{0.032}  
\resdef{e4i.sft_vs_base.per_lineage.olmo31_32b.raw}{0.026}  
\resdef{e4i.sft_vs_base.per_lineage.zephyr_7b.raw}{0.001}  
\resdef{e4m.dpo_vs_sft.pooled.tier}{0.006}  
\resdef{e4m.dpo_vs_sft.pooled.tier_hi}{0.041}  
\resdef{e4m.dpo_vs_sft.pooled.tier_lo}{\ensuremath{-}0.030}  
\resdef{dv.auth.k_both}{1}  
\resdef{dv.auth.n_auth_endive_aae}{500}  
\resdef{dv.auth.n_auth_endive_sge}{500}  
\resdef{dv.auth.n_auth_gro}{225}  
\resdef{dv.auth.n_lin}{4}  
\resdef{dv.auth.n_lin_all}{5}  
\resdef{dv.auth.post_max}{0.017}  
\resdef{dv.auth.ref_dia_dpo}{12}  
\resdef{dv.auth.ref_pair_dpo}{22}  
\resdef{dv.auth.ref_sae_dpo}{17}  
\resdef{dv.cap.max_abs_d_used}{0.006}  
\resdef{dv.cap.n_cells}{6}  
\resdef{dv.cap.trunc1024_max}{0.19}  
\resdef{dv.cap.trunc400_max}{0.79}  
\resdef{dv.cap.trunc400_min}{0.67}  
\resdef{dv.capabl.cap_ins}{0.001}  
\resdef{dv.capabl.cap_sge_ratio}{0.79}  
\resdef{dv.capabl.cap_sge_tail}{0}  
\resdef{dv.capabl.msft_crheld}{\ensuremath{-}0.001}  
\resdef{dv.capabl.msft_ifeval}{\ensuremath{-}5.9}  
\resdef{dv.capabl.mx_crheld}{0.201}  
\resdef{dv.capabl.mx_crheld_hi}{0.269}  
\resdef{dv.capabl.mx_crheld_lo}{0.139}  
\resdef{dv.capabl.mx_ins}{0.378}  
\resdef{dv.capabl.mx_n}{2}  
\resdef{dv.capabl.mx_ostd}{2.42}  
\resdef{dv.capabl.mx_sge_ratio}{10.41}  
\resdef{dv.capabl.mx_sge_tail}{95}  
\resdef{dv.capabl.noA_crheld}{0.008}  
\resdef{dv.capabl.noA_n}{2}  
\resdef{dv.capabl.noO_crheld}{0.583}  
\resdef{dv.capabl.noO_ins}{0.039}  
\resdef{dv.capabl.noO_n}{2}  
\resdef{dv.capabl.noO_ostd}{7.36}  
\resdef{dv.capabl.unm_crheld}{0.043}  
\resdef{dv.capabl.unm_ins}{0.560}  
\resdef{dv.capabl.x15_crheld}{0.340}  
\resdef{dv.capabl.x15_fold}{2.0}  
\resdef{dv.capabl.x15_ins}{\ensuremath{-}0.007}  
\resdef{dv.capres.b1.cap_crg}{0.150}  
\resdef{dv.capres.b1.cap_crheld}{0.173}  
\resdef{dv.capres.b1.cap_crheld_hi}{0.207}  
\resdef{dv.capres.b1.cap_crheld_lo}{0.138}  
\resdef{dv.capres.b1.cap_ifeval}{2.1}  
\resdef{dv.capres.b1.cap_ifeval_hi}{4.9}  
\resdef{dv.capres.b1.cap_ifeval_lo}{\ensuremath{-}0.7}  
\resdef{dv.capres.b1.cap_insertion}{0.001}  
\resdef{dv.capres.b1.cap_insertion_hi}{0.013}  
\resdef{dv.capres.b1.cap_insertion_lo}{\ensuremath{-}0.008}  
\resdef{dv.capres.b1.cap_ostd}{0.680}  
\resdef{dv.capres.b1.cap_ratio_aae}{0.17}  
\resdef{dv.capres.b1.cap_ratio_sge}{0.79}  
\resdef{dv.capres.b1.cap_track}{0.191}  
\resdef{dv.capres.b1.fold}{4.0}  
\resdef{dv.capres.b1.unm_crg}{0.032}  
\resdef{dv.capres.b1.unm_crheld}{0.043}  
\resdef{dv.capres.b1.unm_insertion}{0.560}  
\resdef{dv.capres.b1.unm_ostd}{0.954}  
\resdef{dv.capres.b1.unm_ratio_aae}{0.18}  
\resdef{dv.capres.b1.unm_ratio_sge}{2.59}  
\resdef{dv.capres.b1.unm_track}{3.606}  
\resdef{dv.capres.q3.cap_crheld}{0.017}  
\resdef{dv.capres.q3.cap_crheld_hi}{0.026}  
\resdef{dv.capres.q3.cap_crheld_lo}{0.009}  
\resdef{dv.capres.q3.unm_crheld}{0.154}  
\resdef{dv.capsens.noexcl}{0.173}  
\resdef{dv.capsens.req_hi}{0.198}  
\resdef{dv.capsens.req_lo}{0.151}  
\resdef{dv.capsens.v1}{0.137}  
\resdef{dv.capsens.v1_hi}{0.167}  
\resdef{dv.capsens.v1_lo}{0.108}  
\resdef{dv.comp_corr.l2_aae.comp}{0.567}  
\resdef{dv.comp_corr.l2_sge.comp}{0.288}  
\resdef{dv.comp_corr.l3_aae.comp}{0.411}  
\resdef{dv.comp_corr.l3_sge.comp}{0.391}  
\resdef{dv.comp_corr.l4_aae.comp}{0.588}  
\resdef{dv.comp_corr.l4_sge.comp}{0.251}  
\resdef{dv.dep.aae_max}{0.009}  
\resdef{dv.dep.bre_max}{0.207}  
\resdef{dv.dep.bre_min}{0.089}  
\resdef{dv.dep.inf_max}{0.053}  
\resdef{dv.dep.n}{5}  
\resdef{dv.dep.ngp_max}{0.502}  
\resdef{dv.dep.ngp_min}{0.116}  
\resdef{dv.dep.ngp_open_max}{0.046}  
\resdef{dv.dep.query_date}{2026-10-02}  
\resdef{dv.dfr.n_base_cells}{20}  
\resdef{dv.dfr.n_base_div_pos}{20}  
\resdef{dv.dfr.n_cells_dpo}{24}  
\resdef{dv.dfr.n_cells_sft}{24}  
\resdef{dv.dfr.n_div_dpo}{0}  
\resdef{dv.dfr.n_div_sft}{2}  
\resdef{dv.dfr.n_opp_dpo}{19}  
\resdef{dv.dfr.n_opp_sft}{4}  
\resdef{dv.e560.eff_s0}{0.090}  
\resdef{dv.e560.eff_s1}{0.023}  
\resdef{dv.e560.eff_s2}{0.030}  
\resdef{dv.e560.eff_s3}{0.027}  
\resdef{dv.e560.ftrain_s0}{4.1}  
\resdef{dv.e560.ftrain_s1}{0.6}  
\resdef{dv.e560.ftrain_s2}{1.0}  
\resdef{dv.e560.ftrain_s3}{0.7}  
\resdef{dv.e560.hi_s0}{0.107}  
\resdef{dv.e560.hi_s1}{0.049}  
\resdef{dv.e560.hi_s2}{0.039}  
\resdef{dv.e560.hi_s3}{0.052}  
\resdef{dv.e560.lo_s0}{0.073}  
\resdef{dv.e560.lo_s1}{0.008}  
\resdef{dv.e560.lo_s2}{0.022}  
\resdef{dv.e560.lo_s3}{0.012}  
\resdef{dv.e560.n}{4}  
\resdef{dv.e5dose.share_1k}{0.8}  
\resdef{dv.e5dose.share_20k}{14}  
\resdef{dv.e5dose.share_60k}{33}  
\resdef{dv.fam.aae_final_max}{0.001}  
\resdef{dv.fam.bre_falls}{9}  
\resdef{dv.fam.bre_rise_names}{Falcon3-7B, Yi-1.5-9B and DeepSeek-LLM-7B}  
\resdef{dv.fam.n}{12}  
\resdef{dv.gate.cr_diff}{0.38}  
\resdef{dv.gate.cr_hi}{0.41}  
\resdef{dv.gate.cr_lo}{0.35}  
\resdef{dv.gate.ins_diff}{\ensuremath{-}0.002}  
\resdef{dv.gate.n_checks}{4}  
\resdef{dv.gate.n_pass}{3}  
\resdef{dv.gate.ratio_aae}{0.50}  
\resdef{dv.gate.ratio_sge}{0.26}  
\resdef{dv.gpu.dpo}{185}  
\resdef{dv.gpu.total}{298}  
\resdef{dv.h3.l2_aae.k_ret}{4}  
\resdef{dv.h3.l2_aae.k_ret_corr}{3}  
\resdef{dv.h3.l2_aae.n_ret}{5}  
\resdef{dv.h3.l3_aae.k_ret}{2}  
\resdef{dv.h3.l3_aae.k_ret_corr}{2}  
\resdef{dv.h3.l3_sge.k_ret}{1}  
\resdef{dv.h3.l3_sge.n_ret}{2}  
\resdef{dv.h3.l4_aae.k_ret}{3}  
\resdef{dv.h3.l4_aae.k_ret_corr}{3}  
\resdef{dv.ifeval.c60k.strict}{58.2}  
\resdef{dv.ifeval.d0.strict}{59.1}  
\resdef{dv.ifeval.drop60}{1.2}  
\resdef{dv.ifeval.maxdrop60_run}{2.6}  
\resdef{dv.ifeval.t20k.strict}{60.1}  
\resdef{dv.ifeval.t60k.n}{4}  
\resdef{dv.ifeval.t60k.strict}{57.9}  
\resdef{dv.ifeval.t60k.strict_s0}{58.6}  
\resdef{dv.ifeval.t60k.strict_s1}{57.7}  
\resdef{dv.ifeval.t60k.strict_s2}{56.6}  
\resdef{dv.ifeval.t60k.strict_s3}{59.0}  
\resdef{dv.n_bre_sft_falls_allframes}{3}  
\resdef{dv.n_df_cells_sft}{24}  
\resdef{dv.n_df_div_sft}{9}  
\resdef{dv.n_inf_sft_falls_allframes}{3}  
\resdef{dv.n_int}{10}  
\resdef{dv.n_int_pos}{9}  
\resdef{dv.n_lineages}{10}  
\resdef{dv.n_lineages_allframes}{6}  
\resdef{dv.n_lineages_base}{7}  
\resdef{dv.n_mb_aae_neg}{9}  
\resdef{dv.n_mb_inf_neg}{6}  
\resdef{dv.n_mb_prestige_spared}{9}  
\resdef{dv.n_mb_sge_neg}{6}  
\resdef{dv.n_mixed_tier_pos}{9}  
\resdef{dv.n_mixed_tier_sig}{2}  
\resdef{dv.n_sft_larger_prereg}{7}  
\resdef{dv.n_sft_larger_rate}{5}  
\resdef{dv.n_sge_neg_bre_nonneg}{7}  
\resdef{dv.n_split_falls}{10}  
\resdef{dv.n_split_rate_falls}{10}  
\resdef{dv.opg.chat.zh}{45}  
\resdef{dv.opg.chat_enprompt.zh}{0}  
\resdef{dv.opg.chat_nosys.trunc}{76}  
\resdef{dv.pa.fluency_o}{0.18}  
\resdef{dv.pa.undershoot}{1.9}  
\resdef{dv.pa.within25}{0.39}  
\resdef{dv.pref.aae_morph_7b}{0.65}  
\resdef{dv.pref.floor}{0.03}  
\resdef{dv.pref.olmo2_1b}{0.81}  
\resdef{dv.pref.olmo2_32b}{0.81}  
\resdef{dv.pref.olmo2_7b}{0.81}  
\resdef{dv.pref.tulu3}{0.64}  
\resdef{dv.prefwin.32b.diff}{\ensuremath{-}0.006}  
\resdef{dv.prefwin.32b.hi}{0.009}  
\resdef{dv.prefwin.32b.lo}{\ensuremath{-}0.024}  
\resdef{dv.prefwin.32b.tie}{87}  
\resdef{dv.prefwin.7b.diff}{\ensuremath{-}0.007}  
\resdef{dv.prefwin.7b.hi}{0.005}  
\resdef{dv.prefwin.7b.lo}{\ensuremath{-}0.022}  
\resdef{dv.prefwin.7b.tie}{87}  
\resdef{dv.s5.max_abs_d}{0.006}  
\resdef{dv.s5.n_lin}{4}  
\resdef{dv.syn.aae_copula_r}{0.23}  
\resdef{dv.syn.aae_uninflect_r}{0.52}  
\resdef{dv.syn.fpr_max}{0.9}  
\resdef{dv.syn.n_counted}{13}  
\resdef{dv.syn.n_pool}{8{,}000}  
\resdef{dv.syn.p_min}{0.99}  
\resdef{dv.syn.r_max}{1.00}  
\resdef{dv.syn.r_min}{0.81}  
\resdef{dv.txt.adj}{all 10}  
\resdef{dv.txt.bre_dpo}{9 of 10}  
\resdef{dv.txt.capcrit_b1}{M1 passes, M2 fails, M3 fails, M4 passes, M5 fails}  
\resdef{dv.txt.inf_dpo}{all 10}  
\resdef{dv.txt.s5_bre}{all 4}  
\resdef{dv.txt.s5_inf}{all 4}  
\resdef{dv.txt.strat}{all 10}  
\resdef{dv.wild.aae}{0.50}  
\resdef{dv.wild.n}{162{,}842}  
\resdef{dv.wild.sge}{0.04}  
\resdef{e2.l2.dpo.persona.aae.cr}{0.008}  
\resdef{e2.l2.dpo.persona.bre.cr_rate}{0.136}  
\resdef{e2.l2.dpo.persona.sge.cr_rate}{0.010}  
\resdef{e2.l2.dpo.style.aae.cr}{0.003}  
\resdef{e2.l2.dpo.style.bre.cr_rate}{0.104}  
\resdef{e2.l2.dpo.style.sge.cr_rate}{0.005}  
\resdef{e3.l2.base.aae.blevins_conv_novel.delta}{\ensuremath{-}0.001}  
\resdef{e3.l2.base.aae.blevins_conv_novel.mark}{}  
\resdef{e3.l2.base.aae.copy_rate.delta}{0.003}  
\resdef{e3.l2.base.aae.copy_rate.mark}{}  
\resdef{e3.l2.base.aae.lsm_novel.delta}{\ensuremath{-}0.001}  
\resdef{e3.l2.base.aae.lsm_novel.mark}{}  
\resdef{e3.l2.base.inf.blevins_conv_novel.delta}{0.006}  
\resdef{e3.l2.base.inf.blevins_conv_novel.mark}{\sig{*}}  
\resdef{e3.l2.base.inf.copy_rate.delta}{0.006}  
\resdef{e3.l2.base.inf.copy_rate.mark}{\sig{*}}  
\resdef{e3.l2.base.inf.lsm_novel.delta}{0.002}  
\resdef{e3.l2.base.inf.lsm_novel.mark}{}  
\resdef{e3.l2.base.sge.blevins_conv_novel.delta}{0.005}  
\resdef{e3.l2.base.sge.blevins_conv_novel.mark}{}  
\resdef{e3.l2.base.sge.copy_rate.delta}{\ensuremath{-}0.009}  
\resdef{e3.l2.base.sge.copy_rate.mark}{\sig{*}}  
\resdef{e3.l2.base.sge.lsm_novel.delta}{\ensuremath{-}0.028}  
\resdef{e3.l2.base.sge.lsm_novel.mark}{\sig{*}}  
\resdef{e3.l2.dpo.aae.blevins_conv_novel.delta}{\ensuremath{-}0.002}  
\resdef{e3.l2.dpo.aae.blevins_conv_novel.mark}{\sig{*}}  
\resdef{e3.l2.dpo.aae.copy_rate.delta}{0.000}  
\resdef{e3.l2.dpo.aae.copy_rate.mark}{}  
\resdef{e3.l2.dpo.aae.lsm_novel.delta}{\ensuremath{-}0.002}  
\resdef{e3.l2.dpo.aae.lsm_novel.mark}{}  
\resdef{e3.l2.dpo.inf.blevins_conv_novel.delta}{\ensuremath{-}0.002}  
\resdef{e3.l2.dpo.inf.blevins_conv_novel.mark}{}  
\resdef{e3.l2.dpo.inf.copy_rate.delta}{0.000}  
\resdef{e3.l2.dpo.inf.copy_rate.mark}{}  
\resdef{e3.l2.dpo.inf.lsm_novel.delta}{0.003}  
\resdef{e3.l2.dpo.inf.lsm_novel.mark}{}  
\resdef{e3.l2.dpo.sge.blevins_conv_novel.delta}{\ensuremath{-}0.009}  
\resdef{e3.l2.dpo.sge.blevins_conv_novel.mark}{\sig{*}}  
\resdef{e3.l2.dpo.sge.copy_rate.delta}{\ensuremath{-}0.005}  
\resdef{e3.l2.dpo.sge.copy_rate.mark}{\sig{*}}  
\resdef{e3.l2.dpo.sge.lsm_novel.delta}{\ensuremath{-}0.033}  
\resdef{e3.l2.dpo.sge.lsm_novel.mark}{\sig{*}}  
\resdef{fp.max.AAE_ain_t}{0.001}  
\resdef{fp.max.AAE_completive_done}{0.002}  
\resdef{fp.max.AAE_copula_absence}{0.513}  
\resdef{fp.max.AAE_finna_gon}{0.001}  
\resdef{fp.max.AAE_g_drop}{0.002}  
\resdef{fp.max.AAE_habitual_be}{0.006}  
\resdef{fp.max.AAE_lexicon}{0.006}  
\resdef{fp.max.AAE_negative_concord}{0.009}  
\resdef{fp.max.AAE_uninflect}{0.592}  
\resdef{fp.max.AAE_yall}{0.000}  
\resdef{fp.max.APP_a_prefix}{0.001}  
\resdef{fp.max.APP_ain_t}{0.001}  
\resdef{fp.max.APP_double_modals}{0.000}  
\resdef{fp.max.APP_fixin_to}{0.000}  
\resdef{fp.max.APP_negative_concord}{0.009}  
\resdef{fp.max.APP_yall}{0.000}  
\resdef{fp.max.AUS_aus_lexis}{0.021}  
\resdef{fp.max.AUS_spelling}{0.011}  
\resdef{fp.max.BRE_have_got}{0.007}  
\resdef{fp.max.BRE_lexis}{0.006}  
\resdef{fp.max.BRE_spelling}{0.011}  
\resdef{fp.max.NGP_dey}{0.000}  
\resdef{fp.max.NGP_don}{0.000}  
\resdef{fp.max.NGP_fit}{0.001}  
\resdef{fp.max.NGP_lexicon}{0.004}  
\resdef{fp.max.NGP_na}{0.001}  
\resdef{fp.max.NGP_no}{0.000}  
\resdef{fp.max.NGP_wan}{0.000}  
\resdef{fp.max.SGE_can_cannot}{0.000}  
\resdef{fp.max.SGE_copula_absence}{0.513}  
\resdef{fp.max.SGE_finish_completive}{0.005}  
\resdef{fp.max.SGE_lexicon}{0.002}  
\resdef{fp.max.SGE_particles}{0.001}  
\resdef{fp.max.SGE_zero_plural}{0.036}  
\resdef{fp.n_audited}{39}  
\resdef{fp.n_per_model}{1{,}722}  
\resdef{e4a.breadth.qwen35_9b_instruct_vs_base.adj}{0.010}  
\resdef{e4a.breadth_adj_neg}{0}  
\resdef{e4a.breadth_adj_pos}{12}  
\resdef{e4a.breadth_inconcl_names}{SmolLM3's preference stage and Qwen3.5-4B}  
\resdef{e4a.breadth_n}{14}  
\resdef{e4a.max_abs_ll_slope_others}{0.014}  
\resdef{e4a.n_adj_pos}{10}  
\resdef{e4a.n_lin}{10}  
\resdef{e4a.olmo2_32b.adj}{0.012}  
\resdef{e4a.olmo2_32b.adj_hi}{0.022}  
\resdef{e4a.olmo2_32b.adj_lo}{0.002}  
\resdef{e4a.olmo2_32b.ll_slope}{\ensuremath{-}0.034}  
\resdef{e4a.olmo31_32b.ll_slope}{\ensuremath{-}0.014}  
\resdef{e4a.sftbase.zephyr_7b.adj}{0.010}  
\resdef{e4a.strat.max_width_ratio}{1.02}  
\resdef{e4a.strat.min_width_ratio}{0.93}  
\resdef{e5c.controls_unsolicited_max}{0.00}  
\resdef{e5c.t60k.aae_ratio_max}{0.47}  
\resdef{e5c.t60k.aae_ratio_min}{0.17}  
\resdef{e5c.t60k.n_runs}{4}  
\resdef{e5c.t60k.sge_ratio_max}{9.47}  
\resdef{e5c.t60k.sge_ratio_min}{1.07}  
\resdef{e5c.t60k.unsolicited_max}{1.06}  
\resdef{e5c.t60k.unsolicited_min}{0.10}  
\resdef{audit.cap.cap.n}{20}  
\resdef{audit.cap.cap.n_exaggerated}{0}  
\resdef{audit.cap.n_items}{40}  
\resdef{audit.cap.unmatched.n}{20}  
\resdef{audit.cap.unmatched.n_exaggerated}{0}  
\resdef{e5t.crg.effect_60k_T_minus_C.mean}{0.035}  
\resdef{e5t.crheld.effect_20k_T_minus_C.hi}{0.004}  
\resdef{e5t.crheld.effect_20k_T_minus_C.lo}{\ensuremath{-}0.002}  
\resdef{e5t.crheld.effect_20k_T_minus_C.mean}{0.001}  
\resdef{e5t.crheld.effect_5k_T_minus_C.mean}{0.000}  
\resdef{e5t.crheld.effect_60k_T_minus_C.hi}{0.058}  
\resdef{e5t.crheld.effect_60k_T_minus_C.lo}{0.032}  
\resdef{e5t.crheld.effect_60k_T_minus_C.mean}{0.043}  
\resdef{e5t.crheld.interaction_slope4_T_minus_C_per10k.mean}{0.007}  
\resdef{e5t.crheld.interaction_slope_T_minus_C_per10k.mean}{0.001}  
\resdef{e5t.crheld.interaction_slope_T_minus_C_per10k.p_one_sided}{0.197}  
\resdef{e5t.ftrain.effect_20k_T_minus_C.mean}{0.015}  
\resdef{e5t.split.exploratory_20k_C_minus_d0.mean}{0.024}  
\resdef{e5t.split.interaction_slope_T_minus_C_per10k.mean}{\ensuremath{-}0.016}  
\resdef{e5.aa1k.crg}{0.003}  
\resdef{e5.aa1k.crg_std}{0.000}  
\resdef{e5.aa1k.crheld}{0.001}  
\resdef{e5.aa1k.crheld_std}{0.001}  
\resdef{e5.aa1k.ftrain}{0.000}  
\resdef{e5.aa1k.ftrain_std}{0.000}  
\resdef{e5.aa1k.n_seeds}{2}  
\resdef{e5.aa1k.split}{0.032}  
\resdef{e5.aa1k.split_std}{0.003}  
\resdef{e5.aa20k.crg}{0.006}  
\resdef{e5.aa20k.crg_std}{0.001}  
\resdef{e5.aa20k.crheld}{0.005}  
\resdef{e5.aa20k.crheld_std}{0.000}  
\resdef{e5.aa20k.ftrain}{0.015}  
\resdef{e5.aa20k.ftrain_std}{0.004}  
\resdef{e5.aa20k.n_seeds}{4}  
\resdef{e5.aa20k.split}{0.028}  
\resdef{e5.aa20k.split_std}{0.002}  
\resdef{e5.aa5k.crg}{0.004}  
\resdef{e5.aa5k.crg_std}{0.000}  
\resdef{e5.aa5k.crheld}{0.001}  
\resdef{e5.aa5k.crheld_std}{0.001}  
\resdef{e5.aa5k.ftrain}{0.000}  
\resdef{e5.aa5k.ftrain_std}{0.000}  
\resdef{e5.aa5k.n_seeds}{4}  
\resdef{e5.aa5k.split}{0.032}  
\resdef{e5.aa5k.split_std}{0.002}  
\resdef{e5.aa60k.crg}{0.040}  
\resdef{e5.aa60k.crg_std}{0.036}  
\resdef{e5.aa60k.crheld}{0.047}  
\resdef{e5.aa60k.crheld_std}{0.032}  
\resdef{e5.aa60k.ftrain}{1.595}  
\resdef{e5.aa60k.ftrain_std}{1.683}  
\resdef{e5.aa60k.n_seeds}{4}  
\resdef{e5.aa60k.split}{\ensuremath{-}0.136}  
\resdef{e5.aa60k.split_std}{0.162}  
\resdef{e5.anti1k.crg}{0.003}  
\resdef{e5.anti1k.crg_std}{0.000}  
\resdef{e5.anti1k.crheld}{0.000}  
\resdef{e5.anti1k.crheld_std}{0.001}  
\resdef{e5.anti1k.ftrain}{0.000}  
\resdef{e5.anti1k.ftrain_std}{0.000}  
\resdef{e5.anti1k.n_seeds}{2}  
\resdef{e5.anti1k.split}{0.032}  
\resdef{e5.anti1k.split_std}{0.001}  
\resdef{e5.anti5k.crg}{0.003}  
\resdef{e5.anti5k.crg_std}{0.000}  
\resdef{e5.anti5k.crheld}{0.000}  
\resdef{e5.anti5k.crheld_std}{0.000}  
\resdef{e5.anti5k.ftrain}{0.000}  
\resdef{e5.anti5k.ftrain_std}{0.000}  
\resdef{e5.anti5k.n_seeds}{2}  
\resdef{e5.anti5k.split}{0.035}  
\resdef{e5.anti5k.split_std}{0.002}  
\resdef{e5.ctl20k.crg}{0.005}  
\resdef{e5.ctl20k.crg_std}{0.001}  
\resdef{e5.ctl20k.crheld}{0.003}  
\resdef{e5.ctl20k.crheld_std}{0.001}  
\resdef{e5.ctl20k.ftrain}{0.000}  
\resdef{e5.ctl20k.ftrain_std}{0.000}  
\resdef{e5.ctl20k.n_seeds}{3}  
\resdef{e5.ctl20k.split}{0.058}  
\resdef{e5.ctl20k.split_std}{0.008}  
\resdef{e5.ctl5k.crg}{0.003}  
\resdef{e5.ctl5k.crg_std}{0.000}  
\resdef{e5.ctl5k.crheld}{0.001}  
\resdef{e5.ctl5k.crheld_std}{0.000}  
\resdef{e5.ctl5k.ftrain}{0.000}  
\resdef{e5.ctl5k.ftrain_std}{0.000}  
\resdef{e5.ctl5k.n_seeds}{4}  
\resdef{e5.ctl5k.split}{0.033}  
\resdef{e5.ctl5k.split_std}{0.003}  
\resdef{e5.ctl60k.crg}{0.005}  
\resdef{e5.ctl60k.crg_std}{0.001}  
\resdef{e5.ctl60k.crheld}{0.005}  
\resdef{e5.ctl60k.crheld_std}{0.001}  
\resdef{e5.ctl60k.ftrain}{0.000}  
\resdef{e5.ctl60k.ftrain_std}{0.000}  
\resdef{e5.ctl60k.n_seeds}{4}  
\resdef{e5.ctl60k.split}{0.076}  
\resdef{e5.ctl60k.split_std}{0.002}  
\resdef{e5.d0.crg}{0.003}  
\resdef{e5.d0.crg_std}{0.001}  
\resdef{e5.d0.crheld}{0.001}  
\resdef{e5.d0.crheld_std}{0.000}  
\resdef{e5.d0.ftrain}{0.000}  
\resdef{e5.d0.ftrain_std}{0.000}  
\resdef{e5.d0.n_seeds}{3}  
\resdef{e5.d0.split}{0.033}  
\resdef{e5.d0.split_std}{0.001}  
\resdef{audit.v4.a.full.ac1}{\ensuremath{-}0.062}  
\resdef{audit.v4.a.full.hi}{0.699}  
\resdef{audit.v4.a.full.k}{7}  
\resdef{audit.v4.a.full.kappa}{\ensuremath{-}0.053}  
\resdef{audit.v4.a.full.lo}{0.248}  
\resdef{audit.v4.a.full.n}{15}  
\resdef{audit.v4.a.full.rate}{0.467}  
\resdef{audit.v4.a.n_same}{25}  
\resdef{audit.v4.a.trimmed.ac1}{0.204}  
\resdef{audit.v4.a.trimmed.hi}{0.802}  
\resdef{audit.v4.a.trimmed.k}{9}  
\resdef{audit.v4.a.trimmed.kappa}{0.196}  
\resdef{audit.v4.a.trimmed.kappa_hi}{0.700}  
\resdef{audit.v4.a.trimmed.kappa_lo}{\ensuremath{-}0.316}  
\resdef{audit.v4.a.trimmed.lo}{0.357}  
\resdef{audit.v4.a.trimmed.n}{15}  
\resdef{audit.v4.a.trimmed.rate}{0.600}  
\resdef{audit.v4.b.primary.agreement.ac1}{0.896}  
\resdef{audit.v4.b.primary.agreement.ac1_hi}{1.000}  
\resdef{audit.v4.b.primary.agreement.ac1_lo}{0.719}  
\resdef{audit.v4.b.primary.agreement.hi}{0.982}  
\resdef{audit.v4.b.primary.agreement.k}{28}  
\resdef{audit.v4.b.primary.agreement.kappa}{0.814}  
\resdef{audit.v4.b.primary.agreement.kappa_hi}{1.000}  
\resdef{audit.v4.b.primary.agreement.kappa_lo}{0.510}  
\resdef{audit.v4.b.primary.agreement.lo}{0.787}  
\resdef{audit.v4.b.primary.agreement.n}{30}  
\resdef{audit.v4.b.primary.agreement.rate}{0.933}  
\resdef{audit.v4.b.primary.miss_on_neg.k}{1}  
\resdef{audit.v4.b.primary.miss_on_neg.n}{23}  
\resdef{audit.v4.b.primary.precision.hi}{0.974}  
\resdef{audit.v4.b.primary.precision.k}{6}  
\resdef{audit.v4.b.primary.precision.lo}{0.487}  
\resdef{audit.v4.b.primary.precision.n}{7}  
\resdef{audit.v4.b.primary.precision.rate}{0.857}  
\resdef{audit.v4.b.retained_only.precision.hi}{1.000}  
\resdef{audit.v4.b.retained_only.precision.lo}{0.610}  
\resdef{audit.v4.b.retained_only.precision.n}{6}  
\resdef{audit.v4.b.retained_only.precision.rate}{1.000}  
\resdef{audit.v4.c.by_feature.ain_t.n}{4}  
\resdef{audit.v4.c.by_feature.completive_done.n}{2}  
\resdef{audit.v4.c.by_feature.finish_completive.n}{4}  
\resdef{audit.v4.c.by_feature.negative_concord.n}{4}  
\resdef{audit.v4.c.by_feature.negative_concord.precision.k}{0}  
\resdef{audit.v4.c.by_feature.negative_concord.precision.n}{4}  
\resdef{audit.v4.c.primary.precision.hi}{0.674}  
\resdef{audit.v4.c.primary.precision.k}{6}  
\resdef{audit.v4.c.primary.precision.lo}{0.214}  
\resdef{audit.v4.c.primary.precision.n}{14}  
\resdef{audit.v4.c.primary.precision.rate}{0.429}  
\resdef{audit.v4.n_a}{40}  
\resdef{audit.v4.n_b}{30}  
\resdef{audit.v4.n_c}{14}  
\resdef{audit.v4.n_items}{84}  
\resdef{audit.j.gptoss_qwen.a.k}{30}  
\resdef{audit.j.gptoss_qwen.bc.k}{33}  
\resdef{audit.j.h_gptoss.a.n_same_x}{25}  
\resdef{audit.j.h_gptoss.a.n_same_y}{3}  
\resdef{audit.j.h_gptoss.a_dir.k}{14}  
\resdef{audit.j.h_gptoss.a_dir.n}{15}  
\resdef{audit.j.h_gptoss.bc.k}{43}  
\resdef{audit.j.h_gptoss.bc.kappa}{0.95}  
\resdef{audit.j.h_qwen.a.n_same_y}{0}  
\resdef{audit.j.h_qwen.a_dir.k}{12}  
\resdef{audit.j.h_qwen.a_dir.n}{15}  
\resdef{audit.j.h_qwen.bc.k}{34}  
\resdef{audit.j.h_qwen.bc.kappa}{0.43}  
\resdef{audit.j.n_a}{40}  
\resdef{audit.j.n_bc}{44}  
\resdef{audit.r.full.k}{12}  
\resdef{audit.r.trimmed.ac1}{0.63}  
\resdef{audit.r.trimmed.hi}{0.93}  
\resdef{audit.r.trimmed.k}{12}  
\resdef{audit.r.trimmed.kappa}{0.59}  
\resdef{audit.r.trimmed.lo}{0.55}  
\resdef{audit.r.trimmed.n}{15}  
\resdef{audit.r.trimmed.rate}{0.80}  
\resdef{audit.auth.aaen.hi}{0.503}  
\resdef{audit.auth.aaen.k}{3}  
\resdef{audit.auth.aaen.lo}{0.082}  
\resdef{audit.auth.aaen.n}{13}  
\resdef{audit.auth.aaen.rate}{0.231}  
\resdef{audit.auth.by_feature.aae_ain_t.k}{2}  
\resdef{audit.auth.by_feature.aae_ain_t.n}{2}  
\resdef{audit.auth.by_feature.aae_negative_concord.k}{1}  
\resdef{audit.auth.by_feature.aae_negative_concord.n}{11}  
\resdef{audit.auth.ctrl.n}{2}  
\resdef{audit.auth.flagged.k}{3}  
\resdef{audit.auth.flagged.n}{18}  
\resdef{audit.auth.n_items}{20}  
\resdef{audit.auth.sgen.hi}{0.434}  
\resdef{audit.auth.sgen.k}{0}  
\resdef{audit.auth.sgen.lo}{0.000}  
\resdef{audit.auth.sgen.n}{5}  
\resdef{audit.auth.sgen.rate}{0.000}  
\resdef{audit.auth.wild_aae_corr.est}{0.12}  
\resdef{audit.auth.wild_aae_corr.hi}{0.25}  
\resdef{audit.auth.wild_aae_corr.lo}{0.04}  
\resdef{rh.aligned.aae.aae_hits}{146}  
\resdef{rh.aligned.aae.nc_pct}{88}  
\resdef{rh.aligned.n_replies}{30{,}996}  
\resdef{rh.aligned.sae.aae_hits}{121}  
\resdef{rh.aligned.sae.sge_particles}{2}  
\resdef{rh.aligned.sge.sge_particles}{368}  

\draftfalse

\title{Answering in the User's Variety: \\ Calibrated Accommodation Pairs for Post-Training}

\author{
\bfseries Zhizhuo Kou\textsuperscript{1} \quad
Heming Li\textsuperscript{2} \quad Jian Yang \textsuperscript{1} \quad Junyu Luo \textsuperscript{3} \\
\bfseries Jingshu Peng \textsuperscript{1} \quad
Sirui Han\textsuperscript{1}\NoHyper\footnotemark[1]\endNoHyper \quad Yike Guo\textsuperscript{1} \\[2pt]
\normalfont\textsuperscript{1}The Hong Kong University of Science and Technology \\
\normalfont\textsuperscript{2}The Hong Kong University of Science and Technology (Guangzhou) \quad
\textsuperscript{3} Peking University \\[2pt]
\normalfont\small\texttt{zkouaa@connect.ust.hk, siruihan@ust.hk}
}
\begin{document}
\maketitle

\begin{abstract}
People adapt how they talk to the person in front of them. However, chat models tend to respond in Standard English regardless of the dialect used by the user. A uniform answer can hide whether training treats these users alike, and output audits cannot tell as well. We trace how each post-training stage, from supervised fine-tuning (SFT) to preference optimization (DPO) and reinforcement learning, changes a model's tendency to match the user, and read what each update prefers from two content-matched replies, one matching the user and one not. Across \res{design.n_lineages} open model lineages, post-training standardizes replies to every informal cue, yet after adjusting for how likely each reply already was, the DPO update still favours matching British and Australian spelling over AAE and Singlish on \res{dv.txt.adj} lineages, and models still produce AAE features when asked. We then introduce calibrated accommodation pairs (CAP), preference data that reward matching users' dialect density while preserving Standard English for Standard-English users. On a fully open model, CAP raises convergence on features it was not trained on without a detectable increase in dialect for Standard-English users or a cost in instruction following. In contrast, one-sided pairs reach a similar gain only by overshooting the user and imposing dialect. Our findings show that uniform replies can hide selective updates and audits should read the update as well as the output. Code: \href{https://github.com/kouzhizhuo/cap-accommodation}{cap-accommodation}.
\end{abstract}

\section{Introduction}
\label{sec:intro}

\begin{figure}[t]
  \centering
  \includegraphics[width=\columnwidth]{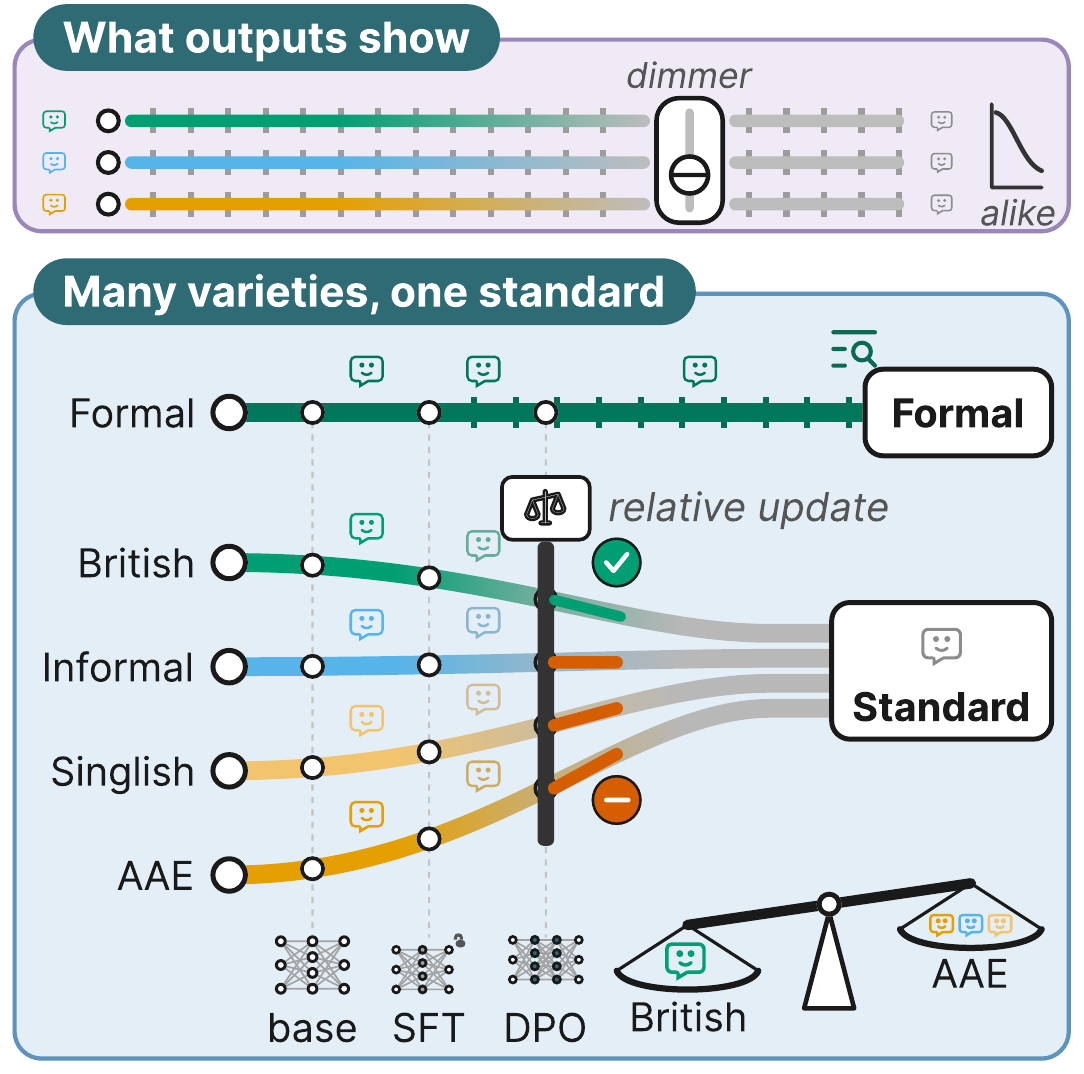}
  \caption{\textbf{Standard replies vs. selective updates.} \textbf{Top:} Post-training uniformly reduces outputs across all English varieties. \textbf{Bottom:} While replies across all varieties converge on Standard English, preference tuning specifically favors matching British spelling over stigmatized varieties.}
  \label{fig:teaser}
  \vspace{-18pt}
\end{figure}

A user asks a chatbot for advice: ``I usually buy theater tickets for my sister, who keep dragging me to the movies. My boss says the budget been canceled this month. Do y'all have any tips?'' (\Cref{fig:teaser}; replies in \Cref{app:case})
The request is in African American English (AAE); its Standard-English twin differs only in \emph{keeps}, \emph{has been} and \emph{you}.
People adjust their speech to the person they talk to, converging toward that person's way of speaking or holding back \citep{giles1991accommodation}, and deployed chatbots over-converge to their users' style \citep{blevins2026accommodation}.
Yet chat models default to Standard English with speakers of AAE, Singlish and Nigerian English \citep{fleisig-etal-2024-linguistic}. Features of these varieties are rare in chatbot traffic and training data (\Cref{app:dpo}).

Instruction-tuned models converge less to dialogue style than pretrained ones \citep{blevins-etal-2026-language}, reward models disprefer AAE \citep{mire-etal-2025-rejected}, and dialect gaps accumulate along the training pipeline \citep{elle2026dialect}.
Two questions remain open. First, closed models and base-versus-instruct comparisons cannot tell \emph{which stage} installs this default: supervised fine-tuning (SFT) on standard responses, or preference optimization.
Second, outputs alone cannot tell \emph{whether training treats users alike}: two models that answer everyone in Standard English may have learned different preferences.

We read accommodation on two channels at every training stage: the \emph{reply} (does it converge toward the user's variety?) and the \emph{update} (does the training step favour converging?).
We follow \res{design.n_lineages} open lineages through SFT, direct preference optimization (DPO) and reinforcement learning with verifiable rewards (RLVR); each checkpoint answers requests like the one above, written in several varieties and in Standard English.
We score only reply text not copied from the request, and read each update through the likelihood it gives to two content-matched replies, one that matches the user (\emph{your sister, who keep dragging you}) and one that does not (\emph{who keeps dragging you}).

Our contributions are as follows: \textbf{\ding{182} Standard replies, selective updates.} Post-training defaults to Standard English, but DPO selectively favors British and Australian spelling over AAE and Singlish on \res{dv.txt.adj} lineages.
\textbf{\ding{183} Controlled re-run.} Accommodation-consistent pairs boost convergence only at high doses, while undesirably introducing dialect for Standard English users.
\textbf{\ding{184} Calibrated accommodation pairs.} Penalizing under-accommodation, over-accommodation, and imposition in DPO data increases held-out convergence on OLMo-2-1B without dialect insertion.
\textbf{\ding{185} Diagnostic \& benchmark.} We introduce a copy-controlled convergence ratio and matching bonus to disentangle output behavior from training preferences (\href{https://anonymous.4open.science/r/cap-accommodation/}{anonymous code and data}).

Several pre-registered predictions did not hold, including that the split between prestige and stigmatized varieties grows at the preference stage, and that convergence rises gradually with the dose of accommodation-consistent preference pairs; \Cref{app:outcomes} reports every outcome.
Whether a model should reply in a user's variety is for its speakers to decide \citep{sandoval-etal-2025-llm}.

\section{Background and Related Works}
\label{sec:background}

\paragraph{Accommodation.}
Communication Accommodation Theory (CAT) holds that speakers adjust their speech to interlocutors for social reasons \citep{giles1991accommodation,giles2016communication}. \emph{Convergence} reduces linguistic differences to seek approval, \emph{divergence} accentuates them, and \emph{maintenance} keeps one's own variety; convergence flows upward, toward the higher-status variety, and a speaker able to converge may choose not to \citep{dragojevic2016accommodative}.
Priming accounts instead predict convergence to every variety \citep{pickering2004toward}, as in a pretrained model continuing a user's turn.
Language models converge strongly when completing human dialogues, instruction-tuned ones less \citep{blevins-etal-2026-language}; these studies measure style, we measure the features that distinguish varieties.\footnote{We use \emph{accommodation} in Giles' sense, not in the pragmatic sense of accommodating an interlocutor's presuppositions \citep{cheng-etal-2026-accommodation}.}

\paragraph{Standard language ideology.}
Standard-language ideology treats one variety as correct and the others as errors \citep{milroy2001language,lippigreen2012english}. It predicts upward convergence toward the standard and prestige cues such as British spelling, but not toward stigmatized grammar.
Language models show covert prejudice against AAE speakers \citep{hofmann2024ai}, reward models disprefer AAE \citep{mire-etal-2025-rejected}, preference tuning creates dialect disparities \citep{ryan-etal-2024-unintended}, and generative AI is argued to reproduce the ideology \citep{smith2026standard}; we ask where in training it enters.

\paragraph{Reading a training update.}
DPO makes the log-ratio between the trained and the reference policy an implicit reward \citep{rafailov2023direct}. We use the ratio between consecutive released stages descriptively, as an \emph{update diagnostic}, not an identified reward: compared across content-matched replies, it shows which reply a stage up- or down-weights. We pre-registered predictions from these accounts before the main runs (\Cref{app:theory}); related work is in \Cref{app:related}.

\section{Method}
\label{sec:method}
\begin{figure*}[t]
  \centering
  \includegraphics[width=\textwidth]{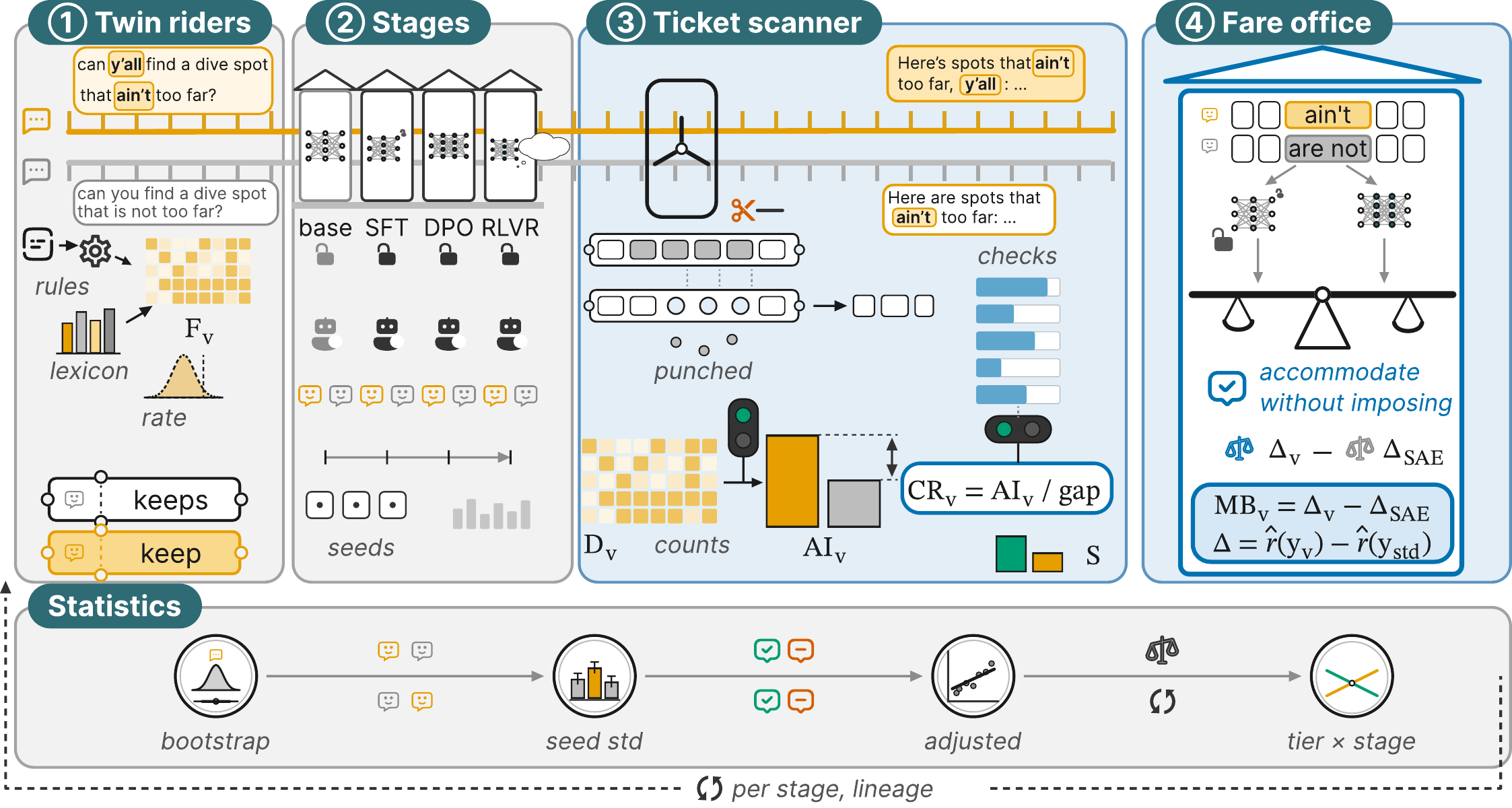}
  \caption{\textbf{Method: two riders, one trip.} The same request in a variety and in Standard English (\circled{1}) is answered at every released stage (\circled{2}). We score novel spans only (\circled{3}, $\CR{v}$) and compare how each update changes the likelihood of matched and standard replies (\circled{4}, $\MB{v}$).}
  \label{fig:method}
\end{figure*}

As shows in \Cref{fig:method} defines how we measure a model's convergence to its user and what each training update prefers and different languages.

\paragraph{\circled{1} Twin requests.}
We wrote \res{design.n_seeds} first-person requests in six task types by local native speaker with Ph.D level education rendered each, content fixed, in \res{design.n_varieties} varieties, in Standard American English (\vty{SAE}) and in two register controls, informal and formal Standard English (\vty{INF}, \vty{FOR}).
Pre-registered prestige tiers from attitude studies place British English (\vty{BRE}) at the top, Australian English (\vty{AUS}) in the middle, and AAE, Colloquial Singapore English (\vty{SGE}), Nigerian Pidgin (\vty{NGP}) and Appalachian English (\vty{APP}) as stigmatized (\Cref{app:method_more}).

\paragraph{\circled{2} Every released stage.}
We study \res{design.n_lineages} open lineages with every post-training stage public, so consecutive checkpoints differ by one stage: OLMo-2 (1B--32B) \citep{olmo2025olmo2}, OLMo-3 \citep{olmo2025olmo3}, T\"ulu-2/3 \citep{ivison2023camels,lambert2024tulu} and Zephyr \citep{tunstall2023zephyr}; popular families and deployed chatbots test breadth (\Cref{app:models}). We sample \res{design.n_samples_per_seed} replies per request and seed over \res{design.n_sampling_seeds} seeds (\Cref{app:method_more}).

\paragraph{\circled{3} Convergence.}
We score only \emph{novel spans}, tokens outside any word 4-gram shared with the request and outside quotations, so a reply that repeats ``keep dragging'' earns nothing.
For each retained feature $f\in F_v$ of variety $v$, a detector returns realizations $\real_f(x)$ and opportunities $\opp_f(x)$, sites where $f$ could apply but the standard variant occurs. The \emph{variety density} of a text $x$ is
\begin{equation}
  \Dens{v}(x) = \frac{\sum_{f\in F_v}\real_f(x)}{\sum_{f\in F_v}\big(\real_f(x)+\opp_f(x)\big)}.
  \label{eq:density}
\end{equation}
The \emph{accommodation} of stage $m$ compares its replies $r$ to the twins of requests $i$, over seeds $s$:
\begin{equation}
  \AIdx{v}(m) = \E_{i,s}\Dens{v}\big(r_{m,v,i,s}\big) - \E_{i,s}\Dens{v}\big(r_{m,\mathrm{SAE},i,s}\big).
  \label{eq:ai}
\end{equation}
The \emph{convergence ratio} normalizes it by the density gap between the requests $p$:
\begin{equation}
  \CR{v}(m) = \frac{\AIdx{v}(m)}{\E_i\,\Dens{v}(p_{v,i}) - \E_i\,\Dens{v}(p_{\mathrm{SAE},i})}.
  \label{eq:cr}
\end{equation}
$\CR{v}=1$: replies differ between twins as much as the requests do; $0$: not at all.
Where replies offer too few opportunities (spelling, \vty{SGE}, \vty{NGP}), we use a \emph{rate-based} ratio with realizations per 100 words (\Cref{app:method_more}).
The \emph{split} $\Split(m)$ is the mean CR over prestige cues (\vty{BRE}, \vty{INF}, \vty{FOR}) minus the mean CR over stigmatized varieties.\phantomsection\label{eq:split}
\emph{Feature competence} $\COMP{v}$ compares feature rates in replies explicitly asked to use $v$ with those in authentic text \citep{groenwold-etal-2020-investigating,chen2013creating}, averaged over features and clipped at 1.

\paragraph{\circled{4} The update diagnostic.}
Let $\hat r_m(x,y)=\tfrac{1}{|y|}\big[\log\pi_m(y\mid x)-\log\pi_{m^-}(y\mid x)\big]$ be the length-normalized log-ratio between stage $m$ and the stage $m^-$ before it. For each request, $y_{\mathrm{std}}$ is the SFT model's reply to the \vty{SAE} twin and $y_v$ its rendering in $v$, differing only in variety features. The \emph{matching bonus}
\begin{equation}
\begin{split}
  \MB{v}(m) ={}& \big[\hat r_m(x_v,y_v)-\hat r_m(x_v,y_{\mathrm{std}})\big]\\
  &- \big[\hat r_m(x_{\mathrm{SAE}},y_v)-\hat r_m(x_{\mathrm{SAE}},y_{\mathrm{std}})\big]
\end{split}
  \label{eq:mb}
\end{equation}
is positive when the update favours the matched reply more for a $v$-speaker than for a Standard-English speaker.

\paragraph{Statistics.}
Intervals are 95\% bootstrap CIs over requests (2{,}000 resamples); $\pm$ are standard deviations over seeds; contrasts are paired within request (\Cref{app:stats}). Design, measures and predictions were pre-registered (\Cref{app:prereg}).

\paragraph{\circled{5} The measurements hold up.}\label{sec:validity}
Detectors agree with blind expert-panel consensus labels on \res{audit.v4.b.primary.agreement.k} of \res{audit.v4.b.primary.agreement.n} requested replies ($\kappa=\res{audit.v4.b.primary.agreement.kappa}$); the recognisers over-fire on raw user requests, so we apply them only to model replies, where false hits occur equally for both twins and cancel (\Cref{app:method_more}).

\section{Replies Converge on Standard English}
\label{sec:replies}
\begin{figure*}[t]
  \centering
  \includegraphics[width=\textwidth]{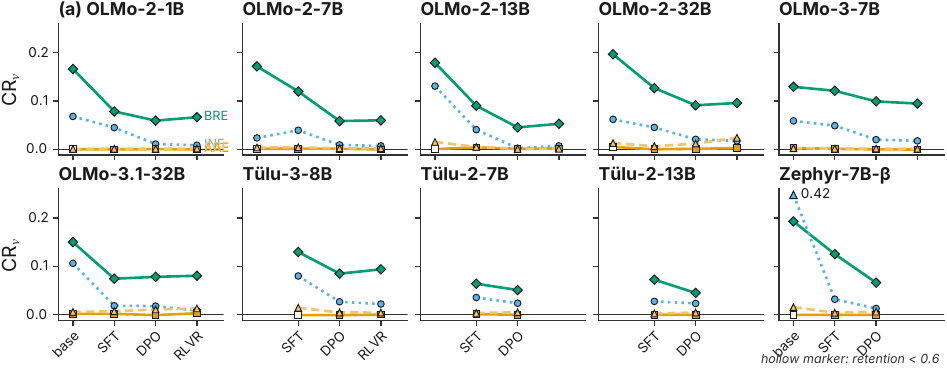}
  \caption{\textbf{Replies lose convergence through post-training.} Convergence ratio per cue and stage, by lineage; the update contrast is in \Cref{tab:contrast} and the two-panel version in \Cref{fig:heatmap_full}.}
  \label{fig:heatmap}
\end{figure*}

As \Cref{fig:heatmap} shows that post-training makes replies standard for every user, while the ability to produce dialect features remains.

\begin{table}[t]
  \centering
  \footnotesize
  \setlength{\tabcolsep}{4pt}
  \begin{tabular}{@{}l cc@{}}
    \toprule
    \textbf{Lineage} & Raw $\Delta$ & Adjusted $\Delta$ \\
    \midrule
    OLMo-3.1-32B & 3.0 {\scriptsize[1.6, 4.3]} & \cellcolor{cCmpA!27}4.7 {\scriptsize[3.0, 6.4]} \\
    OLMo-2-7B & 1.4 {\scriptsize[0.3, 2.4]} & \cellcolor{cCmpA!21}2.7 {\scriptsize[1.7, 3.7]} \\
    T\"ulu-3-8B & 3.1 {\scriptsize[2.0, 4.3]} & \cellcolor{cCmpA!21}2.7 {\scriptsize[1.5, 3.9]} \\
    OLMo-3-7B & 1.0 {\scriptsize[<0.1, 2.0]} & \cellcolor{cCmpA!17}1.8 {\scriptsize[0.4, 3.2]} \\
    T\"ulu-2-13B & 1.1 {\scriptsize[0.8, 1.5]} & \cellcolor{cCmpA!15}1.4 {\scriptsize[1.0, 1.9]} \\
    OLMo-2-1B & 0.4 {\scriptsize[<0.1, 0.9]} & \cellcolor{cCmpA!14}1.3 {\scriptsize[0.8, 1.9]} \\
    T\"ulu-2-7B & 1.0 {\scriptsize[0.7, 1.2]} & \cellcolor{cCmpA!14}1.2 {\scriptsize[0.8, 1.6]} \\
    OLMo-2-13B & 1.5 {\scriptsize[0.8, 2.2]} & \cellcolor{cCmpA!13}1.2 {\scriptsize[0.3, 2.1]} \\
    OLMo-2-32B & \ensuremath{-}3.1 {\scriptsize[\ensuremath{-}4.2, \ensuremath{-}2.1]} & \cellcolor{cCmpA!13}1.2 {\scriptsize[0.2, 2.2]} \\
    Zephyr-7B-$\beta$ & 1.2 {\scriptsize[0.9, 1.5]} & \cellcolor{cCmpA!11}0.9 {\scriptsize[0.4, 1.3]} \\
    \bottomrule
  \end{tabular}
  \caption{\textbf{After adjustment, every DPO update favours prestige matching.} Prestige-minus-stigmatized matching bonus, DPO vs.\ SFT, $\times$100 (nats/token), 95\% CI; grey: CI includes 0. Full panel: \Cref{fig:heatmap_full}.}
  \label{tab:contrast}
\end{table}

\begin{table*}[t]
  \centering
  \footnotesize
  \setlength{\tabcolsep}{3.2pt}
  \begin{tabular}{l cc cc cc ccc}
    \toprule
    & \multicolumn{2}{c}{\textbf{British spelling}} & \multicolumn{2}{c}{\textbf{Informal register}} & \multicolumn{2}{c}{\textbf{AAE}} & \multicolumn{3}{c}{\textbf{Update}} \\
    \cmidrule(lr){2-3} \cmidrule(lr){4-5} \cmidrule(lr){6-7} \cmidrule(lr){8-10}
    \textbf{Model} & first & last & first & last & first & last & $\MB{AAE}$ & $\Delta_{\text{raw}}$ & $\Delta_{\text{adj}}$ \\
    \midrule
    \multicolumn{10}{@{}l}{\textit{Popular families: base and released model}} \\
    Qwen2.5-7B & 11.8\std{0.7} & 3.4\std{0.6} & 3.3\std{0.6} & 1.9\std{0.7} & \textcolor{gray}{\ensuremath{-}0.1\std{0.2}} & \textcolor{gray}{\ensuremath{-}0.1\std{0.1}} & \ensuremath{-}0.9 & 2.3 & \cellcolor{cCmpA!23}\underline{3.2} \\
    Qwen3-8B & 12.4 & 6.1 & 9.7 & \textbf{3.6} & \textcolor{gray}{0.4} & \textcolor{gray}{<0.1} & 1.5 & 1.9 & \cellcolor{cCmpA!21}2.8 \\
    Qwen3.5-4B & 11.3 & 5.5 & 4.8 & 1.2 & \textcolor{gray}{<0.1} & \underline{\ensuremath{-}0.1} & 1.1 & \textcolor{gray}{\ensuremath{-}0.1} & \cellcolor{cCmpA!7}\textcolor{gray}{\ensuremath{-}0.4} \\
    Qwen3.5-9B & 7.6 & 4.7 & 4.3 & \textcolor{gray}{0.7} & \textcolor{gray}{0.1} & \textcolor{gray}{\ensuremath{-}0.1} & 0.9 & \textcolor{gray}{\ensuremath{-}0.4} & \cellcolor{cCmpA!12}1.0 \\
    \makecell[l]{DeepSeek-\\V2-Lite} & \textbf{22.0} & 7.0 & \underline{38.5} & 1.2 & \textbf{0.7} & \textcolor{gray}{0.1} & \textcolor{gray}{0.1} & \textcolor{gray}{0.4} & \cellcolor{cCmpA!17}1.8 \\
    DeepSeek-LLM-7B & 2.6 & 3.4 & 11.4 & 1.9 & \textcolor{gray}{1.0} & \textcolor{gray}{<0.1} & \ensuremath{-}0.8 & 1.6 & \cellcolor{cCmpA!14}1.3 \\
    MiMo-7B & 12.7 & 3.9 & 27.6 & \ensuremath{-}0.8 & \textcolor{gray}{0.1} & \textcolor{gray}{\ensuremath{-}0.1} & 0.5 & 0.7 & \cellcolor{cCmpA!13}1.2 \\
    \addlinespace[0.5mm]
    \multicolumn{10}{@{}l}{\textit{Open-recipe lineages: base and DPO}} \\
    OLMo-2-1B & 16.6\std{0.8} & 5.9\std{0.5} & 6.8\std{0.6} & 1.1\std{0.5} & \textcolor{gray}{<0.1\std{0.2}} & \textcolor{gray}{<0.1\std{0.1}} & \ensuremath{-}0.9 & 0.4 & \cellcolor{cCmpA!14}1.3 \\
    OLMo-2-7B & 17.1\std{0.5} & 5.8\std{0.1} & 2.3\std{0.3} & \textcolor{gray}{0.9\std{0.6}} & \textcolor{gray}{0.1\std{0.1}} & \textbf{0.1\std{0.1}} & \ensuremath{-}1.8 & 1.4 & \cellcolor{cCmpA!21}2.7 \\
    OLMo-2-13B & 17.9\std{1.0} & 4.5\std{0.5} & 13.1\std{1.4} & \textcolor{gray}{0.2\std{0.3}} & \textcolor{gray}{<0.1\std{<0.1}} & \textcolor{gray}{0.1\std{0.1}} & \underline{\ensuremath{-}2.1} & 1.5 & \cellcolor{cCmpA!13}1.2 \\
    OLMo-2-32B & \underline{19.6\std{1.8}} & \underline{9.1\std{0.8}} & 6.2\std{0.2} & 2.0\std{0.3} & \underline{0.5\std{0.1}} & \textcolor{gray}{0.1\std{0.1}} & 2.0 & \textbf{\ensuremath{-}3.1} & \cellcolor{cCmpA!13}1.2 \\
    OLMo-3-7B & 12.9\std{1.5} & \textbf{9.9\std{0.3}} & 5.9\std{0.8} & 2.0\std{1.2} & \textcolor{gray}{0.2\std{0.5}} & \textcolor{gray}{\ensuremath{-}<0.1\std{0.1}} & \textbf{\ensuremath{-}2.6} & 1.0 & \cellcolor{cCmpA!17}1.8 \\
    T\"ulu-3-8B$^{\dagger}$ & 13.0\std{0.6} & 8.5\std{0.3} & 8.0\std{0.4} & \underline{2.7\std{0.8}} & \textcolor{gray}{\ensuremath{-}0.1\std{0.3}} & \textcolor{gray}{\ensuremath{-}0.1\std{0.2}} & \ensuremath{-}1.8 & \underline{3.1} & \cellcolor{cCmpA!21}2.7 \\
    T\"ulu-2-7B$^{\dagger}$ & 6.4 & 5.1 & 3.6 & 2.4 & 0.1 & \textcolor{gray}{<0.1} & \ensuremath{-}1.1 & 1.0 & \cellcolor{cCmpA!14}1.2 \\
    T\"ulu-2-13B$^{\dagger}$ & 7.3 & 4.5 & 2.8 & 2.3 & \textcolor{gray}{<0.1} & \textcolor{gray}{\ensuremath{-}<0.1} & \ensuremath{-}1.0 & 1.1 & \cellcolor{cCmpA!15}1.4 \\
    Zephyr-7B-$\beta$ & 19.3 & 6.6 & \textbf{42.2} & 1.3 & \textcolor{gray}{0} & \textcolor{gray}{\ensuremath{-}<0.1} & \ensuremath{-}1.4 & 1.2 & \cellcolor{cCmpA!11}0.9 \\
    OLMo-3.1-32B & 15.0 & 7.8 & 10.6 & 1.8 & \textcolor{gray}{0.2} & \textcolor{gray}{\ensuremath{-}0.1} & \ensuremath{-}1.3 & 3.0 & \cellcolor{cCmpA!27}\textbf{4.7} \\
    \bottomrule
  \end{tabular}
  \caption{\textbf{Post-training lowers register convergence everywhere and spelling convergence in 16 of 17 models; the adjusted update favours prestige matching in all but one.} $\times$100; bold, underline: top two; grey: CI.}
  \label{tab:main}
  \par\vspace{2pt}{\footnotesize\raggedright First: base; $^{\dagger}$SFT where the base is gated. Last: DPO, or the released model for popular families, whose update spans all post-training. $\Delta$: prestige-minus-stigmatized contrast, raw and adjusted for baseline likelihood and length. Chat-only releases: \Cref{tab:mainfull}. Last for MiMo-7B: RL. Every stage: \Cref{tab:mainfull}.\par}
\end{table*}

\subsection{Post-Training Standardizes Replies to Every Informal Cue}
\label{sec:exp:h1}

\textbf{Post-training standardizes replies to every informal cue} ( \Cref{tab:contrast}).
Base OLMo-2-7B reproduces a user's British spelling at \res{e1.l2.base.bre.cr_rate} (95\% CI \rci{e1.l2.base.bre.cr_rate}); after SFT the rate is \res{e1.l2.sft.bre.cr_rate}, after DPO \res{e1.l2.dpo.bre.cr_rate} (\rci{e1.l2.dpo.bre.cr_rate}), and RLVR leaves it at \res{e1.l2.instruct.bre.cr_rate}.
Informal register follows the same path (\res{e1.l2.base_f1.inf.cr} at base under the chat template, \res{e1.l2.dpo.inf.cr} after DPO).
The preference stage, with the chat template fixed, lowers convergence to informal register on \res{dv.txt.inf_dpo} lineages and to British spelling on \res{dv.txt.bre_dpo} (stage and framing details in \Cref{app:results_more}).
\textbf{The exception is formal register.} Replies keep converging to formal requests at every stage (OLMo-2-7B after DPO: \res{e1.l2.dpo.for.cr}), so post-training not replies formal.

\textbf{Dialect grammar was mirrored only weakly to begin with.} Continuing a user's text, base OLMo-2-7B reproduces AAE features at \res{e1.l2.base_raw.aae.cr} (\rci{e1.l2.base_raw.aae.cr}) and Singlish lexis at \res{e1.l2.base_raw.sge.cr_lex}. After SFT both are near zero, also on human-written AAE tweets (\Cref{app:authentic}).

\textbf{The ordering by prestige is already present in the base model.} Convergence follows the pre-registered prestige tiers at base ($\rho=\res{e1.l2.base.rho}$ on OLMo-2-7B) and after DPO ($\rho=\res{e1.l2.dpo.rho}$): post-training lowers the level of convergence but does not install the order.

\begin{figure}[t]
  \centering
  \includegraphics[width=\columnwidth]{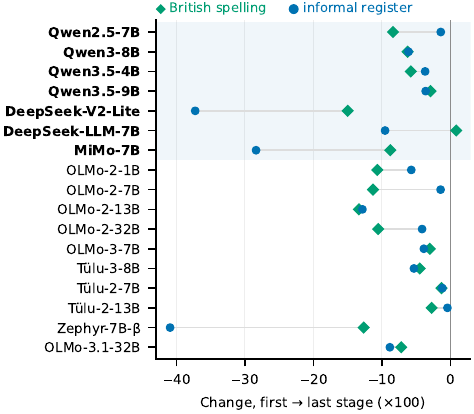}
  \caption{\textbf{Post-training lowers convergence in popular and open-recipe models alike.} Change from first to last stage per model; bold: popular families. Levels and update: \Cref{tab:main,fig:models_full}.}
  \label{fig:models}
\end{figure}

\textbf{The decline holds at scale and across popular families} (\Cref{tab:main,fig:models}).

From base to final release, informal-register convergence falls in all \res{dv.fam.n} further families, British-spelling convergence in \res{dv.fam.bre_falls}, and AAE ends at most \res{dv.fam.aae_final_max}.
\textbf{Deployed chatbots standardize AAE too, but partly answer Nigerian Pidgin in kind:} across \res{dv.dep.n}, convergence is at most \res{dv.dep.inf_max} for informal register and \res{dv.dep.aae_max} for AAE, while British spelling is reproduced at \res{dv.dep.bre_min}--\res{dv.dep.bre_max} and Nigerian Pidgin at \res{dv.dep.ngp_min}--\res{dv.dep.ngp_max}, above any post-trained open model (at most \res{dv.dep.ngp_open_max}; \Cref{tab:mainfull}). In open models, post-training standardizes replies for users of every variety, not singling out stigmatized varieties.

\subsection{Models Still Produce AAE Features on Request}
\label{sec:exp:h3}

\textbf{Two of three families still produce AAE features at more than half their authentic rate when asked} (\Cref{tab:comp} in \Cref{app:results}).
Asked to reply in the user's variety, OLMo-2-7B after DPO realizes AAE features at \res{dv.comp_corr.l2_aae.comp} of their authentic rate, against \res{comp.l2.dpo.aae.default_ratio} in its default replies; T\"ulu-3-8B behaves alike (\res{dv.comp_corr.l4_aae.comp}).
OLMo-3-7B reaches \res{dv.comp_corr.l3_aae.comp}; rates are corrected by the blind human audit (\Cref{app:results_more}).
Requested imitation increases stereotyping \citep{fleisig-etal-2024-linguistic}, so competence measures ability, not reply quality. AAE features are produced on request in all three families, at more than half their authentic rate in two rarely by default.

\section{The Update Is Selective}
\label{sec:updates}
This section shows that, beneath these uniform replies, the DPO update still prefers matching some users over others.

\subsection{The DPO Update Favours Matching Prestige Spelling over Stigmatized Varieties}
\label{sec:exp:h4}

\begin{table}[t]
  \centering
  \scriptsize
  \setlength{\tabcolsep}{1.8pt}
  \begin{tabular}{@{}l ccc cc@{}}
    \toprule
    & \multicolumn{3}{c}{\textbf{Matching bonus $\MB{v}$}} & \multicolumn{2}{c}{\textbf{Contrast $\Delta$}} \\
    \cmidrule(lr){2-4} \cmidrule(lr){5-6}
    \textbf{Update} & \vty{AAE} & \vty{SGE} & \vty{BRE} & raw & adj. \\
    \midrule
    \multicolumn{6}{@{}l}{\textit{DPO vs.\ SFT}} \\
    OLMo-2-1B & \ensuremath{-}0.9 & \textcolor{gray}{0.5} & 0.4 & 0.4 & \cellcolor{cCmpA!14}1.3 \\
    OLMo-2-7B & \ensuremath{-}1.8 & \textcolor{gray}{\ensuremath{-}0.7} & \textcolor{gray}{0.2} & 1.4 & \cellcolor{cCmpA!21}2.7 \\
    OLMo-2-13B & \ensuremath{-}2.1 & \ensuremath{-}1.5 & \textcolor{gray}{\ensuremath{-}0.2} & 1.5 & \cellcolor{cCmpA!13}1.2 \\
    OLMo-2-32B & 2.0 & 7.1 & 1.7 & \ensuremath{-}3.1 & \cellcolor{cCmpA!13}1.2 \\
    OLMo-3-7B & \ensuremath{-}2.6 & \ensuremath{-}2.5 & \ensuremath{-}1.4 & 1.0 & \cellcolor{cCmpA!17}1.8 \\
    T\"ulu-3-8B$^{\dagger}$ & \ensuremath{-}1.8 & \ensuremath{-}2.3 & 0.9 & 3.1 & \cellcolor{cCmpA!21}2.7 \\
    T\"ulu-2-7B$^{\dagger}$ & \ensuremath{-}1.1 & \ensuremath{-}1.2 & \textcolor{gray}{\ensuremath{-}<0.1} & 1.0 & \cellcolor{cCmpA!13}1.2 \\
    T\"ulu-2-13B$^{\dagger}$ & \ensuremath{-}1.0 & \ensuremath{-}0.8 & \textcolor{gray}{0.2} & 1.1 & \cellcolor{cCmpA!15}1.4 \\
    Zephyr-7B-$\beta$ & \ensuremath{-}1.4 & \ensuremath{-}0.9 & \textcolor{gray}{\ensuremath{-}<0.1} & 1.2 & \cellcolor{cCmpA!11}0.9 \\
    OLMo-3.1-32B & \ensuremath{-}1.3 & \textcolor{gray}{\ensuremath{-}0.4} & 1.9 & 3.0 & \cellcolor{cCmpA!27}4.7 \\
    \multicolumn{6}{@{}l}{\textit{SFT vs.\ base}} \\
    OLMo-2-1B & \ensuremath{-}1.7 & \ensuremath{-}3.2 & 1.0 & 3.5 & \cellcolor{cCmpA!20}2.6 \\
    OLMo-2-7B & \ensuremath{-}1.5 & \ensuremath{-}2.1 & 1.1 & 3.2 & \cellcolor{cCmpA!23}3.3 \\
    Zephyr-7B-$\beta$ & \ensuremath{-}1.1 & 2.0 & 0.6 & \textcolor{gray}{0.1} & \cellcolor{cCmpA!12}1.0 \\
    OLMo-3.1-32B & \ensuremath{-}1.2 & \textcolor{gray}{\ensuremath{-}0.5} & 1.8 & 2.6 & \cellcolor{cCmpA!22}2.9 \\
    \multicolumn{6}{@{}l}{\textit{Final vs.\ base (other families)}} \\
    Mistral-7B-v0.3 & \ensuremath{-}0.6 & \textcolor{gray}{0.9} & 1.8 & 1.6 & \cellcolor{cCmpA!18}2.1 \\
    Mistral-Small-24B & \ensuremath{-}0.6 & \textcolor{gray}{0.4} & 1.8 & 1.8 & \cellcolor{cCmpA!18}2.1 \\
    Qwen3-8B & 1.5 & \textcolor{gray}{0.5} & 3.0 & 1.9 & \cellcolor{cCmpA!21}2.8 \\
    Falcon3-7B & \textcolor{gray}{\ensuremath{-}0.1} & 3.2 & 2.4 & 1.0 & \cellcolor{cCmpA!19}2.2 \\
    Granite-3.3-8B & 2.1 & 5.8 & 4.3 & \textcolor{gray}{0.2} & \cellcolor{cCmpA!16}1.6 \\
    Yi-1.5-9B & \textcolor{gray}{\ensuremath{-}0.2} & \textcolor{gray}{0.7} & 1.8 & 1.8 & \cellcolor{cCmpA!19}2.3 \\
    SmolLM3-3B & \ensuremath{-}0.6 & 0.5 & 0.9 & 0.9 & \cellcolor{cCmpA!16}1.6 \\
    SmolLM3 APO vs.\ SFT & \ensuremath{-}0.2 & 0.4 & \textcolor{gray}{\ensuremath{-}<0.1} & \ensuremath{-}0.1 & \cellcolor{cCmpA!5}\textcolor{gray}{0.2} \\
    DeepSeek-V2-Lite & \textcolor{gray}{0.1} & 3.8 & 2.2 & \textcolor{gray}{0.4} & \cellcolor{cCmpA!17}1.8 \\
    DeepSeek-LLM-7B & \ensuremath{-}0.8 & \ensuremath{-}1.3 & 0.5 & 1.6 & \cellcolor{cCmpA!14}1.3 \\
    Qwen2.5-7B & \ensuremath{-}0.9 & \textcolor{gray}{\ensuremath{-}0.4} & 1.6 & 2.3 & \cellcolor{cCmpA!23}3.2 \\
    MiMo-7B & 0.5 & 1.0 & 1.5 & 0.7 & \cellcolor{cCmpA!13}1.2 \\
    Qwen3.5-4B & 1.1 & 1.4 & 1.2 & \textcolor{gray}{\ensuremath{-}0.1} & \cellcolor{cCmpA!7}\textcolor{gray}{\ensuremath{-}0.4} \\
    Qwen3.5-9B & 0.9 & 2.8 & 1.4 & \textcolor{gray}{\ensuremath{-}0.4} & \cellcolor{cCmpA!12}1.0 \\
    \bottomrule
  \end{tabular}
  \caption{\textbf{After adjustment, no update significantly favours stigmatized over prestige matching.} Values $\times$100; grey: 95\% CI includes 0; adj.: adjusted for baseline likelihood and length.}
  \label{tab:update}
\end{table}

Within a request, the DPO update favours matching British and Australian spelling over AAE and Singlish on \res{dv.txt.adj} lineages after adjusting for baseline likelihood and length (\Cref{tab:contrast,tab:main}).
Pooled over lineages (the analysis fixed at the gate), the contrast is \res{e4i.dpo_vs_sft.adjusted.coef} nats per token (CI [\res{e4i.dpo_vs_sft.adjusted.lo}, \res{e4i.dpo_vs_sft.adjusted.hi}]) and stays positive when any one lineage is left out; the per-lineage fits, made after seeing the 32B data, are all positive. This paired test postdates the pre-registration; the pre-registered mixed model gives a positive tier effect on \res{dv.n_mixed_tier_pos} of \res{dv.n_lineages} lineages but a CI above zero only on \res{dv.n_mixed_tier_sig} (\Cref{app:outcomes}).
Per variety (\Cref{tab:update}), the bonus is negative for AAE on \res{dv.n_mb_aae_neg} of \res{dv.n_lineages} lineages (OLMo-2-7B: \res{e4.l2.dpo_vs_sft.aae.mb}, CI \rci{e4.l2.dpo_vs_sft.aae.mb}), for Singlish on \res{dv.n_mb_sge_neg} and for informal register on \res{dv.n_mb_inf_neg}, and not detectably negative for British and Australian spelling on \res{dv.n_mb_prestige_spared}.

\textbf{Raw and adjusted contrasts differ on OLMo-2-32B,} whose update up-weights replies the SFT model found unlikely far more than any other: its raw contrast reverses (\res{e4i.dpo_vs_sft.per_lineage.olmo2_32b.raw}) while the adjusted one is positive (\res{e4a.olmo2_32b.adj}); OLMo-3.1-32B does not reverse, so size alone does not explain it (\Cref{app:results_more}).
\textbf{The supervised update shows the same contrast, and RLVR adds little} (\Cref{app:results_more}).
\textbf{The contrast also appears within lexical features.} Singlish particles and British spelling, both lexical, get opposite signs on \res{dv.n_sge_neg_bre_nonneg} of \res{dv.n_lineages} lineages.

\textbf{Popular families point the same way.} Across \res{e4a.breadth_n} further updates (base to final release in popular families, and SmolLM3's preference stage), the adjusted contrast is positive in \res{e4a.breadth_adj_pos}, reversed in \res{e4a.breadth_adj_neg}, and inconclusive for \res{e4a.breadth_inconcl_names} (\Cref{tab:update}). Replies are standardized for every user, but the update that produces them treats varieties differently.

\subsection{Added Preference Pairs Raise Convergence Only at the Largest Dose Tested}
\label{sec:exp:h5}
\textbf{Dialect-bearing prompts are rare in the released preference data.} Only \res{dv.pref.olmo2_7b}\% of the prompts in the OLMo-2-7B preference mix and \res{dv.pref.tulu3}\% in T\"ulu-3's trigger the dialect recognisers (\Cref{app:dpo}), so the selectivity arises from few dialect-bearing pairs.

\begin{figure}[t]
  \centering
  \includegraphics[width=\columnwidth]{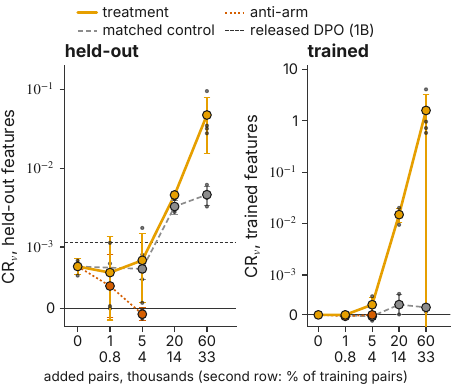}
  \caption{\textbf{Added pairs raise convergence only at the largest dose, by different amounts across runs.} Convergence on (a) held-out and (b) trained features by added pairs; mean\std{std} over runs; dashed: released DPO.}
  \label{fig:e5}
\end{figure}

\textbf{Adding accommodation-consistent pairs raises convergence only at the largest dose we tested.} We re-ran DPO from the released OLMo-2-1B SFT checkpoint with its recipe on a fixed subset of its preference mix, adding pairs whose chosen reply is the AAE or Singlish rendering of the rejected standard reply; matched controls add the same number of standard pairs (\Cref{fig:e5}; \Cref{tab:mitigation}).
At 1k (\res{dv.e5dose.share_1k}\% of pairs, about the natural share), 5k and 20k added pairs (\res{dv.e5dose.share_20k}\%), held-out convergence stays at its dose-0 level and matches the controls (20k: \res{e5t.crheld.effect_20k_T_minus_C.mean}, CI [\res{e5t.crheld.effect_20k_T_minus_C.lo}, \res{e5t.crheld.effect_20k_T_minus_C.hi}]).
At 60k (\res{dv.e5dose.share_60k}\%; added after pre-registration), held-out convergence rises above the control in all \res{dv.e560.n} training runs, by different amounts (\res{dv.e560.eff_s0}, \res{dv.e560.eff_s1}, \res{dv.e560.eff_s2} and \res{dv.e560.eff_s3}, each with a CI above zero; \res{e5t.crheld.effect_60k_T_minus_C.mean} on average, CI [\res{e5t.crheld.effect_60k_T_minus_C.lo}, \res{e5t.crheld.effect_60k_T_minus_C.hi}]). The gain transfers to the held-out varieties (\res{e5t.crg.effect_60k_T_minus_C.mean} on average), and IFEval stays within the pre-registered two-point bound (prompt-level accuracy \res{dv.ifeval.t60k.strict}\% over the \res{dv.ifeval.t60k.n} runs, against \res{dv.ifeval.d0.strict}\% at dose 0 and \res{dv.ifeval.c60k.strict}\% for the matched controls).

\textbf{The added pairs raise convergence without calibrating it.} In all \res{e5c.t60k.n_runs} runs at 60k, Standard-English replies gain \res{e5c.t60k.unsolicited_min}--\res{e5c.t60k.unsolicited_max} trained dialect features per 100 words (controls: \res{e5c.controls_unsolicited_max}); Singlish replies carry \res{e5c.t60k.sge_ratio_min}--\res{e5c.t60k.sge_ratio_max} times the user's trained-feature rate, AAE replies only \res{e5c.t60k.aae_ratio_min}--\res{e5c.t60k.aae_ratio_max} times. Preference data can raise convergence, but only at a dose far above the share of dialect-bearing prompts, unevenly across runs, and without calibration to the user.

\subsection{Sensitivity Analysis}
\label{sec:sensitivity}

\textbf{The findings hold under other analysis choices.} Changing one choice at a time (estimator, bootstrap, reply length, sampling seeds, human-audit correction) leaves both findings in place (\Cref{tab:sensitivity}; details in \Cref{app:results_more}).

\section{Calibrated Accommodation Pairs}
\label{sec:cap}
This section introduces calibrated accommodation pairs and tests whether they make models match their users without imposing dialect.

\begin{figure}[t]
  \centering
  \includegraphics[width=\columnwidth]{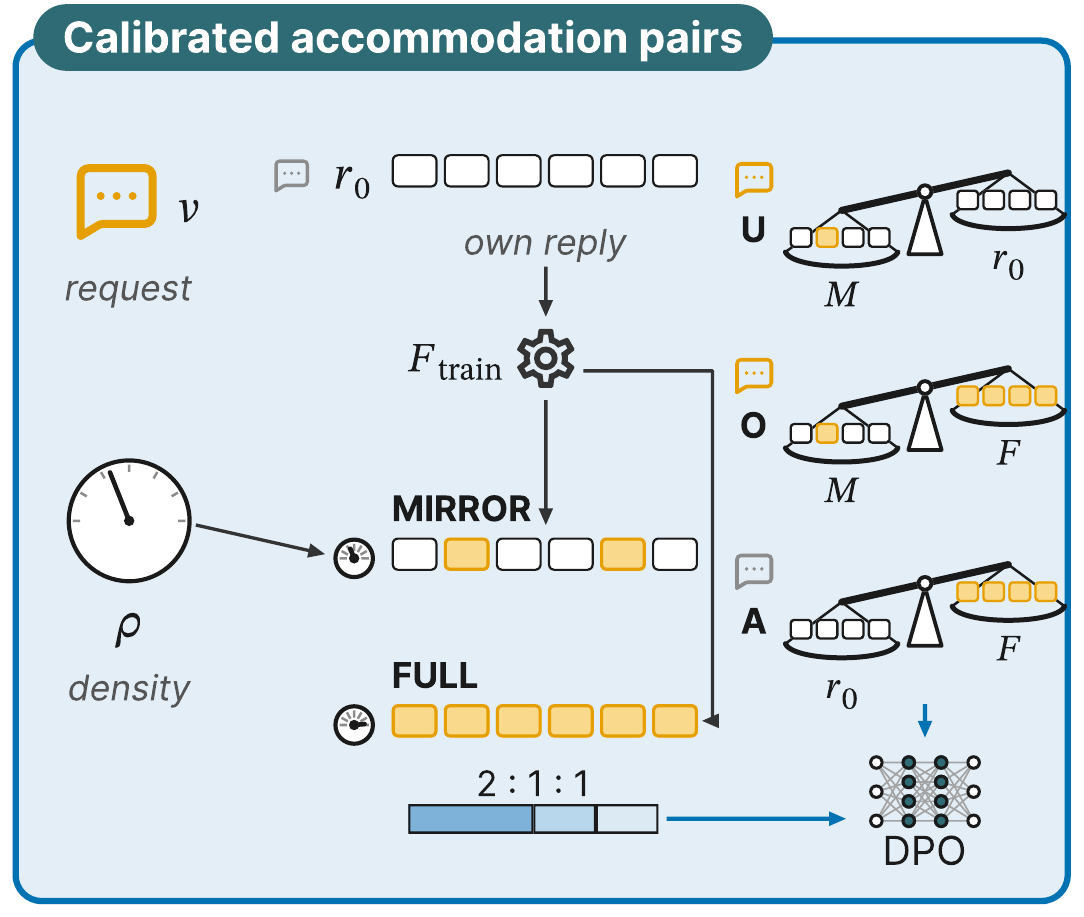}
  \caption{\textbf{Calibrated accommodation pairs.} The model's reply $r_0$ to the Standard-English twin is rendered at the user's density ($M$) and in every sentence ($F$). DPO pairs prefer $M$ over $r_0$ and over $F$ for a dialect user, and $r_0$ over $F$ for a Standard-English user.}
  \label{fig:cap}
\end{figure}

The re-run raises convergence without calibrating it (\Cref{sec:exp:h5}): Standard-English replies gain dialect features, Singlish is over-produced and AAE under-produced. We turn this diagnosis into a recipe for preference data.

\paragraph{Method.} \emph{Calibrated accommodation pairs} (CAP) re-render the model's own reply $r_0$ to a request's Standard-English twin with the trained features only (\Cref{fig:cap}). For a dialect request, $M$ renders $r_0$ at the trained-feature density closest to the user's, at most 1.25 times it. U pairs prefer $M$ over $r_0$, O pairs prefer $M$ over $F$, a rendering with features in every sentence, and, for the Standard-English twin, A pairs prefer $r_0$ over $F$. These pairwise preferences penalize under-accommodation, over-accommodation and imposition, without guaranteeing calibrated generation. The pairs (U\,:\,O\,:\,A $=2:1:1$) make up a third of the DPO data, the share at which one-sided pairs raised convergence; held-out features and varieties test transfer.

\subsection{Experimental Settings}
\label{sec:cap:settings}

\paragraph{Setup.} We train OLMo-2-1B, whose stages are all released and whose controls carry over from \Cref{sec:exp:h5}, at a fixed budget per arm, against a control preferring one standard reply over another and the one-sided pairs; the test set is 300 fresh requests from a disjoint pool (\Cref{tab:capsetup}).

\paragraph{Metrics.} Criteria frozen before any result (\Cref{app:cap}): held-out convergence at least 0.02 above the control (M1); calibration closer to the user's density than the one-sided recipe, without insertion (M2); tracking of a within-request density intervention (M3); IFEval within two points (M4); compliance with requests for Standard English (M5).

\paragraph{Pilot.} The pre-declared pilot passed \res{dv.gate.n_pass} of \res{dv.gate.n_checks} validation checks but under-accommodates; we continued as pre-declared, without tuning (\Cref{app:cap_more}).

\subsection{Main Results}
\label{sec:cap:results}
\begin{table*}[t]
  \centering
  \footnotesize
  \setlength{\tabcolsep}{2.3pt}
  \begin{tabular}{@{}ll c cc cc c c c cc@{}}
    \toprule
    & & & \multicolumn{2}{c}{\textbf{Convergence}} & \multicolumn{2}{c}{\textbf{Reply / user density}} & & & & \multicolumn{2}{c}{\textbf{Capability}} \\
    \cmidrule(lr){4-5} \cmidrule(lr){6-7} \cmidrule(lr){11-12}
    \textbf{Model} & \textbf{Arm} & Runs & held-out & varieties & AAE & Singlish & Insertion & Tracking & Std.\ request & IFEval & GSM8K \\
    \midrule
     & Control & 4 & \cellcolor{cCmpA!4}0.2 & 0.5 & 0.00 & \underline{0.00} & 0.01 & \underline{0.00} & \textbf{0.00} & \underline{58.4} & 60.5 \\
     & One-sided & 4 & \cellcolor{cCmpA!17}\underline{4.5} & \underline{3.7} & \textbf{0.18} & 2.59 & 0.57 & 3.61 & 0.96 & 58.0 & \textbf{61.0} \\
    \rowcolor{gray!10} \multirow{-3}{*}{OLMo-2-1B} & \textbf{CAP} & 3 & \cellcolor{cCmpA!27}\textbf{17.6} & \textbf{15.5} & \underline{0.17} & \textbf{0.79} & 0.01 & \textbf{0.19} & \underline{0.68} & \textbf{60.4} & \underline{60.6} \\
    \bottomrule
  \end{tabular}
  \caption{\textbf{On OLMo-2-1B, CAP raises held-out convergence without inserting dialect; one-sided pairs insert it.} Absolute convergence $\times$100; rates per 100 words; accuracy in \%; bold, underline: best, second.}
  \label{tab:cap}
  \par\vspace{2pt}{\footnotesize\raggedright Ratio: reply over user density of the trained features (1 = mirroring). Tracking: within-request slope of reply on user density, Singlish. Std.\ request: dialect features after ``Please answer in Standard English.'' Registered criteria for CAP: \res{dv.txt.capcrit_b1}.\par}
\end{table*}

\paragraph{CAP raises held-out convergence without a detectable increase in insertion.} On OLMo-2-1B (\Cref{tab:cap}), CAP raises held-out convergence by \res{dv.capres.b1.cap_crheld} over the matched controls (CI [\res{dv.capres.b1.cap_crheld_lo}, \res{dv.capres.b1.cap_crheld_hi}]), \res{dv.capres.b1.fold} times the gain of the registered one-sided pairs (\res{dv.capres.b1.unm_crheld}; the ratio also reflects their older pair builder, \Cref{sec:cap:ablation}), and on the held-out varieties (Nigerian Pidgin, Appalachian English) by \res{dv.capres.b1.cap_crg} (one-sided: \res{dv.capres.b1.unm_crg}). For Standard-English users, insertion changes by \res{dv.capres.b1.cap_insertion} features per 100 words (CI [\res{dv.capres.b1.cap_insertion_lo}, \res{dv.capres.b1.cap_insertion_hi}]), where the one-sided pairs insert \res{dv.capres.b1.unm_insertion}, at no cost in instruction following (IFEval \res{dv.capres.b1.cap_ifeval} points, CI [\res{dv.capres.b1.cap_ifeval_lo}, \res{dv.capres.b1.cap_ifeval_hi}]). Its Singlish replies no longer overshoot (\res{dv.capres.b1.cap_ratio_sge} of the user's density, against \res{dv.capres.b1.unm_ratio_sge}), but calibration, tracking and compliance with requests for Standard English fail as registered (M2, M3, M5; \Cref{app:cap_more}).

On Qwen3-8B, CAP gives \res{dv.capres.q3.cap_crheld} (CI [\res{dv.capres.q3.cap_crheld_lo}, \res{dv.capres.q3.cap_crheld_hi}]), below the one-sided recipe's \res{dv.capres.q3.unm_crheld} and short of M1 (\Cref{app:cap}).

\subsection{Ablation Study}
\label{sec:cap:ablation}
Each ablation removes one component of CAP at the same budget (\Cref{tab:capabl}; variants in \Cref{app:cap_more}). 
\begin{table}[t]
  \centering
  \scriptsize
  \setlength{\tabcolsep}{2pt}
  \begin{tabular}{@{}l c cc cc c@{}}
    \toprule
    & & \multicolumn{2}{c}{\textbf{Convergence}} & \multicolumn{2}{c}{\textbf{Dialect /100 words}} & \\
    \cmidrule(lr){3-4} \cmidrule(lr){5-6}
    \textbf{Variant} & Runs & held-out & varieties & insertion & Std.\ req. & IFEval \\
    \midrule
    \rowcolor{gray!10} \textbf{CAP} & 3 & \cellcolor{cCmpA!20}17.3 & 15.0 & 0.00 & 0.68 & 2.1 \\
    w/o O pairs & 2 & \cellcolor{cCmpA!27}58.3 & 69.2 & 0.04 & 7.36 & 0.0 \\
    w/o A pairs & 2 & \cellcolor{cCmpA!6}0.8 & \ensuremath{-}0.1 & 0.01 & 0.19 & 1.0 \\
    mirrored SFT & 1 & \cellcolor{cCmpA!3}\ensuremath{-}0.1 & 0.1 & 0.03 & 0.12 & \ensuremath{-}5.9 \\
    target $1.5\times$ (expl.) & 1 & \cellcolor{cCmpA!24}34.0 & 28.8 & \ensuremath{-}0.01 & 1.61 & 0.6 \\
    one-sided (rebuilt) & 2 & \cellcolor{cCmpA!21}20.1 & 8.3 & 0.38 & 2.42 & 0.0 \\
    one-sided & 4 & \cellcolor{cCmpA!12}4.3 & 3.2 & 0.56 & 0.95 & \ensuremath{-}0.3 \\
    \bottomrule
  \end{tabular}
  \caption{\textbf{The O pairs buy restraint, the A pairs and the preference contrast carry the gain, and density matching curbs overshoot.} OLMo-2-1B, each variant minus the matched control; convergence $\times$100; IFEval in points.}
  \label{tab:capabl}
\end{table}

Without the O pairs, held-out convergence rises to \res{dv.capabl.noO_crheld} but Standard-English replies gain \res{dv.capabl.noO_ins} dialect features per 100 words, so the O pairs buy restraint. Without the A pairs, convergence barely moves (\res{dv.capabl.noA_crheld}); fine-tuning on the mirrored replies instead of preferring them leaves it unchanged (\res{dv.capabl.msft_crheld}): the gain comes from the preference contrast. One-sided pairs from CAP's builder gain similarly (\res{dv.capabl.mx_crheld}) only by overshooting, with Singlish replies at \res{dv.capabl.mx_sge_ratio} times the user's density (CAP: \res{dv.capabl.cap_sge_ratio}) and \res{dv.capabl.mx_ins} inserted features per 100 words.

\subsection{Sensitivity Analysis}
\label{sec:cap:sensitivity}
CAP's held-out gain holds under a request-only bootstrap, without the false-positive exclusion, and on the requests of \Cref{sec:replies} (\res{dv.capsens.v1}; \Cref{app:cap_more}).

On the lineage whose every stage is released, CAP raises held-out convergence with no detectable increase in dialect for Standard-English users; one-sided pairs built the same way reach a similar gain only by overshooting the user and imposing it.
\section{Discussion}
\label{sec:discussion}

Output audits find one Standard-English answer for every user; the update diagnostic, read from public checkpoints without retraining, shows training moving the policy toward matching some users and away from others. Model releases should report it next to output audits. The standard default arises with little dialect-specific data (under 1\% of preference prompts) and is not inevitable, since deployed chatbots answer Nigerian Pidgin in kind. The prestige ordering in the update is consistent with prestige-directed convergence but does not identify prestige as its cause, and raw and adjusted contrasts answer different questions, so update diagnostics should report both (\Cref{app:discussion}).

\section{Conclusion}
\label{sec:conclusion}

Along every released stage of open model lineages, post-training standardizes replies to every informal cue, prestige spelling included, while the training update favours matching prestige spelling over stigmatized varieties, and models still produce AAE features on request; adding accommodation-consistent pairs to a third of the DPO data raises held-out convergence over matched controls in every run, though without calibrating it.
Calibrated accommodation pairs address that failure: on OLMo-2-1B they raise held-out convergence with no detectable increase in dialect for Standard-English users, though explicit requests for Standard English still receive some, whereas one-sided pairs built the same way reach a similar gain only by overshooting the user's density and imposing it.
Uniform replies can hide a selective update; audits should read the update as well as the output.

\section*{Limitations}
\label{sec:limitations}

We study English, six varieties and open lineages up to 32B whose every stage is public; closed chatbots enter only through their behaviour. All requests are rule-based renderings that are not calibrated to the feature densities of authentic text, so convergence on real users' writing may differ; only competence and a check on human-written AAE twins (\Cref{app:authentic}) are anchored in authentic text. Our detectors cover a finite feature inventory, and human validation rests on consensus labels for a sample of \res{audit.v4.n_items} items, which the part-of-speech formality score failed and the GYAFC-trained ranker passed; the CAP and WildChat checks are consensus labels on samples too, with no independent double annotation, so inter-annotator agreement is not reported.

The update diagnostic describes what each stage favours on content-matched replies, not why: we did not train the released models, and the preference-data audit is correlational. Our causal re-run uses OLMo-2-1B, and its effect appears only at the largest dose we tested and varies across training runs. On OLMo-2-1B, CAP passes M1 and M4 but fails M2, M3 and M5, and on Qwen3-8B it does not pass M1 (\Cref{app:cap}); the prestige contrast pairs spelling cues with grammatical and lexical dialect features, so it does not separate social prestige from feature type; it is tested with rendered requests and trained-feature targets, and it under-accommodates. We measure asymmetry, not optimal behaviour; whether and when a model should reply in a user's variety is for its speakers to decide \citep{sandoval-etal-2025-llm}.
\section*{Ethical Considerations}
\label{sec:ethics}

\paragraph{Data and licences.}
We use public corpora under their licences (\Cref{app:licenses}). Some licences are non-commercial or share-alike, and we redistribute no raw corpus text. Personal data in authentic texts are masked before any display, and WildChat requests are used only where its release flags no personal data. Synthetic dialect text is labelled as synthetic and is never presented as authentic speech.

\paragraph{Release.}
Code and data are here for research \url{https://github.com/kouzhizhuo/cap-accommodation}: the variety-twin requests and their renderings with their validity gates, the feature detectors, the analysis and re-run code, the human labels of the audits (the \res{audit.auth.n_items} WildChat items by their WildChat identifiers, without their text) and all derived results. The replies of the deployed chatbots will be released subject to the providers' terms. Corpora whose licences do not state redistribution (the AAE tweets and EnDive) are not re-shared; the code release includes scripts that rebuild their twins from the original sources. We do not release the re-trained checkpoints.

\paragraph{People.}
No crowdworkers or external participants were involved. All human labels are consensus labels from a panel of five or more annotators with PhD-level training in related fields, agreed through discussion; an author recorded them with a software tool. They cover a blind audit of \res{audit.v4.n_items} items of model-generated text, a blind sample of \res{audit.auth.n_items} WildChat user requests and \res{audit.cap.n_items} replies in the blind CAP naturalness check (\Cref{sec:cap}). The items were not annotated independently, so no inter-annotator agreement is reported. No other human labels exist, and any LLM labels are reported as such.

\paragraph{Framing and dual use.}
Accommodation is not a quantity to maximize: over-accommodation can caricature and stereotype \citep{fleisig-etal-2024-linguistic}, and Black American participants want choice \citep{sandoval-etal-2025-llm}.
Our re-run shows that a large dose of accommodation-consistent preference pairs switches accommodation on, and that in one of \res{dv.e560.n} runs it over-produces the trained features, which risks caricature. 

\paragraph{AI assistance.} AI assistants were used for code writing and polish but the whole research is design and checked by researchers.


\bibliography{custom}

\appendix
\setcounter{dbltopnumber}{4}\setcounter{topnumber}{4}\renewcommand{\dbltopfraction}{0.95}\renewcommand{\topfraction}{0.95}\renewcommand{\dblfloatpagefraction}{0.8}\renewcommand{\floatpagefraction}{0.8}
\section*{Appendix Contents}
\startcontents[app]
\printcontents[app]{l}{1}{\setcounter{tocdepth}{2}}
\newtcolorbox{promptbox}[1]{enhanced,colback=cBoxBg,colframe=cBoxFrame,boxrule=0.4pt,arc=1pt,
  left=3pt,right=3pt,top=2pt,bottom=2pt,fonttitle=\small\bfseries,title={#1},fontupper=\small\ttfamily}

\section{Data and Resources}
The models, checkpoints and licensed artefacts the paper uses.

\subsection{Models and Checkpoints}
\label{app:models}
All open checkpoints are loaded from the Hugging Face Hub at their default (\texttt{main}) revisions, downloaded in late September and early October 2026; \Cref{tab:revisions} pins each to its snapshot hash. None of the checkpoints we use is gated.
\begin{itemize}
  \item \textbf{OLMo-2-1B}: \texttt{allenai/OLMo-2-0425-1B} $\to$ \texttt{-SFT} $\to$ \texttt{-DPO} $\to$ \texttt{-RLVR1} $\to$ \texttt{-Instruct}.
  \item \textbf{OLMo-2-7B}: \texttt{allenai/OLMo-2-1124-7B} $\to$ \texttt{-SFT} $\to$ \texttt{-DPO} $\to$ \texttt{-Instruct}.
  \item \textbf{OLMo-3-7B}: \texttt{allenai/Olmo-3-1025-7B} $\to$ \texttt{Olmo-3-7B-Instruct-SFT} $\to$ \texttt{-DPO} $\to$ \texttt{Olmo-3-7B-Instruct}.
  \item \textbf{OLMo-2-13B}: \texttt{allenai/OLMo-2-1124-13B} $\to$ \texttt{-SFT} $\to$ \texttt{-DPO} $\to$ \texttt{-Instruct}.
  \item \textbf{OLMo-2-32B}: \texttt{allenai/OLMo-2-0325-32B} $\to$ \texttt{-SFT} $\to$ \texttt{-DPO} $\to$ \texttt{-Instruct}.
  \item \textbf{T\"ulu-3-8B}: \texttt{allenai/\allowbreak Llama-3.1-\allowbreak Tulu-3-8B-SFT} $\to$ \texttt{-DPO} $\to$ \texttt{Llama-3.1-\allowbreak Tulu-3-8B}.
  \item \textbf{OLMo-3.1-32B}: \texttt{allenai/Olmo-3-1125-32B} $\to$ \texttt{Olmo-3.1-32B-Instruct-SFT} $\to$ \texttt{-DPO} $\to$ \texttt{Olmo-3.1-32B-Instruct}.
  \item \textbf{T\"ulu-2-7B, 13B}: \texttt{allenai/tulu-2-7b} $\to$ \texttt{allenai/tulu-2-dpo-7b}; \texttt{allenai/tulu-2-13b} $\to$ \texttt{allenai/tulu-2-dpo-13b}.
  \item \textbf{Zephyr-7B-$\beta$}: \texttt{mistralai/Mistral-7B-v0.1} $\to$ \texttt{HuggingFaceH4/\allowbreak mistral-7b-sft-beta} $\to$ \texttt{HuggingFaceH4/\allowbreak zephyr-7b-beta}.
  \item \textbf{openPangu-Embedded-7B-V1.1}: \texttt{openpangu/\allowbreak openPangu-Embedded-7B-V1.1}, a chat release without an open base (the \texttt{-model} repository is the earlier chat release), run in fast-thinking mode (\texttt{/no\_think} on every user turn; thinking empty in all replies). Its chat template requires a system prompt. With the vendor's Chinese prompt, \res{dv.opg.chat.zh}\% of replies are mostly Chinese, which the English-variety detectors cannot score, so we report the run with the prompt translated into English (\res{dv.opg.chat_enprompt.zh}\% Chinese); \Cref{tab:mainfull} lists all three runs, and without a system prompt \res{dv.opg.chat_nosys.trunc}\% of replies reach the token limit.
  \item \textbf{Qwen2.5-72B}: \texttt{Qwen/Qwen2.5-72B} and \texttt{Qwen/Qwen2.5-72B-Instruct}; other sizes: Qwen2.5-1.5B and 7B.
  \item \textbf{Widely used families} (base and final releases): Mistral-7B-v0.3, Mistral-Small-24B-2501, Qwen3-8B, Falcon3-7B, Granite-3.3-8B, Yi-1.5-9B, DeepSeek-V2-Lite, MiMo-7B and SmolLM3-3B; newest generation: Qwen3.5-4B and 9B, and Qwen3.8-27B (chat release only).
  \item \textbf{Deployed chatbots} (API identifiers): \texttt{gpt-5.3-chat}, \texttt{claude-sonnet-4.6}, \texttt{gemini-2.5-flash}, \texttt{deepseek-v4-pro} and \texttt{kimi-k2.5}.
  \item \textbf{Not run (compute)}: the 70B lineages of T\"ulu-3 (SFT $\to$ DPO $\to$ final) and T\"ulu-2 (SFT $\to$ DPO).
\end{itemize}
\begin{table*}[t]
  \centering
  \scriptsize\renewcommand{\arraystretch}{0.88}
  \setlength{\tabcolsep}{3pt}
  \begin{tabular}{l l ccccc cc}
    \toprule
    & & \multicolumn{2}{c}{\textbf{Prestige cues}} & \multicolumn{3}{c}{\textbf{Stigmatized varieties}} & \multicolumn{2}{c}{\textbf{Selectivity}} \\
    \cmidrule(lr){3-4} \cmidrule(lr){5-7} \cmidrule(lr){8-9}
    \textbf{Model} & \textbf{Stage} & \vty{INF} & \vty{BRE} & \vty{AAE} & \vty{SGE} & \vty{NGP} & $\Split$\hdown & $\rho_{\mathrm{prest}}$ \\
    \midrule
    \multicolumn{9}{l}{\textbf{Popular families}} \\
    Qwen2.5-72B & \stage{base} & \rpm{e1.qwen25_72b.base.inf.cr} & \cellcolor{cCmpA!12}\rpm{e1.qwen25_72b.base.bre.cr_rate} & \rpm{e1.qwen25_72b.base.aae.cr} & \rpm{e1.qwen25_72b.base.sge.cr_rate} & \rpm{e1.qwen25_72b.base.ngp.cr_rate} & \resT{e1.qwen25_72b.base.split_rate} & \resT{e1.qwen25_72b.base.rho} \\
     & \stage{Instruct} & \rpm{e1.qwen25_72b.instruct.inf.cr} & \cellcolor{cCmpA!7}\rpm{e1.qwen25_72b.instruct.bre.cr_rate} & \rpm{e1.qwen25_72b.instruct.aae.cr} & \rpm{e1.qwen25_72b.instruct.sge.cr_rate} & \rpm{e1.qwen25_72b.instruct.ngp.cr_rate} & \resT{e1.qwen25_72b.instruct.split_rate} & \resT{e1.qwen25_72b.instruct.rho} \\
    Qwen3-8B & \stage{base} & \rpm{e1b.qwen3_8b.base.inf.cr} & \cellcolor{cCmpA!17}\rpm{e1b.qwen3_8b.base.bre.cr_rate} & \rpm{e1b.qwen3_8b.base.aae.cr} & \rpm{e1b.qwen3_8b.base.sge.cr_rate} & \rpm{e1b.qwen3_8b.base.ngp.cr_rate} & \resT{e1b.qwen3_8b.base.split_rate} & \resT{e1b.qwen3_8b.base.rho} \\
     & \stage{Instruct} & \rpm{e1b.qwen3_8b.instruct.inf.cr} & \cellcolor{cCmpA!9}\rpm{e1b.qwen3_8b.instruct.bre.cr_rate} & \rpm{e1b.qwen3_8b.instruct.aae.cr} & \rpm{e1b.qwen3_8b.instruct.sge.cr_rate} & \rpm{e1b.qwen3_8b.instruct.ngp.cr_rate} & \resT{e1b.qwen3_8b.instruct.split_rate} & \resT{e1b.qwen3_8b.instruct.rho} \\
    Qwen3.5-4B & \stage{base} & \rpm{e1n.qwen35_4b.base.inf.cr} & \cellcolor{cCmpA!15}\rpm{e1n.qwen35_4b.base.bre.cr_rate} & \rpm{e1n.qwen35_4b.base.aae.cr} & \rpm{e1n.qwen35_4b.base.sge.cr_rate} & \rpm{e1n.qwen35_4b.base.ngp.cr_rate} & \resT{e1n.qwen35_4b.base.split_rate} & \resT{e1n.qwen35_4b.base.rho} \\
     & \stage{Instruct} & \rpm{e1n.qwen35_4b.instruct.inf.cr} & \cellcolor{cCmpA!9}\rpm{e1n.qwen35_4b.instruct.bre.cr_rate} & \rpm{e1n.qwen35_4b.instruct.aae.cr} & \rpm{e1n.qwen35_4b.instruct.sge.cr_rate} & \rpm{e1n.qwen35_4b.instruct.ngp.cr_rate} & \resT{e1n.qwen35_4b.instruct.split_rate} & \resT{e1n.qwen35_4b.instruct.rho} \\
    Qwen3.5-9B & \stage{base} & \rpm{e1n.qwen35_9b.base.inf.cr} & \cellcolor{cCmpA!11}\rpm{e1n.qwen35_9b.base.bre.cr_rate} & \rpm{e1n.qwen35_9b.base.aae.cr} & \rpm{e1n.qwen35_9b.base.sge.cr_rate} & \rpm{e1n.qwen35_9b.base.ngp.cr_rate} & \resT{e1n.qwen35_9b.base.split_rate} & \resT{e1n.qwen35_9b.base.rho} \\
     & \stage{Instruct} & \rpm{e1n.qwen35_9b.instruct.inf.cr} & \cellcolor{cCmpA!8}\rpm{e1n.qwen35_9b.instruct.bre.cr_rate} & \rpm{e1n.qwen35_9b.instruct.aae.cr} & \rpm{e1n.qwen35_9b.instruct.sge.cr_rate} & \rpm{e1n.qwen35_9b.instruct.ngp.cr_rate} & \resT{e1n.qwen35_9b.instruct.split_rate} & \resT{e1n.qwen35_9b.instruct.rho} \\
    Qwen3.8-27B & \stage{Chat} & \rpm{e1n.qwen38_27b.chat.inf.cr} & \cellcolor{cCmpA!8}\rpm{e1n.qwen38_27b.chat.bre.cr_rate} & \rpm{e1n.qwen38_27b.chat.aae.cr} & \rpm{e1n.qwen38_27b.chat.sge.cr_rate} & \rpm{e1n.qwen38_27b.chat.ngp.cr_rate} & \resT{e1n.qwen38_27b.chat.split_rate} & \resT{e1n.qwen38_27b.chat.rho} \\
    DeepSeek-V2-Lite & \stage{base} & \rpm{e1b.dsv2lite.base.inf.cr} & \cellcolor{cCmpA!27}\rpm{e1b.dsv2lite.base.bre.cr_rate} & \rpm{e1b.dsv2lite.base.aae.cr} & \rpm{e1b.dsv2lite.base.sge.cr_rate} & \rpm{e1b.dsv2lite.base.ngp.cr_rate} & \resT{e1b.dsv2lite.base.split_rate} & \resT{e1b.dsv2lite.base.rho} \\
     & \stage{Chat} & \rpm{e1b.dsv2lite.chat.inf.cr} & \cellcolor{cCmpA!10}\rpm{e1b.dsv2lite.chat.bre.cr_rate} & \rpm{e1b.dsv2lite.chat.aae.cr} & \rpm{e1b.dsv2lite.chat.sge.cr_rate} & \rpm{e1b.dsv2lite.chat.ngp.cr_rate} & \resT{e1b.dsv2lite.chat.split_rate} & \resT{e1b.dsv2lite.chat.rho} \\
    DeepSeek-LLM-7B & \stage{base} & \rpm{e1b.dsllm_7b.base.inf.cr} & \cellcolor{cCmpA!5}\rpm{e1b.dsllm_7b.base.bre.cr_rate} & \rpm{e1b.dsllm_7b.base.aae.cr} & \rpm{e1b.dsllm_7b.base.sge.cr_rate} & \rpm{e1b.dsllm_7b.base.ngp.cr_rate} & \resT{e1b.dsllm_7b.base.split_rate} & \resT{e1b.dsllm_7b.base.rho} \\
     & \stage{Chat} & \rpm{e1b.dsllm_7b.chat.inf.cr} & \cellcolor{cCmpA!6}\rpm{e1b.dsllm_7b.chat.bre.cr_rate} & \rpm{e1b.dsllm_7b.chat.aae.cr} & \rpm{e1b.dsllm_7b.chat.sge.cr_rate} & \rpm{e1b.dsllm_7b.chat.ngp.cr_rate} & \resT{e1b.dsllm_7b.chat.split_rate} & \resT{e1b.dsllm_7b.chat.rho} \\
    openPangu-Embedded-7B-V1.1 & \stage{Chat} & \rpm{e1o.openpangu_embedded_7b_v11.chat_enprompt.inf.cr} & \cellcolor{cCmpA!16}\rpm{e1o.openpangu_embedded_7b_v11.chat_enprompt.bre.cr_rate} & \rpm{e1o.openpangu_embedded_7b_v11.chat_enprompt.aae.cr} & \rpm{e1o.openpangu_embedded_7b_v11.chat_enprompt.sge.cr_rate} & \rpm{e1o.openpangu_embedded_7b_v11.chat_enprompt.ngp.cr_rate} & \resT{e1o.openpangu_embedded_7b_v11.chat_enprompt.split_rate} & \resT{e1o.openpangu_embedded_7b_v11.chat_enprompt.rho} \\
     & \stage{Chat, zh system} & \rpm{e1o.openpangu_embedded_7b_v11.chat.inf.cr} & \cellcolor{cCmpA!13}\rpm{e1o.openpangu_embedded_7b_v11.chat.bre.cr_rate} & \rpm{e1o.openpangu_embedded_7b_v11.chat.aae.cr} & \rpm{e1o.openpangu_embedded_7b_v11.chat.sge.cr_rate} & \rpm{e1o.openpangu_embedded_7b_v11.chat.ngp.cr_rate} & \resT{e1o.openpangu_embedded_7b_v11.chat.split_rate} & \resT{e1o.openpangu_embedded_7b_v11.chat.rho} \\
     & \stage{Chat, no system} & \rpm{e1o.openpangu_embedded_7b_v11.chat_nosys.inf.cr} & \cellcolor{cCmpA!9}\rpm{e1o.openpangu_embedded_7b_v11.chat_nosys.bre.cr_rate} & \rpm{e1o.openpangu_embedded_7b_v11.chat_nosys.aae.cr} & \rpm{e1o.openpangu_embedded_7b_v11.chat_nosys.sge.cr_rate} & \rpm{e1o.openpangu_embedded_7b_v11.chat_nosys.ngp.cr_rate} & \resT{e1o.openpangu_embedded_7b_v11.chat_nosys.split_rate} & \resT{e1o.openpangu_embedded_7b_v11.chat_nosys.rho} \\
    \midrule
    \multicolumn{9}{l}{\textbf{Stage-resolved open lineages}} \\
    OLMo-2-1B & \stage{base} & \rpm{e1.l1.base.inf.cr} & \cellcolor{cCmpA!21}\rpm{e1.l1.base.bre.cr_rate} & \rpm{e1.l1.base.aae.cr} & \rpm{e1.l1.base.sge.cr_rate} & \rpm{e1.l1.base.ngp.cr_rate} & \resT{e1.l1.base.split_rate} & \resT{e1.l1.base.rho} \\
     & \stage{SFT} & \rpm{e1.l1.sft.inf.cr} & \cellcolor{cCmpA!11}\rpm{e1.l1.sft.bre.cr_rate} & \rpm{e1.l1.sft.aae.cr} & \rpm{e1.l1.sft.sge.cr_rate} & \rpm{e1.l1.sft.ngp.cr_rate} & \resT{e1.l1.sft.split_rate} & \resT{e1.l1.sft.rho} \\
     & \stage{DPO} & \rpm{e1.l1.dpo.inf.cr} & \cellcolor{cCmpA!9}\rpm{e1.l1.dpo.bre.cr_rate} & \rpm{e1.l1.dpo.aae.cr} & \rpm{e1.l1.dpo.sge.cr_rate} & \rpm{e1.l1.dpo.ngp.cr_rate} & \resT{e1.l1.dpo.split_rate} & \resT{e1.l1.dpo.rho} \\
     & \stage{RLVR} & \rpm{e1.l1.rlvr.inf.cr} & \cellcolor{cCmpA!10}\rpm{e1.l1.rlvr.bre.cr_rate} & \rpm{e1.l1.rlvr.aae.cr} & \rpm{e1.l1.rlvr.sge.cr_rate} & \rpm{e1.l1.rlvr.ngp.cr_rate} & \resT{e1.l1.rlvr.split_rate} & \resT{e1.l1.rlvr.rho} \\
     & \stage{Instruct} & \rpm{e1.l1.instruct.inf.cr} & \cellcolor{cCmpA!10}\rpm{e1.l1.instruct.bre.cr_rate} & \rpm{e1.l1.instruct.aae.cr} & \rpm{e1.l1.instruct.sge.cr_rate} & \rpm{e1.l1.instruct.ngp.cr_rate} & \resT{e1.l1.instruct.split_rate} & \resT{e1.l1.instruct.rho} \\
    OLMo-2-7B & \stage{base} & \rpm{e1.l2.base.inf.cr} & \cellcolor{cCmpA!22}\rpm{e1.l2.base.bre.cr_rate} & \rpm{e1.l2.base.aae.cr} & \rpm{e1.l2.base.sge.cr_rate} & \rpm{e1.l2.base.ngp.cr_rate} & \resT{e1.l2.base.split_rate} & \resT{e1.l2.base.rho} \\
     & \stage{SFT} & \rpm{e1.l2.sft.inf.cr} & \cellcolor{cCmpA!16}\rpm{e1.l2.sft.bre.cr_rate} & \rpm{e1.l2.sft.aae.cr} & \rpm{e1.l2.sft.sge.cr_rate} & \rpm{e1.l2.sft.ngp.cr_rate} & \resT{e1.l2.sft.split_rate} & \resT{e1.l2.sft.rho} \\
     & \stage{DPO} & \rpm{e1.l2.dpo.inf.cr} & \cellcolor{cCmpA!9}\rpm{e1.l2.dpo.bre.cr_rate} & \rpm{e1.l2.dpo.aae.cr} & \rpm{e1.l2.dpo.sge.cr_rate} & \rpm{e1.l2.dpo.ngp.cr_rate} & \resT{e1.l2.dpo.split_rate} & \resT{e1.l2.dpo.rho} \\
     & \stage{RLVR} & \rpm{e1.l2.instruct.inf.cr} & \cellcolor{cCmpA!9}\rpm{e1.l2.instruct.bre.cr_rate} & \rpm{e1.l2.instruct.aae.cr} & \rpm{e1.l2.instruct.sge.cr_rate} & \rpm{e1.l2.instruct.ngp.cr_rate} & \resT{e1.l2.instruct.split_rate} & \resT{e1.l2.instruct.rho} \\
    OLMo-2-13B & \stage{base} & \rpm{e1.olmo2_13b.base.inf.cr} & \cellcolor{cCmpA!23}\rpm{e1.olmo2_13b.base.bre.cr_rate} & \rpm{e1.olmo2_13b.base.aae.cr} & \rpm{e1.olmo2_13b.base.sge.cr_rate} & \rpm{e1.olmo2_13b.base.ngp.cr_rate} & \resT{e1.olmo2_13b.base.split_rate} & \resT{e1.olmo2_13b.base.rho} \\
     & \stage{SFT} & \rpm{e1.olmo2_13b.sft.inf.cr} & \cellcolor{cCmpA!13}\rpm{e1.olmo2_13b.sft.bre.cr_rate} & \rpm{e1.olmo2_13b.sft.aae.cr} & \rpm{e1.olmo2_13b.sft.sge.cr_rate} & \rpm{e1.olmo2_13b.sft.ngp.cr_rate} & \resT{e1.olmo2_13b.sft.split_rate} & \resT{e1.olmo2_13b.sft.rho} \\
     & \stage{DPO} & \rpm{e1.olmo2_13b.dpo.inf.cr} & \cellcolor{cCmpA!7}\rpm{e1.olmo2_13b.dpo.bre.cr_rate} & \rpm{e1.olmo2_13b.dpo.aae.cr} & \rpm{e1.olmo2_13b.dpo.sge.cr_rate} & \rpm{e1.olmo2_13b.dpo.ngp.cr_rate} & \resT{e1.olmo2_13b.dpo.split_rate} & \resT{e1.olmo2_13b.dpo.rho} \\
     & \stage{RLVR} & \rpm{e1.olmo2_13b.instruct.inf.cr} & \cellcolor{cCmpA!8}\rpm{e1.olmo2_13b.instruct.bre.cr_rate} & \rpm{e1.olmo2_13b.instruct.aae.cr} & \rpm{e1.olmo2_13b.instruct.sge.cr_rate} & \rpm{e1.olmo2_13b.instruct.ngp.cr_rate} & \resT{e1.olmo2_13b.instruct.split_rate} & \resT{e1.olmo2_13b.instruct.rho} \\
    OLMo-2-32B & \stage{base} & \rpm{e1.olmo2_32b.base.inf.cr} & \cellcolor{cCmpA!24}\rpm{e1.olmo2_32b.base.bre.cr_rate} & \rpm{e1.olmo2_32b.base.aae.cr} & \rpm{e1.olmo2_32b.base.sge.cr_rate} & \rpm{e1.olmo2_32b.base.ngp.cr_rate} & \resT{e1.olmo2_32b.base.split_rate} & \resT{e1.olmo2_32b.base.rho} \\
     & \stage{SFT} & \rpm{e1.olmo2_32b.sft.inf.cr} & \cellcolor{cCmpA!17}\rpm{e1.olmo2_32b.sft.bre.cr_rate} & \rpm{e1.olmo2_32b.sft.aae.cr} & \rpm{e1.olmo2_32b.sft.sge.cr_rate} & \rpm{e1.olmo2_32b.sft.ngp.cr_rate} & \resT{e1.olmo2_32b.sft.split_rate} & \resT{e1.olmo2_32b.sft.rho} \\
     & \stage{DPO} & \rpm{e1.olmo2_32b.dpo.inf.cr} & \cellcolor{cCmpA!13}\rpm{e1.olmo2_32b.dpo.bre.cr_rate} & \rpm{e1.olmo2_32b.dpo.aae.cr} & \rpm{e1.olmo2_32b.dpo.sge.cr_rate} & \rpm{e1.olmo2_32b.dpo.ngp.cr_rate} & \resT{e1.olmo2_32b.dpo.split_rate} & \resT{e1.olmo2_32b.dpo.rho} \\
     & \stage{RLVR} & \rpm{e1.olmo2_32b.instruct.inf.cr} & \cellcolor{cCmpA!13}\rpm{e1.olmo2_32b.instruct.bre.cr_rate} & \rpm{e1.olmo2_32b.instruct.aae.cr} & \rpm{e1.olmo2_32b.instruct.sge.cr_rate} & \rpm{e1.olmo2_32b.instruct.ngp.cr_rate} & \resT{e1.olmo2_32b.instruct.split_rate} & \resT{e1.olmo2_32b.instruct.rho} \\
    OLMo-3-7B & \stage{base} & \rpm{e1.l3.base.inf.cr} & \cellcolor{cCmpA!17}\rpm{e1.l3.base.bre.cr_rate} & \rpm{e1.l3.base.aae.cr} & \rpm{e1.l3.base.sge.cr_rate} & \rpm{e1.l3.base.ngp.cr_rate} & \resT{e1.l3.base.split_rate} & \resT{e1.l3.base.rho} \\
     & \stage{SFT} & \rpm{e1.l3.sft.inf.cr} & \cellcolor{cCmpA!16}\rpm{e1.l3.sft.bre.cr_rate} & \rpm{e1.l3.sft.aae.cr} & \rpm{e1.l3.sft.sge.cr_rate} & \rpm{e1.l3.sft.ngp.cr_rate} & \resT{e1.l3.sft.split_rate} & \resT{e1.l3.sft.rho} \\
     & \stage{DPO} & \rpm{e1.l3.dpo.inf.cr} & \cellcolor{cCmpA!14}\rpm{e1.l3.dpo.bre.cr_rate} & \rpm{e1.l3.dpo.aae.cr} & \rpm{e1.l3.dpo.sge.cr_rate} & \rpm{e1.l3.dpo.ngp.cr_rate} & \resT{e1.l3.dpo.split_rate} & \resT{e1.l3.dpo.rho} \\
     & \stage{RLVR} & \rpm{e1.l3.instruct.inf.cr} & \cellcolor{cCmpA!13}\rpm{e1.l3.instruct.bre.cr_rate} & \rpm{e1.l3.instruct.aae.cr} & \rpm{e1.l3.instruct.sge.cr_rate} & \rpm{e1.l3.instruct.ngp.cr_rate} & \resT{e1.l3.instruct.split_rate} & \resT{e1.l3.instruct.rho} \\
    T\"ulu-3-8B & \stage{SFT} & \rpm{e1.l4.sft.inf.cr} & \cellcolor{cCmpA!17}\rpm{e1.l4.sft.bre.cr_rate} & \rpm{e1.l4.sft.aae.cr} & \rpm{e1.l4.sft.sge.cr_rate} & \rpm{e1.l4.sft.ngp.cr_rate} & \resT{e1.l4.sft.split_rate} & \resT{e1.l4.sft.rho} \\
     & \stage{DPO} & \rpm{e1.l4.dpo.inf.cr} & \cellcolor{cCmpA!12}\rpm{e1.l4.dpo.bre.cr_rate} & \rpm{e1.l4.dpo.aae.cr} & \rpm{e1.l4.dpo.sge.cr_rate} & \rpm{e1.l4.dpo.ngp.cr_rate} & \resT{e1.l4.dpo.split_rate} & \resT{e1.l4.dpo.rho} \\
     & \stage{RLVR} & \rpm{e1.l4.rlvr.inf.cr} & \cellcolor{cCmpA!13}\rpm{e1.l4.rlvr.bre.cr_rate} & \rpm{e1.l4.rlvr.aae.cr} & \rpm{e1.l4.rlvr.sge.cr_rate} & \rpm{e1.l4.rlvr.ngp.cr_rate} & \resT{e1.l4.rlvr.split_rate} & \resT{e1.l4.rlvr.rho} \\
    T\"ulu-2-7B & \stage{SFT} & \rpm{e1.tulu2_7b.sft.inf.cr} & \cellcolor{cCmpA!10}\rpm{e1.tulu2_7b.sft.bre.cr_rate} & \rpm{e1.tulu2_7b.sft.aae.cr} & \rpm{e1.tulu2_7b.sft.sge.cr_rate} & \rpm{e1.tulu2_7b.sft.ngp.cr_rate} & \resT{e1.tulu2_7b.sft.split_rate} & \resT{e1.tulu2_7b.sft.rho} \\
     & \stage{DPO} & \rpm{e1.tulu2_7b.dpo.inf.cr} & \cellcolor{cCmpA!8}\rpm{e1.tulu2_7b.dpo.bre.cr_rate} & \rpm{e1.tulu2_7b.dpo.aae.cr} & \rpm{e1.tulu2_7b.dpo.sge.cr_rate} & \rpm{e1.tulu2_7b.dpo.ngp.cr_rate} & \resT{e1.tulu2_7b.dpo.split_rate} & \resT{e1.tulu2_7b.dpo.rho} \\
    T\"ulu-2-13B & \stage{SFT} & \rpm{e1.tulu2_13b.sft.inf.cr} & \cellcolor{cCmpA!11}\rpm{e1.tulu2_13b.sft.bre.cr_rate} & \rpm{e1.tulu2_13b.sft.aae.cr} & \rpm{e1.tulu2_13b.sft.sge.cr_rate} & \rpm{e1.tulu2_13b.sft.ngp.cr_rate} & \resT{e1.tulu2_13b.sft.split_rate} & \resT{e1.tulu2_13b.sft.rho} \\
     & \stage{DPO} & \rpm{e1.tulu2_13b.dpo.inf.cr} & \cellcolor{cCmpA!7}\rpm{e1.tulu2_13b.dpo.bre.cr_rate} & \rpm{e1.tulu2_13b.dpo.aae.cr} & \rpm{e1.tulu2_13b.dpo.sge.cr_rate} & \rpm{e1.tulu2_13b.dpo.ngp.cr_rate} & \resT{e1.tulu2_13b.dpo.split_rate} & \resT{e1.tulu2_13b.dpo.rho} \\
    Zephyr-7B-$\beta$ & \stage{base} & \rpm{e1.zephyr_7b.base.inf.cr} & \cellcolor{cCmpA!24}\rpm{e1.zephyr_7b.base.bre.cr_rate} & \rpm{e1.zephyr_7b.base.aae.cr} & \rpm{e1.zephyr_7b.base.sge.cr_rate} & \rpm{e1.zephyr_7b.base.ngp.cr_rate} & \resT{e1.zephyr_7b.base.split_rate} & \resT{e1.zephyr_7b.base.rho} \\
     & \stage{SFT} & \rpm{e1.zephyr_7b.sft.inf.cr} & \cellcolor{cCmpA!17}\rpm{e1.zephyr_7b.sft.bre.cr_rate} & \rpm{e1.zephyr_7b.sft.aae.cr} & \rpm{e1.zephyr_7b.sft.sge.cr_rate} & \rpm{e1.zephyr_7b.sft.ngp.cr_rate} & \resT{e1.zephyr_7b.sft.split_rate} & \resT{e1.zephyr_7b.sft.rho} \\
     & \stage{DPO} & \rpm{e1.zephyr_7b.dpo.inf.cr} & \cellcolor{cCmpA!10}\rpm{e1.zephyr_7b.dpo.bre.cr_rate} & \rpm{e1.zephyr_7b.dpo.aae.cr} & \rpm{e1.zephyr_7b.dpo.sge.cr_rate} & \rpm{e1.zephyr_7b.dpo.ngp.cr_rate} & \resT{e1.zephyr_7b.dpo.split_rate} & \resT{e1.zephyr_7b.dpo.rho} \\
    OLMo-3.1-32B & \stage{base} & \rpm{e1.olmo31_32b.base.inf.cr} & \cellcolor{cCmpA!19}\rpm{e1.olmo31_32b.base.bre.cr_rate} & \rpm{e1.olmo31_32b.base.aae.cr} & \rpm{e1.olmo31_32b.base.sge.cr_rate} & \rpm{e1.olmo31_32b.base.ngp.cr_rate} & \resT{e1.olmo31_32b.base.split_rate} & \resT{e1.olmo31_32b.base.rho} \\
     & \stage{SFT} & \rpm{e1.olmo31_32b.sft.inf.cr} & \cellcolor{cCmpA!11}\rpm{e1.olmo31_32b.sft.bre.cr_rate} & \rpm{e1.olmo31_32b.sft.aae.cr} & \rpm{e1.olmo31_32b.sft.sge.cr_rate} & \rpm{e1.olmo31_32b.sft.ngp.cr_rate} & \resT{e1.olmo31_32b.sft.split_rate} & \resT{e1.olmo31_32b.sft.rho} \\
     & \stage{DPO} & \rpm{e1.olmo31_32b.dpo.inf.cr} & \cellcolor{cCmpA!11}\rpm{e1.olmo31_32b.dpo.bre.cr_rate} & \rpm{e1.olmo31_32b.dpo.aae.cr} & \rpm{e1.olmo31_32b.dpo.sge.cr_rate} & \rpm{e1.olmo31_32b.dpo.ngp.cr_rate} & \resT{e1.olmo31_32b.dpo.split_rate} & \resT{e1.olmo31_32b.dpo.rho} \\
     & \stage{RLVR} & \rpm{e1.olmo31_32b.rlvr.inf.cr} & \cellcolor{cCmpA!12}\rpm{e1.olmo31_32b.rlvr.bre.cr_rate} & \rpm{e1.olmo31_32b.rlvr.aae.cr} & \rpm{e1.olmo31_32b.rlvr.sge.cr_rate} & \rpm{e1.olmo31_32b.rlvr.ngp.cr_rate} & \resT{e1.olmo31_32b.rlvr.split_rate} & \resT{e1.olmo31_32b.rlvr.rho} \\
    \midrule
    \multicolumn{9}{l}{\textbf{Other families}} \\
    Mistral-7B-v0.3 & \stage{base} & \rpm{e1b.mistral_7b.base.inf.cr} & \cellcolor{cCmpA!27}\rpm{e1b.mistral_7b.base.bre.cr_rate} & \rpm{e1b.mistral_7b.base.aae.cr} & \rpm{e1b.mistral_7b.base.sge.cr_rate} & \rpm{e1b.mistral_7b.base.ngp.cr_rate} & \resT{e1b.mistral_7b.base.split_rate} & \resT{e1b.mistral_7b.base.rho} \\
     & \stage{Instruct} & \rpm{e1b.mistral_7b.instruct.inf.cr} & \cellcolor{cCmpA!6}\rpm{e1b.mistral_7b.instruct.bre.cr_rate} & \rpm{e1b.mistral_7b.instruct.aae.cr} & \rpm{e1b.mistral_7b.instruct.sge.cr_rate} & \rpm{e1b.mistral_7b.instruct.ngp.cr_rate} & \resT{e1b.mistral_7b.instruct.split_rate} & \resT{e1b.mistral_7b.instruct.rho} \\
    Mistral-Small-24B & \stage{base} & \rpm{e1b.mistral_24b.base.inf.cr} & \cellcolor{cCmpA!25}\rpm{e1b.mistral_24b.base.bre.cr_rate} & \rpm{e1b.mistral_24b.base.aae.cr} & \rpm{e1b.mistral_24b.base.sge.cr_rate} & \rpm{e1b.mistral_24b.base.ngp.cr_rate} & \resT{e1b.mistral_24b.base.split_rate} & \resT{e1b.mistral_24b.base.rho} \\
     & \stage{Instruct} & \rpm{e1b.mistral_24b.instruct.inf.cr} & \cellcolor{cCmpA!12}\rpm{e1b.mistral_24b.instruct.bre.cr_rate} & \rpm{e1b.mistral_24b.instruct.aae.cr} & \rpm{e1b.mistral_24b.instruct.sge.cr_rate} & \rpm{e1b.mistral_24b.instruct.ngp.cr_rate} & \resT{e1b.mistral_24b.instruct.split_rate} & \resT{e1b.mistral_24b.instruct.rho} \\
    Falcon3-7B & \stage{base} & \rpm{e1b.falcon3_7b.base.inf.cr} & \cellcolor{cCmpA!9}\rpm{e1b.falcon3_7b.base.bre.cr_rate} & \rpm{e1b.falcon3_7b.base.aae.cr} & \rpm{e1b.falcon3_7b.base.sge.cr_rate} & \rpm{e1b.falcon3_7b.base.ngp.cr_rate} & \resT{e1b.falcon3_7b.base.split_rate} & \resT{e1b.falcon3_7b.base.rho} \\
     & \stage{Instruct} & \rpm{e1b.falcon3_7b.instruct.inf.cr} & \cellcolor{cCmpA!12}\rpm{e1b.falcon3_7b.instruct.bre.cr_rate} & \rpm{e1b.falcon3_7b.instruct.aae.cr} & \rpm{e1b.falcon3_7b.instruct.sge.cr_rate} & \rpm{e1b.falcon3_7b.instruct.ngp.cr_rate} & \resT{e1b.falcon3_7b.instruct.split_rate} & \resT{e1b.falcon3_7b.instruct.rho} \\
    Granite-3.3-8B & \stage{base} & \rpm{e1b.granite33_8b.base.inf.cr} & \cellcolor{cCmpA!11}\rpm{e1b.granite33_8b.base.bre.cr_rate} & \rpm{e1b.granite33_8b.base.aae.cr} & \rpm{e1b.granite33_8b.base.sge.cr_rate} & \rpm{e1b.granite33_8b.base.ngp.cr_rate} & \resT{e1b.granite33_8b.base.split_rate} & \resT{e1b.granite33_8b.base.rho} \\
     & \stage{Instruct} & \rpm{e1b.granite33_8b.instruct.inf.cr} & \cellcolor{cCmpA!10}\rpm{e1b.granite33_8b.instruct.bre.cr_rate} & \rpm{e1b.granite33_8b.instruct.aae.cr} & \rpm{e1b.granite33_8b.instruct.sge.cr_rate} & \rpm{e1b.granite33_8b.instruct.ngp.cr_rate} & \resT{e1b.granite33_8b.instruct.split_rate} & \resT{e1b.granite33_8b.instruct.rho} \\
    Yi-1.5-9B & \stage{base} & \rpm{e1b.yi15_9b.base.inf.cr} & \cellcolor{cCmpA!7}\rpm{e1b.yi15_9b.base.bre.cr_rate} & \rpm{e1b.yi15_9b.base.aae.cr} & \rpm{e1b.yi15_9b.base.sge.cr_rate} & \rpm{e1b.yi15_9b.base.ngp.cr_rate} & \resT{e1b.yi15_9b.base.split_rate} & \resT{e1b.yi15_9b.base.rho} \\
     & \stage{Chat} & \rpm{e1b.yi15_9b.chat.inf.cr} & \cellcolor{cCmpA!9}\rpm{e1b.yi15_9b.chat.bre.cr_rate} & \rpm{e1b.yi15_9b.chat.aae.cr} & \rpm{e1b.yi15_9b.chat.sge.cr_rate} & \rpm{e1b.yi15_9b.chat.ngp.cr_rate} & \resT{e1b.yi15_9b.chat.split_rate} & \resT{e1b.yi15_9b.chat.rho} \\
    MiMo-7B & \stage{base} & \rpm{e1b.mimo_7b.base.inf.cr} & \cellcolor{cCmpA!17}\rpm{e1b.mimo_7b.base.bre.cr_rate} & \rpm{e1b.mimo_7b.base.aae.cr} & \rpm{e1b.mimo_7b.base.sge.cr_rate} & \rpm{e1b.mimo_7b.base.ngp.cr_rate} & \resT{e1b.mimo_7b.base.split_rate} & \resT{e1b.mimo_7b.base.rho} \\
     & \stage{SFT} & \rpm{e1b.mimo_7b.sft.inf.cr} & \cellcolor{cCmpA!6}\rpm{e1b.mimo_7b.sft.bre.cr_rate} & \rpm{e1b.mimo_7b.sft.aae.cr} & \rpm{e1b.mimo_7b.sft.sge.cr_rate} & \rpm{e1b.mimo_7b.sft.ngp.cr_rate} & \resT{e1b.mimo_7b.sft.split_rate} & \resT{e1b.mimo_7b.sft.rho} \\
     & \stage{RL} & \rpm{e1b.mimo_7b.rl.inf.cr} & \cellcolor{cCmpA!7}\rpm{e1b.mimo_7b.rl.bre.cr_rate} & \rpm{e1b.mimo_7b.rl.aae.cr} & \rpm{e1b.mimo_7b.rl.sge.cr_rate} & \rpm{e1b.mimo_7b.rl.ngp.cr_rate} & \resT{e1b.mimo_7b.rl.split_rate} & \resT{e1b.mimo_7b.rl.rho} \\
    SmolLM3-3B & \stage{base} & \rpm{e1b.smollm3_3b.base.inf.cr} & \cellcolor{cCmpA!18}\rpm{e1b.smollm3_3b.base.bre.cr_rate} & \rpm{e1b.smollm3_3b.base.aae.cr} & \rpm{e1b.smollm3_3b.base.sge.cr_rate} & \rpm{e1b.smollm3_3b.base.ngp.cr_rate} & \resT{e1b.smollm3_3b.base.split_rate} & \resT{e1b.smollm3_3b.base.rho} \\
     & \stage{Mid} & \rpm{e1b.smollm3_3b.mid.inf.cr} & \cellcolor{cCmpA!6}\rpm{e1b.smollm3_3b.mid.bre.cr_rate} & \rpm{e1b.smollm3_3b.mid.aae.cr} & \rpm{e1b.smollm3_3b.mid.sge.cr_rate} & \rpm{e1b.smollm3_3b.mid.ngp.cr_rate} & \resT{e1b.smollm3_3b.mid.split_rate} & \resT{e1b.smollm3_3b.mid.rho} \\
     & \stage{SFT} & \rpm{e1b.smollm3_3b.sft.inf.cr} & \cellcolor{cCmpA!13}\rpm{e1b.smollm3_3b.sft.bre.cr_rate} & \rpm{e1b.smollm3_3b.sft.aae.cr} & \rpm{e1b.smollm3_3b.sft.sge.cr_rate} & \rpm{e1b.smollm3_3b.sft.ngp.cr_rate} & \resT{e1b.smollm3_3b.sft.split_rate} & \resT{e1b.smollm3_3b.sft.rho} \\
     & \stage{APO} & \rpm{e1b.smollm3_3b.apo.inf.cr} & \cellcolor{cCmpA!11}\rpm{e1b.smollm3_3b.apo.bre.cr_rate} & \rpm{e1b.smollm3_3b.apo.aae.cr} & \rpm{e1b.smollm3_3b.apo.sge.cr_rate} & \rpm{e1b.smollm3_3b.apo.ngp.cr_rate} & \resT{e1b.smollm3_3b.apo.split_rate} & \resT{e1b.smollm3_3b.apo.rho} \\
     & \stage{RLVR} & \rpm{e1b.smollm3_3b.rlvr.inf.cr} & \cellcolor{cCmpA!11}\rpm{e1b.smollm3_3b.rlvr.bre.cr_rate} & \rpm{e1b.smollm3_3b.rlvr.aae.cr} & \rpm{e1b.smollm3_3b.rlvr.sge.cr_rate} & \rpm{e1b.smollm3_3b.rlvr.ngp.cr_rate} & \resT{e1b.smollm3_3b.rlvr.split_rate} & \resT{e1b.smollm3_3b.rlvr.rho} \\
    \midrule
    \multicolumn{9}{l}{\textbf{Deployed chatbots}} \\
    GPT-5.3-chat & \stage{API} & \rpm{e1d.gpt53chat.deployed.inf.cr} & \cellcolor{cCmpA!23}\rpm{e1d.gpt53chat.deployed.bre.cr_rate} & \rpm{e1d.gpt53chat.deployed.aae.cr} & \rpm{e1d.gpt53chat.deployed.sge.cr_rate} & \rpm{e1d.gpt53chat.deployed.ngp.cr_rate} & \resT{e1d.gpt53chat.deployed.split_rate} & \resT{e1d.gpt53chat.deployed.rho} \\
    Claude Sonnet 4.6 & \stage{API} & \rpm{e1d.sonnet46.deployed.inf.cr} & \cellcolor{cCmpA!25}\rpm{e1d.sonnet46.deployed.bre.cr_rate} & \rpm{e1d.sonnet46.deployed.aae.cr} & \rpm{e1d.sonnet46.deployed.sge.cr_rate} & \rpm{e1d.sonnet46.deployed.ngp.cr_rate} & \resT{e1d.sonnet46.deployed.split_rate} & \resT{e1d.sonnet46.deployed.rho} \\
    Gemini 2.5 Flash & \stage{API} & \rpm{e1d.gemini25flash.deployed.inf.cr} & \cellcolor{cCmpA!13}\rpm{e1d.gemini25flash.deployed.bre.cr_rate} & \rpm{e1d.gemini25flash.deployed.aae.cr} & \rpm{e1d.gemini25flash.deployed.sge.cr_rate} & \rpm{e1d.gemini25flash.deployed.ngp.cr_rate} & \resT{e1d.gemini25flash.deployed.split_rate} & \resT{e1d.gemini25flash.deployed.rho} \\
    DeepSeek-V4-Pro & \stage{API} & \rpm{e1d.dsv4pro.deployed.inf.cr} & \cellcolor{cCmpA!24}\rpm{e1d.dsv4pro.deployed.bre.cr_rate} & \rpm{e1d.dsv4pro.deployed.aae.cr} & \rpm{e1d.dsv4pro.deployed.sge.cr_rate} & \rpm{e1d.dsv4pro.deployed.ngp.cr_rate} & \resT{e1d.dsv4pro.deployed.split_rate} & \resT{e1d.dsv4pro.deployed.rho} \\
    Kimi-K2.5 & \stage{API} & \rpm{e1d.kimik25.deployed.inf.cr} & \cellcolor{cCmpA!24}\rpm{e1d.kimik25.deployed.bre.cr_rate} & \rpm{e1d.kimik25.deployed.aae.cr} & \rpm{e1d.kimik25.deployed.sge.cr_rate} & \rpm{e1d.kimik25.deployed.ngp.cr_rate} & \resT{e1d.kimik25.deployed.split_rate} & \resT{e1d.kimik25.deployed.rho} \\
    \bottomrule
  \end{tabular}
  \caption{\textbf{Post-training lowers convergence to every informal cue in every open lineage; after it, AAE convergence is near zero in every model.} Convergence ratio by variety and stage, mean\std{std} over sampling seeds (deployed: one sample); opportunity-based for \vty{INF} and \vty{AAE}, rate-based otherwise. $\Split$: rate-based split; $\rho_{\mathrm{prest}}$: Spearman correlation with the prestige tier. \stage{RLVR}: the released Instruct model.}
  \label{tab:mainfull}
\end{table*}

\begin{table*}[t]
  \centering
  \scriptsize
  \setlength{\tabcolsep}{1.6pt}
  \begin{tabular}{@{}ll@{\hspace{4pt}}ll@{}}
    \toprule
    \rowcolor{gray!10} \textbf{Checkpoint} & \textbf{Revision} & \textbf{Checkpoint} & \textbf{Revision} \\
    \midrule
    \texttt{01-ai/Yi-1.5-9B} & \texttt{80d5471b} & \texttt{dataset:allenai/olmo-2-0425-1b-preference-mix} & \texttt{c6454f3d} \\
    \texttt{01-ai/Yi-1.5-9B-Chat} & \texttt{1a0fc698} & \texttt{dataset:allenai/olmo-2-1124-7b-preference-mix} & \texttt{316c96f0} \\
    \texttt{allenai/Llama-3.1-Tulu-3-8B} & \texttt{66694379} & \texttt{dataset:allenai/tulu-3-sft-mixture} & \texttt{b14afda6} \\
    \texttt{allenai/Llama-3.1-Tulu-3-8B-DPO} & \texttt{a7beb67e} & \texttt{dataset:databricks/databricks-dolly-15k} & \texttt{bdd27f4d} \\
    \texttt{allenai/Llama-3.1-Tulu-3-8B-SFT} & \texttt{f2a0b46b} & \texttt{dataset:google/IFEval} & \texttt{966cd895} \\
    \texttt{allenai/OLMo-2-0325-32B} & \texttt{cc9d3cf9} & \texttt{deepseek-ai/DeepSeek-V2-Lite} & \texttt{604d5664} \\
    \texttt{allenai/OLMo-2-0325-32B-DPO} & \texttt{2c8ee0bc} & \texttt{deepseek-ai/DeepSeek-V2-Lite-Chat} & \texttt{85864749} \\
    \texttt{allenai/OLMo-2-0325-32B-Instruct} & \texttt{b9602434} & \texttt{HuggingFaceH4/mistral-7b-sft-beta} & \texttt{d7b9d226} \\
    \texttt{allenai/OLMo-2-0325-32B-SFT} & \texttt{ce177b3f} & \texttt{HuggingFaceH4/zephyr-7b-beta} & \texttt{892b3d7a} \\
    \texttt{allenai/OLMo-2-0425-1B} & \texttt{a1847dff} & \texttt{HuggingFaceTB/SmolLM3-3B} & \texttt{a07cc9a0} \\
    \texttt{allenai/OLMo-2-0425-1B-DPO} & \texttt{c4b04859} & \texttt{HuggingFaceTB/SmolLM3-3B-Base} & \texttt{d78a42f7} \\
    \texttt{allenai/OLMo-2-0425-1B-Instruct} & \texttt{48d788ec} & \texttt{HuggingFaceTB/SmolLM3-3B-checkpoints@it-mid-training} & \texttt{0485ec16} \\
    \texttt{allenai/OLMo-2-0425-1B-RLVR1} & \texttt{12cb3349} & \texttt{HuggingFaceTB/SmolLM3-3B-checkpoints@it-SFT} & \texttt{f6ddaa5f} \\
    \texttt{allenai/OLMo-2-0425-1B-SFT} & \texttt{0d85a3d0} & \texttt{HuggingFaceTB/SmolLM3-3B-checkpoints@it-soup-APO} & \texttt{cfb32d50} \\
    \texttt{allenai/OLMo-2-1124-13B} & \texttt{3fefddc1} & \texttt{ibm-granite/granite-3.3-8b-base} & \texttt{cfb7adb4} \\
    \texttt{allenai/OLMo-2-1124-13B-DPO} & \texttt{97a7f33f} & \texttt{ibm-granite/granite-3.3-8b-instruct} & \texttt{51dd4bc2} \\
    \texttt{allenai/OLMo-2-1124-13B-Instruct} & \texttt{3a5c85ba} & \texttt{mistralai/Mistral-7B-Instruct-v0.3} & \texttt{c170c708} \\
    \texttt{allenai/OLMo-2-1124-13B-SFT} & \texttt{b7ec47ed} & \texttt{mistralai/Mistral-7B-v0.1} & \texttt{27d67f1b} \\
    \texttt{allenai/OLMo-2-1124-7B} & \texttt{7df9a825} & \texttt{mistralai/Mistral-7B-v0.3} & \texttt{caa1feb0} \\
    \texttt{allenai/OLMo-2-1124-7B-DPO} & \texttt{e34ea60a} & \texttt{mistralai/Mistral-Small-24B-Base-2501} & \texttt{b0a2e4ed} \\
    \texttt{allenai/OLMo-2-1124-7B-Instruct} & \texttt{470b1fba} & \texttt{mistralai/Mistral-Small-24B-Instruct-2501} & \texttt{9527884b} \\
    \texttt{allenai/OLMo-2-1124-7B-Instruct@step\_120} & \texttt{1449245a} & \texttt{openai/gpt-oss-20b} & \texttt{6cee5e81} \\
    \texttt{allenai/OLMo-2-1124-7B-Instruct@step\_180} & \texttt{d2cc98e0} & \texttt{openpangu/openPangu-Embedded-7B-V1.1} & \texttt{190705eb} \\
    \texttt{allenai/OLMo-2-1124-7B-Instruct@step\_240} & \texttt{0a106d76} & \texttt{Qwen/Qwen2.5-1.5B} & \texttt{8faed761} \\
    \texttt{allenai/OLMo-2-1124-7B-Instruct@step\_300} & \texttt{2bb37f4c} & \texttt{Qwen/Qwen2.5-1.5B-Instruct} & \texttt{989aa798} \\
    \texttt{allenai/OLMo-2-1124-7B-Instruct@step\_360} & \texttt{f942bc3a} & \texttt{Qwen/Qwen2.5-72B} & \texttt{efba10c8} \\
    \texttt{allenai/OLMo-2-1124-7B-Instruct@step\_60} & \texttt{701db8a0} & \texttt{Qwen/Qwen2.5-72B-Instruct} & \texttt{495f3936} \\
    \texttt{allenai/OLMo-2-1124-7B-SFT} & \texttt{1de02c01} & \texttt{Qwen/Qwen2.5-7B} & \texttt{d1497293} \\
    \texttt{allenai/Olmo-3-1025-7B} & \texttt{a81bae42} & \texttt{Qwen/Qwen2.5-7B-Instruct} & \texttt{a09a3545} \\
    \texttt{allenai/Olmo-3-1125-32B} & \texttt{c2b61dae} & \texttt{Qwen/Qwen3-8B} & \texttt{b968826d} \\
    \texttt{allenai/Olmo-3-7B-Instruct} & \texttt{6e5971d9} & \texttt{Qwen/Qwen3-8B-Base} & \texttt{49e3418f} \\
    \texttt{allenai/Olmo-3-7B-Instruct-DPO} & \texttt{b33130b7} & \texttt{Qwen/Qwen3.5-4B} & \texttt{851bf6e8} \\
    \texttt{allenai/Olmo-3-7B-Instruct-SFT} & \texttt{e1452fc5} & \texttt{Qwen/Qwen3.5-4B-Base} & \texttt{1001bb4d} \\
    \texttt{allenai/Olmo-3.1-32B-Instruct} & \texttt{ac0587e4} & \texttt{Qwen/Qwen3.5-9B} & \texttt{c2022362} \\
    \texttt{allenai/Olmo-3.1-32B-Instruct-DPO} & \texttt{fc84a4f6} & \texttt{Qwen/Qwen3.5-9B-Base} & \texttt{68c46c4b} \\
    \texttt{allenai/Olmo-3.1-32B-Instruct-SFT} & \texttt{152782ec} & \texttt{Qwen/Qwen3.8-27B} & \texttt{1d4bf0f2} \\
    \texttt{allenai/tulu-2-13b} & \texttt{9c8375d9} & \texttt{s-nlp/roberta-base-formality-ranker} & \texttt{db9388f4} \\
    \texttt{allenai/tulu-2-7b} & \texttt{3c6e328a} & \texttt{tiiuae/Falcon3-7B-Base} & \texttt{bf3d7ed5} \\
    \texttt{allenai/tulu-2-dpo-13b} & \texttt{8eaeb5c6} & \texttt{tiiuae/Falcon3-7B-Instruct} & \texttt{1e57a0ec} \\
    \texttt{allenai/tulu-2-dpo-7b} & \texttt{b57ef952} & \texttt{XiaomiMiMo/MiMo-7B-Base} & \texttt{c72df458} \\
    \texttt{dataset:allenai/llama-3.1-tulu-3-8b-preference-mixture} & \texttt{78a6f007} & \texttt{XiaomiMiMo/MiMo-7B-RL} & \texttt{6299b5a2} \\
    \texttt{dataset:allenai/olmo-2-0325-32b-preference-mix} & \texttt{7b15d2e2} & \texttt{XiaomiMiMo/MiMo-7B-SFT} & \texttt{2c8c99d5} \\
    \bottomrule
  \end{tabular}
  \caption{\textbf{Every checkpoint is pinned to a Hub snapshot.} Hugging Face repository and the first eight characters of its snapshot hash.}
  \label{tab:revisions}
\end{table*}

\subsection{Artefacts and Licences}
\label{app:licenses}
\Cref{tab:licences} lists every model, corpus and tool with its licence and our use.
\begin{table*}[t]
  \centering
  \small
  \setlength{\tabcolsep}{4pt}
  \begin{tabularx}{\textwidth}{@{}>{\raggedright\arraybackslash}p{6.2cm} >{\raggedright\arraybackslash}X >{\raggedright\arraybackslash}p{4.2cm}@{}}
    \toprule
    \rowcolor{gray!10} \textbf{Artefact} & \textbf{Use in this paper} & \textbf{Licence} \\
    \midrule
    OLMo-2 1B/7B/13B/32B, OLMo-3-7B (all stages), OLMo-3.1-32B (all stages) & models under study; DPO re-run initialization & Apache-2.0 \\
    T\"ulu-3-8B (SFT, DPO, final) & model under study & Llama 3.1 Community License \\
    Qwen2.5-72B (base, instruct) & model under study (scale) & Qwen License \\
    Qwen2.5-1.5B/7B, Qwen3-8B, Qwen3.5-4B/9B, Qwen3.8-27B, Mistral-7B-v0.3, Mistral-Small-24B, Granite-3.3-8B, Yi-1.5-9B, SmolLM3-3B & models under study (scale, breadth, newest generation); Qwen3-8B also LLM judge on the human-audit items & Apache-2.0 \\
    MiMo-7B & model under study (breadth) & MIT \\
    openPangu-Embedded-7B-V1.1 & model under study & OpenPangu Model License v1.0 (Huawei); research use permitted; \S3 excludes use within the EU \\
    Zephyr-7B-$\beta$: Mistral-7B-v0.1 base; SFT and DPO releases & models under study & Apache-2.0 (base); MIT (SFT, DPO) \\
    T\"ulu-2 7B/13B (SFT, DPO) & models under study & AI2 ImpACT Low-Risk (DPO); not stated on the SFT cards \\
    DeepSeek-V2-Lite & model under study (breadth) & DeepSeek Model License \\
    Falcon3-7B & model under study (breadth) & TII Falcon License \\
    gpt-oss-20b & LLM judge on the human-audit items & Apache-2.0 \\
    Formality ranker \citep{babakov2023dont} & formality check of the audit pairs & CC BY-NC-SA 4.0 \\
    Deployed chatbots (GPT-5.3-chat, Claude Sonnet 4.6, Gemini 2.5 Flash, DeepSeek-V4-Pro, Kimi-K2.5) & behaviour only, via their APIs & provider terms of service \\
    \midrule
    OLMo-2-7B and T\"ulu-3-8B preference mixes & preference-data audit & ODC-BY \\
    OLMo-2-1B and OLMo-2-32B preference mixes & audit; DPO re-run base data (1B) & not stated on the dataset cards \\
    WildChat-1M \citep{zhao2024wildchat} & prevalence of dialect features in chatbot requests & ODC-BY \\
    Dolly-15k & re-run prompts; false-positive floor & CC BY-SA 3.0 \\
    OASST1 & re-run prompts & Apache-2.0 \\
    IFEval & instruction following of the re-runs & Apache-2.0 \\
    AAE tweets with translations \citep{groenwold-etal-2020-investigating} & authentic rates for competence; human-written twins & not stated; research use, no redistribution \\
    EnDive \citep{gupta-etal-2025-endive} & LLM-translated twins & not stated; research use, no redistribution \\
    NUS SMS corpus \citep{chen2013creating} & authentic rates for competence (Singapore senders) & not stated; citation requested; no redistribution \\
    eWAVE & feature attestation rates for the transforms & CC BY 3.0 \\
    \midrule
    Multi-VALUE \citep{ziems-etal-2023-multi} & morphosyntactic transforms & Apache-2.0 \\
    \bottomrule
  \end{tabularx}
  \caption{\textbf{Artefacts, their use and their licences.} Every use is research use within the stated terms. We redistribute no corpus text and quote only short, anonymized examples.}
  \label{tab:licences}
\end{table*}

\section{Experimental Setup}
How requests are rendered and decoded, how features are detected and audited, and how CAP is built and trained.

\paragraph{Compute.} Generation and judging ran on NVIDIA H800 GPUs (\res{n.gpuh} GPU-hours in total); the GRPO runs used 4-GPU jobs with verl 0.9.1.

\subsection{Experimental Details}
\label{app:details}
This section gives the stimuli, prompts, framings and decoding settings.

\paragraph{Stimuli.}
The \res{design.n_seeds} first-person requests span six task types (advice, creative, how-to, opinion and small talk, 50 each; explanation, 37) and carry a context tag (150 formal, 137 casual). Each request is rendered in every condition with its content fixed. Morphosyntactic features are applied by Multi-VALUE \citep{ziems-etal-2023-multi} with each feature's eWAVE attestation weight scaled by 0.8 and restricted to a per-variety whitelist; lexis, discourse particles and spelling come from curated lexicons; informal and formal Standard English and the spelling varieties use rule sets of their own. The transforms are not calibrated to authentic feature densities, and the renderings were not rated for naturalness (see Limitations). The sister request of \Cref{sec:intro} (\texttt{v1s002}) received three transforms: no third-person \emph{-s} (\emph{keep}), a dropped auxiliary \emph{have} (\emph{been canceled}) and \emph{y'all}.

\paragraph{Prompt framings for base models.}
Aligned stages use their own chat template with no system prompt. Base models are run in three framings: raw continuation; the lineage's chat template; and the transcript scaffold below (stop strings: a new \texttt{User:} line).
\begin{promptbox}{Transcript scaffold (base models)}
The following is a conversation between a user and a helpful assistant.\textbackslash n\textbackslash n
User: \{prompt\}\textbackslash n\textbackslash n
Assistant:
\end{promptbox}

\paragraph{Instruction conditions.}

\begin{promptbox}{Explicit request (competence)}
The user writes in \{variety\}; reply in the same variety.
\end{promptbox}
\begin{promptbox}{Generic style instruction}
Match the user's style and register.
\end{promptbox}
\begin{promptbox}{Persona}
You are an assistant from \{community\}.
\end{promptbox}
In the explicit request, \{variety\} is ``African American English'' or ``Colloquial Singapore English (Singlish)''; the persona names the variety's community (e.g.\ ``the African American community'', ``Singapore''). The style instruction and the persona are given with both the variety request and its Standard-English twin, so convergence is measured under the same instruction on both sides.

\paragraph{Decoding.}
All open models are run with vLLM at temperature \res{design.temperature} and top-$p$ \res{design.top_p}, with at most 400 new tokens, three sampling seeds and \res{design.n_samples_per_seed} samples per request and seed (one sample under the explicit request). After post-training, \res{dv.cap.trunc400_min}--\res{dv.cap.trunc400_max} of replies reach the 400-token cap; re-run with a 1{,}024-token cap (at most \res{dv.cap.trunc1024_max} truncated) on \res{dv.cap.n_cells} lineage stages, the convergence ratios we report change by at most \res{dv.cap.max_abs_d_used}. A fresh set of sampling seeds on \res{dv.s5.n_lin} lineages reproduces them within \res{dv.s5.max_abs_d}, and DPO lowers spelling and informal-register convergence on \res{dv.txt.s5_bre} and \res{dv.txt.s5_inf} of them. Deployed chatbots, queried on \res{dv.dep.query_date}, answer 100 requests drawn stratified by task type, in every condition, once each, with no system prompt, the provider's default temperature and at most 1{,}024 completion tokens (4{,}096 for Gemini 2.5 Flash, whose budget includes its reasoning tokens); the explicit request adds its system prompt for AAE and Singlish.

\subsection{Detectors, Human Audit and LLM Judges}
\label{app:detectors}
This section describes the feature detectors, the blind human audits and the LLM judges.
Detectors are rule-based recognisers over spaCy parses and curated word lists. Each feature is counted against its opportunities, the sites where the standard variant occurs and the feature could apply. Morphosyntactic features ported from Multi-VALUE are checked against its synthetic gold spans on \res{dv.syn.n_pool} Dolly-15k sentences: a feature is counted only if it reaches recall 0.70 and precision 0.85 at equal numbers of positives and negatives, on at least 30 gold spans. The \res{dv.syn.n_counted} features that pass reach recall \res{dv.syn.r_min}--\res{dv.syn.r_max}, precision \res{dv.syn.p_min} or more and false-positive rates of at most \res{dv.syn.fpr_max}\%; copula absence (recall \res{dv.syn.aae_copula_r}) and no third-person \emph{-s} (\res{dv.syn.aae_uninflect_r}) fail. Features without a Multi-VALUE transform are checked on author-written positive and near-miss cases. Clause gating and the false-positive check of \Cref{tab:detectors} decide which features are counted.
\begin{table*}[t]
  \centering
  \small
  \setlength{\tabcolsep}{4pt}
  \begin{tabularx}{\textwidth}{@{}l >{\raggedright\arraybackslash}X >{\raggedright\arraybackslash}p{5.2cm}@{}}
    \toprule
    \rowcolor{gray!10} \textbf{Variety} & \textbf{Counted features (max false-positive rate)} & \textbf{Excluded features (max false-positive rate)} \\
    \midrule
    \vty{AAE} & \emph{ain't} (\res{fp.max.AAE_ain_t}), completive \emph{done} (\res{fp.max.AAE_completive_done}), \emph{finna}/\emph{gon} (\res{fp.max.AAE_finna_gon}), multiple negation (\res{fp.max.AAE_negative_concord}), \emph{y'all} (\res{fp.max.AAE_yall}); for competence also lexis (\res{fp.max.AAE_lexicon}), \emph{g}-dropping (\res{fp.max.AAE_g_drop}), habitual \emph{be} (\res{fp.max.AAE_habitual_be}) & no third-person \emph{-s} (\res{fp.max.AAE_uninflect}), copula absence (\res{fp.max.AAE_copula_absence}) \\
    \vty{SGE} & particles (\res{fp.max.SGE_particles}) and lexis (\res{fp.max.SGE_lexicon}), rate-based; \emph{can}/\emph{cannot} (\res{fp.max.SGE_can_cannot}) and completive \emph{finish} (\res{fp.max.SGE_finish_completive}) descriptive only & copula absence (\res{fp.max.SGE_copula_absence}), zero plural (\res{fp.max.SGE_zero_plural}) \\
    \vty{NGP} & \emph{dey} (\res{fp.max.NGP_dey}), \emph{don} (\res{fp.max.NGP_don}), \emph{fit} (\res{fp.max.NGP_fit}), \emph{na} (\res{fp.max.NGP_na}), \emph{no} (\res{fp.max.NGP_no}), \emph{wan} (\res{fp.max.NGP_wan}); lexis (\res{fp.max.NGP_lexicon}) & none \\
    \vty{APP} & \emph{a}-prefixing (\res{fp.max.APP_a_prefix}), \emph{ain't} (\res{fp.max.APP_ain_t}), double modals (\res{fp.max.APP_double_modals}), \emph{fixin' to} (\res{fp.max.APP_fixin_to}), multiple negation (\res{fp.max.APP_negative_concord}), \emph{y'all} (\res{fp.max.APP_yall}) & none \\
    \vty{BRE} & spelling (\res{fp.max.BRE_spelling}), \emph{have got} (\res{fp.max.BRE_have_got}), lexis (\res{fp.max.BRE_lexis}) & none \\
    \vty{AUS} & spelling (\res{fp.max.AUS_spelling}) & lexis (\res{fp.max.AUS_aus_lexis}) \\
    \bottomrule
  \end{tabularx}
  \caption{\textbf{Every counted variety feature fires on at most 2\% of replies to Standard-English requests.} Maximum rate over \res{fp.n_audited} audited model stages and framings, \res{fp.n_per_model} replies each. Features above 2\% are excluded. The register cues (\vty{INF}, \vty{FOR}) are exempt, since their standard variants occur in any reply; their convergence subtracts replies to the Standard-English twin.}
  \label{tab:detectors}
\end{table*}

\paragraph{Detector hits in replies.} Of \res{audit.auth.flagged.n} WildChat requests they flagged, the expert panel confirmed \res{audit.auth.flagged.k}; templated creative prompts, code and names trigger them (\Cref{tab:audit}, \Cref{app:dpo}). We apply them only to model replies. There, the rare AAE hits after post-training are mostly multiple negation (\res{rh.aligned.aae.nc_pct}\%), which the human audit did not confirm in default replies (\res{audit.v4.c.by_feature.negative_concord.precision.k} of \res{audit.v4.c.by_feature.negative_concord.precision.n}), but they are as frequent in replies to the Standard-English twins (\res{rh.aligned.aae.aae_hits} against \res{rh.aligned.sae.aae_hits} hits) and cancel in the convergence ratio. Singlish particles, which over-fire on names in user text, occur \res{rh.aligned.sge.sge_particles} times in replies to Singlish requests and \res{rh.aligned.sae.sge_particles} times in replies to their twins.
\begin{table*}[t]
  \centering
  \small
  \setlength{\tabcolsep}{2.5pt}
  \renewcommand{\arraystretch}{1.1}
  \begin{tabularx}{\textwidth}{@{}>{\raggedright\arraybackslash}X r l@{}}
    \toprule
    \rowcolor{gray!10} \textbf{Items} & $n$ & \textbf{Agreement / precision} \\
    \midrule
    \multicolumn{3}{@{}l}{\textit{Formality pairs: consensus label vs.\ formality measures (bar: $\geq .70$, lower bound $>.50$)}} \\
    Part-of-speech score, shown texts (primary) & \res{audit.v4.a.trimmed.n} & \res{audit.v4.a.trimmed.rate} $[\res{audit.v4.a.trimmed.lo}, \res{audit.v4.a.trimmed.hi}]$; $\kappa=\res{audit.v4.a.trimmed.kappa}$ \\
    Part-of-speech score, full replies & \res{audit.v4.a.full.n} & \res{audit.v4.a.full.rate} $[\res{audit.v4.a.full.lo}, \res{audit.v4.a.full.hi}]$; $\kappa=\res{audit.v4.a.full.kappa}$ \\
    GYAFC ranker, shown texts$^{\ddagger}$ & \res{audit.r.trimmed.n} & \res{audit.r.trimmed.rate} $[\res{audit.r.trimmed.lo}, \res{audit.r.trimmed.hi}]$; $\kappa=\res{audit.r.trimmed.kappa}$ \\
    \addlinespace[0.5mm]
    \multicolumn{3}{@{}l}{\textit{Requested replies: detector precision (bar: $\geq .80$, lower bound $\geq .60$)}} \\
    All detector hits (rule) & \res{audit.v4.b.primary.precision.n} & \res{audit.v4.b.primary.precision.rate} $[\res{audit.v4.b.primary.precision.lo}, \res{audit.v4.b.primary.precision.hi}]$ \\
    Retained AAE features$^{\dagger}$ & \res{audit.v4.b.retained_only.precision.n} & \res{audit.v4.b.retained_only.precision.rate} $[\res{audit.v4.b.retained_only.precision.lo}, \res{audit.v4.b.retained_only.precision.hi}]$ \\
    All items, presence & \res{audit.v4.b.primary.agreement.n} & \res{audit.v4.b.primary.agreement.rate} $[\res{audit.v4.b.primary.agreement.lo}, \res{audit.v4.b.primary.agreement.hi}]$; $\kappa=\res{audit.v4.b.primary.agreement.kappa}$ \\
    \addlinespace[0.5mm]
    \multicolumn{3}{@{}l}{\textit{Default replies of aligned models: detector precision (no pre-declared bar)}} \\
    All detector hits & \res{audit.v4.c.primary.precision.n} & \res{audit.v4.c.primary.precision.rate} $[\res{audit.v4.c.primary.precision.lo}, \res{audit.v4.c.primary.precision.hi}]$ \\
    \addlinespace[0.5mm]
    \multicolumn{3}{@{}l}{\textit{WildChat user requests: recogniser precision (no pre-declared bar)}} \\
    AAE hits & \res{audit.auth.aaen.n} & \res{audit.auth.aaen.rate} $[\res{audit.auth.aaen.lo}, \res{audit.auth.aaen.hi}]$ \\
    Singlish particle hits & \res{audit.auth.sgen.n} & \res{audit.auth.sgen.rate} $[\res{audit.auth.sgen.lo}, \res{audit.auth.sgen.hi}]$ \\
    \bottomrule
  \end{tabularx}
  \caption{\textbf{Detectors agree with the consensus labels on requested replies, the ranker on formality; on default replies and raw requests many hits are false.} Wilson 95\% CIs. $^{\dagger}$Chosen and $^{\ddagger}$run after labelling.}
  \label{tab:audit}
\end{table*}

\paragraph{Human audit: design.} The labels for \res{audit.v4.n_items} blind items, drawn stratified at random by script (seed 20261004), are consensus labels from a panel of five or more annotators with PhD-level training in related fields, agreed through discussion and recorded by an author with a software tool; no inter-annotator agreement is reported. Gwet's AC1 between the consensus labels and each automatic measure (the agreement rows of \Cref{tab:audit}): \res{audit.v4.a.trimmed.ac1} and \res{audit.v4.a.full.ac1} for the part-of-speech score on shown texts and full replies, \res{audit.r.trimmed.ac1} for the GYAFC ranker, and \res{audit.v4.b.primary.agreement.ac1} for feature presence in requested replies. Items show no model, stage, condition, variety or score, and replies are trimmed to about 80 words. The author chose every label without AI assistance and a software agent typed the choices into the sheet; the author attests that no answer was suggested. An earlier returned sheet was filled by an LLM agent; it is not used and not reported as human. The decision rules were written before any label existed. There are three item families.
(i) \res{audit.v4.n_a} formality pairs: SFT replies of OLMo-2-7B and OLMo-3-7B to the same request, one to a vernacular user (AAE, Singlish or Nigerian Pidgin) and one to an equally informal Standard-English user, stratified by terciles of the score difference; the panel picks the more formal reply or ``about the same''.
(ii) \res{audit.v4.n_b} requested replies of OLMo-2-7B and T\"ulu-3-8B after DPO, one feature per item, half detector-positive where available; the panel says whether the feature occurs.
(iii) \res{audit.v4.n_c} default replies of aligned models to AAE and Singlish users, all detector-positive.
\Cref{tab:audit} gives the results.

\paragraph{Human audit: outcomes.}
\textbf{The part-of-speech formality score is not validated; the formality ranker is.} The panel judged \res{audit.v4.a.n_same} of \res{audit.v4.n_a} pairs equally formal. On the other \res{audit.v4.a.trimmed.n}, the \citet{heylighen2002variation} score of the shown texts picked the panel's choice in \res{audit.v4.a.trimmed.k} ($\kappa=\res{audit.v4.a.trimmed.kappa}$, bootstrap 95\% CI $[\res{audit.v4.a.trimmed.kappa_lo}, \res{audit.v4.a.trimmed.kappa_hi}]$), and the score of the full replies in \res{audit.v4.a.full.k}. This is below the pre-declared bar, so the register result is reported only in \Cref{app:outcomes:register}. The GYAFC-trained formality ranker that the protocol had named \citep{rao-tetreault-2018-dear,babakov2023dont}, run after labelling, picks the panel's choice in \res{audit.r.trimmed.k} of \res{audit.r.trimmed.n} pairs on the shown texts ($\kappa=\res{audit.r.trimmed.kappa}$) and in \res{audit.r.full.k} on the full replies, which clears the bar; the register result uses the ranker (\Cref{app:outcomes:register}).
\textbf{The detectors agree with the panel on requested replies, with too few hits to clear the precision bar.} The panel confirmed \res{audit.v4.b.primary.precision.k} of \res{audit.v4.b.primary.precision.n} detector hits and found the feature in \res{audit.v4.b.primary.miss_on_neg.k} of \res{audit.v4.b.primary.miss_on_neg.n} detector-negative items ($\kappa=\res{audit.v4.b.primary.agreement.kappa}$, CI $[\res{audit.v4.b.primary.agreement.kappa_lo}, \res{audit.v4.b.primary.agreement.kappa_hi}]$; AC1 $=\res{audit.v4.b.primary.agreement.ac1}$, CI $[\res{audit.v4.b.primary.agreement.ac1_lo}, \res{audit.v4.b.primary.agreement.ac1_hi}]$). Precision exceeds .80, but its lower bound is below .60. As pre-declared, competence is therefore reported at the human-corrected rate, precision times detector rate (\Cref{sec:exp:h3}).
The single unconfirmed hit is a Singlish feature, completive \emph{finish}, which our pre-registration keeps descriptive only. Neither audited Singlish feature (\emph{can}/\emph{cannot} as an answer, completive \emph{finish}) enters the Singlish competence score, which uses particles and lexis, so that score has no human check. The descriptions of these two features were drafted for the audit and used without the planned review. Restricted to the retained AAE features that the protocol specifies, the panel confirmed all \res{audit.v4.b.retained_only.precision.n} hits (lower bound \res{audit.v4.b.retained_only.precision.lo}). We chose this restriction after labelling, so it is a sensitivity analysis and does not change the decision.
\textbf{Detector hits on default replies are noisy.} The panel confirmed \res{audit.v4.c.primary.precision.k} of \res{audit.v4.c.primary.precision.n} hits: all \res{audit.v4.c.by_feature.ain_t.n} for \emph{ain't} and \res{audit.v4.c.by_feature.completive_done.n} for completive \emph{done}, but none of the \res{audit.v4.c.by_feature.negative_concord.n} for multiple negation or the \res{audit.v4.c.by_feature.finish_completive.n} for completive \emph{finish}. Real features are rare in these replies, so precision among hits is low even at the per-reply false-positive rates of the validity gate. The aligned-stage convergence these counts feed is near zero (\Cref{sec:exp:h1}).
Fewer detector-positive items than planned existed for \emph{can}/\emph{cannot}, completive \emph{finish} and completive \emph{done}; per-item provenance, shortages and the scoring script are in the code release (\url{https://anonymous.4open.science/r/cap-accommodation/}). 
\paragraph{LLM judges.} Two open-weight LLMs from different families, \texttt{openai/gpt-oss-20b} and \texttt{Qwen/Qwen3-8B}, labelled the same \res{audit.v4.n_items} items blind, with the panel's instructions and options, a fixed prompt and greedy decoding (prompt and settings in the code release). On the \res{audit.j.n_bc} feature items, gpt-oss-20b gives the consensus label on \res{audit.j.h_gptoss.bc.k} ($\kappa=\res{audit.j.h_gptoss.bc.kappa}$) and Qwen3-8B on \res{audit.j.h_qwen.bc.k} ($\kappa=\res{audit.j.h_qwen.bc.kappa}$). On the \res{audit.j.n_a} register pairs, the judges call \res{audit.j.h_gptoss.a.n_same_y} and \res{audit.j.h_qwen.a.n_same_y} pairs equally formal, against the panel's \res{audit.j.h_gptoss.a.n_same_x}; where the panel picked a reply, gpt-oss-20b picked the same one in \res{audit.j.h_gptoss.a_dir.k} of \res{audit.j.h_gptoss.a_dir.n} pairs and Qwen3-8B in \res{audit.j.h_qwen.a_dir.k} of \res{audit.j.h_qwen.a_dir.n}. The two judges agree with each other on \res{audit.j.gptoss_qwen.bc.k} feature items and \res{audit.j.gptoss_qwen.a.k} register pairs. The judges ran after the human labels were in, and no validation rule or reported effect uses their labels.

\subsection{Calibrated Accommodation Pairs}
\label{app:cap}
This section gives the CAP setup, its pre-declared criteria, pair construction and training details.
\paragraph{Pre-declared criteria and pair audit.} Analysis code and criteria were frozen before any result (D24, in the code release). M1: held-out convergence above the control by at least 0.02 with a 95\% CI above zero, from a bootstrap over training runs and requests. M2: per variety, a reply-to-user density ratio closer to 1 than the one-sided recipe's, fewer replies above twice the user's density, and at most 0.05 inserted features per 100 words for Standard-English users. M3: a positive within-request tracking slope, above the one-sided recipe's. M4: IFEval within two points, on the CI. M5: no more dialect than the control under a request for Standard English. Before training, an audit found no held-out features in the pairs and length or punctuation differences of one to two words in opposite directions across pair types; only \res{dv.pa.within25} of the mirrored replies land within 25\% of the user's density, \res{dv.pa.undershoot} features below it on average, because many replies offer too few sentences to render, and the over-dense replies are also less fluent under the SFT model (\res{dv.pa.fluency_o} nats per token).
\begin{itemize}[leftmargin=*,itemsep=1pt,topsep=2pt]
  \item \textbf{U}, against under-accommodation: a dialect request; chosen, the reply re-rendered at the user's density (the subset of sentences whose achieved density is closest to the user's, at most 1.25 times it); rejected, the standard reply.
  \item \textbf{O}, against over-accommodation: the same request; chosen as in U; rejected, every sentence re-rendered, kept only with at least twice the chosen reply's features.
  \item \textbf{A}, against imposition: the Standard-English twin; chosen, the standard reply; rejected, every sentence re-rendered.
\end{itemize}
The mix is U\,:\,O\,:\,A $=2:1:1$ over AAE and Singlish, at a third of the DPO pairs (\Cref{fig:cap}, \Cref{sec:cap}). Candidates carry only the trained features, so held-out features and varieties test transfer.
\paragraph{OLMo-2-1B.} The released recipe on the 120k subset of its preference mix plus 60k CAP pairs, three training runs, compared with the matched controls and one-sided pairs of \Cref{sec:exp:h5}; an ablation replaces the O pairs by U pairs, and a further baseline fine-tunes on the re-rendered replies before DPO; two runs drop the A pairs, and two runs build the one-sided recipe with CAP's pair builder (pre-declared confound control, filed before its result: the registered one-sided pairs predate a formatting fix).
\paragraph{Qwen3-8B.} As pre-registered (D24, in the code release), we also trained CAP on Qwen3-8B (LoRA-DPO, three runs per arm); it did not pass M1, raising held-out convergence by \res{dv.capres.q3.cap_crheld} (CI [\res{dv.capres.q3.cap_crheld_lo}, \res{dv.capres.q3.cap_crheld_hi}]) against the 0.02 minimum. In a blind check of \res{audit.cap.n_items} of its replies, the expert panel's consensus labels mark \res{audit.cap.cap.n_exaggerated} of \res{audit.cap.cap.n} CAP replies and \res{audit.cap.unmatched.n_exaggerated} of \res{audit.cap.unmatched.n} one-sided replies exaggerated, so the check does not separate them.
\paragraph{Evaluation.} 300 fresh requests from a disjoint pool, rendered as in \Cref{sec:method}: held-out-feature and held-out-variety convergence; the ratio of reply to user density and the share of replies above twice the user's; dialect features in replies to Standard-English users; a density intervention that renders each request at two densities; compliance with explicit requests for Standard English or for a variety; IFEval and GSM8K.

\subsection{Further Method Details}
\label{app:method_more}
Details of the requests, lineages, measures and measurement checks summarized in \Cref{sec:intro,sec:method}.

\paragraph{Rarity in traffic and training data.} After a human check of our recognisers' hits, about \res{audit.auth.wild_aae_corr.est}\% of \res{dv.wild.n} English first turns in WildChat \citep{zhao2024wildchat} carry an audited AAE feature (95\% CI \res{audit.auth.wild_aae_corr.lo}--\res{audit.auth.wild_aae_corr.hi}\%), and the recognisers fire on only \res{dv.pref.aae_morph_7b}\% of the prompts in an open preference mix (\Cref{app:dpo}).

\paragraph{Lineages.} OLMo-2 at four sizes, OLMo-3-7B, OLMo-3.1-32B, T\"ulu-3, T\"ulu-2 at two sizes and Zephyr \citep{olmo2025olmo2,olmo2025olmo3,lambert2024tulu,ivison2023camels,tunstall2023zephyr}, starting from the base model in \res{design.n_open_lineages} of them; each checkpoint answers \res{design.n_seeds} requests, each written in \res{design.n_varieties} varieties and in Standard English.

\paragraph{Released artefacts.} Code and data are at \url{https://anonymous.4open.science/r/cap-accommodation/}, including \res{design.n_seeds} requests rendered in \res{design.n_varieties} varieties, two registers and Standard English with their validity gates and feature detectors checked against a blind audit of \res{audit.v4.n_items} items and a WildChat precision check of \res{audit.auth.n_items} items, both labelled by consensus of a panel of five or more annotators with PhD-level training in related fields; the replies of \res{dv.dep.n} deployed chatbots follow on publication.

\paragraph{Prestige tiers and renderings.} The varieties are ordered by prestige tiers that we pre-registered from attitude studies \citep{coupland2007ideologised,bernaisch2016attitudes,cavallaro2014singapore,purnell1999perceptual,rickford2016language,luhman1990appalachian,obi2014language}. Morphosyntax follows the eWAVE rules of Multi-VALUE \citep{ziems-etal-2023-multi}, applied at the eWAVE attestation rates scaled by 0.8; lexis, particles and spelling follow curated lexicons (\Cref{app:details}).

\paragraph{Lineages and sampling.} We study lineages whose every post-training stage is public, so consecutive checkpoints differ by one training stage: OLMo-2 at 1B, 7B, 13B and 32B \citep{olmo2025olmo2} and OLMo-3-7B \citep{olmo2025olmo3}, each with base, SFT, DPO and RLVR checkpoints, T\"ulu-3-8B's SFT, DPO and RLVR stages \citep{lambert2024tulu} and T\"ulu-2's SFT and DPO stages at 7B and 13B \citep{ivison2023camels}, whose bases are gated, Zephyr-7B-$\beta$'s base, SFT and DPO stages \citep{tunstall2023zephyr}, and OLMo-3.1-32B's base, SFT, DPO and RLVR stages \citep{olmo2025olmo3}. Qwen2.5-72B tests scale, and widely used families and deployed chatbots test breadth, at their released stages (\Cref{app:models}).
Aligned stages use their own chat template; base stages are run as raw continuation, under the chat template, and in a transcript scaffold (\Cref{app:details}).
We sample \res{design.n_samples_per_seed} replies per request for each of \res{design.n_sampling_seeds} sampling seeds, at temperature \res{design.temperature} and top-$p$ \res{design.top_p}.

\paragraph{Uncounted features.} Some salient features fail our false-positive check on standard text and are not counted; the sister example's \emph{keep} (no third-person \emph{-s}) is one, while its \emph{y'all} is counted (\Cref{app:detectors}).

\paragraph{Validity checks.} Two pre-registered validity checks decide how each cue is read: every counted variety feature must fire on at most 2\% of replies to \vty{SAE} requests at every stage (\Cref{tab:detectors}), and at least 60\% of replies must offer three or more opportunities. Spelling and the lexis-carried varieties (\vty{SGE}, \vty{NGP}) fail the second check at some or all stages, so for them we use the \emph{rate-based} ratio, with realizations per 100 words in place of densities; for AAE the two ratios agree.

\paragraph{Reading the ratio.} $\CR{v}=1$ means that the replies differ between the variety and its Standard-English twin by as much as the requests do, $0$ that they do not differ, and negative values a reversed difference; the ratio does not measure absolute matching to the user's density.

\paragraph{Implicit reward.} For the length-normalized DPO recipe of OLMo~2 and T\"ulu~3, $\beta\hat r_m$ is the score inside the training loss, so its differences between replies to the same request are implicit-reward differences; for the other recipes we read it as a descriptive update score.

The matched and standard replies differ only in the rendered variety features, including some that the convergence ratio does not count (\emph{keeps} versus \emph{keep}).

\paragraph{Compute.} At least \res{dv.gpu.total} GPU hours on NVIDIA H800s, \res{dv.gpu.dpo} of them for the DPO re-runs; API queries to deployed chatbots are logged with their cost in the released repository.

\paragraph{Measurement validity.}
An expert panel assigned consensus labels to \res{audit.v4.n_items} blind items (\Cref{tab:audit} in \Cref{app:detectors}). On requested replies, the consensus labels and detectors agree on \res{audit.v4.b.primary.agreement.k} of \res{audit.v4.b.primary.agreement.n} items ($\kappa=\res{audit.v4.b.primary.agreement.kappa}$), and the panel confirmed \res{audit.v4.b.primary.precision.k} of \res{audit.v4.b.primary.precision.n} detector hits. The GYAFC-trained formality ranker that the protocol named, run after labelling, picks the panel's more formal reply in \res{audit.r.trimmed.k} of \res{audit.r.trimmed.n} pairs; the part-of-speech score does not pass (\res{audit.v4.a.trimmed.k} of \res{audit.v4.a.trimmed.n}), and no claim rests on it. On raw user requests the recognisers over-fire: of \res{audit.auth.flagged.n} WildChat requests they flagged, the expert panel confirmed \res{audit.auth.flagged.k}. We therefore apply them only to model replies. Default replies carry few hits, and the panel confirmed \res{audit.v4.c.primary.precision.k} of \res{audit.v4.c.primary.precision.n}; the false ones, mostly multiple negation (confirmed in \res{audit.v4.c.by_feature.negative_concord.precision.k} of \res{audit.v4.c.by_feature.negative_concord.precision.n}), occur about as often in replies to the Standard-English twins (\res{rh.aligned.aae.aae_hits} against \res{rh.aligned.sae.aae_hits} hits over \res{rh.aligned.n_replies} replies each after post-training), so they largely cancel in the ratio, and default convergence remains subject to this detector error (\Cref{app:detectors}). Prior function-word measures barely move between varieties and stages, and instructions, personas and copying do not explain the pattern (\Cref{app:results}).

\subsection{Further CAP Details}
\label{app:cap_more}
Settings, pilot, registered criteria, ablations and sensitivity of \Cref{sec:cap} in full.

\paragraph{What the pairs express.} CAP's pairs express preferences against sampled under- and over-accommodation; we do not claim a reward that peaks at the user's density. The three pair types do different work: the matching and anchoring contrasts carry the gain, the over-accommodation contrast buys restraint, and density matching curbs the Singlish overshoot (\Cref{sec:cap:ablation}).

\paragraph{Settings.} We train OLMo-2-1B, whose stages are all released and whose controls from \Cref{sec:exp:h5} carry over, with a fixed budget in every arm (\Cref{tab:capsetup} in \Cref{app:cap}).
\begin{table}[!htb]
  \centering
  \scriptsize
  \setlength{\tabcolsep}{2pt}
  \begin{tabular}{@{}l p{0.75\columnwidth}@{}}
    \toprule
    \rowcolor{gray!10} & \textbf{OLMo-2-1B} (all stages released) \\
    \midrule
    Start & released SFT checkpoint \\
    Training & released DPO recipe, full fine-tuning \\
    Data per arm & 120k released pairs + 60k arm pairs \\
    Runs (planned) & three per arm; completed: \Cref{tab:cap} \\
    \addlinespace[0.5mm]
    \multicolumn{2}{@{}l}{\textit{Trained comparisons: same prompts, same number of pairs}} \\
    Control & a standard reply over another \\
    One-sided & dialect rendering over the standard reply \\
    Ablations & no O pairs; no A pairs; SFT on mirrored replies; one-sided pairs from CAP's builder; mirror target $1.5\times$ (exploratory) \\
    \multicolumn{2}{@{}l}{\textit{Without training}} \\
    Baselines & released DPO \\
    \addlinespace[0.5mm]
    Evaluation & \multicolumn{1}{p{0.75\columnwidth}@{}}{300 fresh requests in AAE, Singlish, Nigerian Pidgin, Appalachian and Standard English; a density intervention; requests for Standard English or a variety; IFEval; GSM8K} \\
    \bottomrule
  \end{tabular}
  \caption{\textbf{CAP is tested at a fixed budget against matched controls.} Arms share prompts and pair counts.}
  \label{tab:capsetup}
\end{table}

The test set is 300 fresh requests from a pool disjoint from training, validation and the requests of \Cref{sec:replies}.

We compare CAP with a control that prefers one standard reply over another and with the one-sided pairs of \Cref{sec:exp:h5}, at the same budget, and, without training, with the unchanged model (\Cref{tab:capsetup}).

\paragraph{Pilot.} The pre-declared pilot on OLMo-2-1B passed \res{dv.gate.n_pass} of \res{dv.gate.n_checks} checks on validation requests. Its trained-feature convergence exceeds the control by \res{dv.gate.cr_diff} (CI [\res{dv.gate.cr_lo}, \res{dv.gate.cr_hi}]), and it inserts no dialect into replies to Standard-English users (\res{dv.gate.ins_diff} features per 100 words against the control). Its calibration falls below the pre-declared band: CAP under-accommodates, reproducing \res{dv.gate.ratio_aae} of the user's density for AAE and \res{dv.gate.ratio_sge} for Singlish, as the audit of its pairs anticipated (\Cref{app:cap}). We continue as pre-declared, without tuning on the pilot.

\paragraph{Registered criteria M2, M3 and M5.} Its AAE replies stay as far below the user's density as the one-sided pairs' (\res{dv.capres.b1.cap_ratio_aae} against \res{dv.capres.b1.unm_ratio_aae} of it), and its Singlish replies no longer overshoot (\res{dv.capres.b1.cap_ratio_sge} against \res{dv.capres.b1.unm_ratio_sge}); the Singlish difference is not resolved, because the one-sided runs vary widely, so M2 fails as registered. The one-sided pairs track the user's Singlish density more steeply, by overshooting it (slope \res{dv.capres.b1.unm_track} against \res{dv.capres.b1.cap_track}; M3 fails), and neither recipe makes the 1B model follow a request for Standard English (\res{dv.capres.b1.cap_ostd} and \res{dv.capres.b1.unm_ostd} more features per 100 words than the control; M5 fails).

\paragraph{Ablation variants.} Each ablation removes one component of CAP on OLMo-2-1B at the same budget: the O pairs (their slots filled with U pairs), the A pairs, the preference contrast itself (SFT on the mirrored replies, then DPO on control pairs), and density matching (the one-sided pairs of \Cref{sec:exp:h5}, and, as a pre-declared control for their older pair builder, the same recipe built with CAP's builder); an exploratory run raises the mirror target to 1.5 times the user's density.

\paragraph{Ablation results.} \Cref{tab:capabl} isolates the O pairs. Without them, held-out convergence rises to \res{dv.capabl.noO_crheld} over the control (\res{dv.capabl.noO_n} runs), but the model imposes dialect: replies to Standard-English users gain \res{dv.capabl.noO_ins} features per 100 words, and replies to a request for Standard English keep \res{dv.capabl.noO_ostd}. The O pairs thus trade convergence for restraint, the behaviour CAP is built to keep; the registered one-sided pairs gain less (\res{dv.capabl.unm_crheld}) and insert more (\res{dv.capabl.unm_ins}). Imitation is not enough: fine-tuning on the same mirrored replies before DPO leaves held-out convergence unchanged (\res{dv.capabl.msft_crheld}) and costs \res{dv.capabl.msft_ifeval} IFEval points, so the gain comes from the preference contrast. Within CAP's mix, the A pairs matter as much: without them, held-out convergence barely moves (\res{dv.capabl.noA_crheld}, \res{dv.capabl.noA_n} runs). The mirror target sets how far CAP accommodates: raising it to 1.5 times the user's density, in an exploratory run, multiplies held-out convergence by \res{dv.capabl.x15_fold} (\res{dv.capabl.x15_crheld}) without insertion (\res{dv.capabl.x15_ins}). Density matching is what curbs the overshoot, although AAE replies remain well below the user's density. Built with CAP's builder, the one-sided pairs raise held-out convergence by \res{dv.capabl.mx_crheld} (CI [\res{dv.capabl.mx_crheld_lo}, \res{dv.capabl.mx_crheld_hi}], \res{dv.capabl.mx_n} runs), a gain similar to CAP's, but by overshooting: their Singlish replies reach \res{dv.capabl.mx_sge_ratio} times the user's density, \res{dv.capabl.mx_sge_tail}\% of them above twice it (CAP: \res{dv.capabl.cap_sge_ratio}, \res{dv.capabl.cap_sge_tail}\%), replies to Standard-English users gain \res{dv.capabl.mx_ins} features per 100 words, and replies to a request for Standard English keep \res{dv.capabl.mx_ostd}.

\paragraph{Sensitivity of the CAP gain.} CAP's held-out gain on OLMo-2-1B does not depend on how it is estimated or where it is tested. A bootstrap over requests alone narrows its interval to [\res{dv.capsens.req_lo}, \res{dv.capsens.req_hi}] (registered, over runs and requests: [\res{dv.capres.b1.cap_crheld_lo}, \res{dv.capres.b1.cap_crheld_hi}]); dropping the 2\% false-positive exclusion leaves it at \res{dv.capsens.noexcl}; and on the \res{design.n_seeds} requests of \Cref{sec:replies}, used before the method existed, it is \res{dv.capsens.v1} (CI [\res{dv.capsens.v1_lo}, \res{dv.capsens.v1_hi}]).

\section{Additional Results}
Results that support the main text, the preference-data audit with the DPO re-runs, and an extended discussion.

\subsection{Additional Results}
\label{app:results}
This section reports results that support the main text: competence, sensitivity, every stage of every model, the update per variety, authentic text and model size.
\Cref{tab:mechanism} gives the matching bonus of every update and variety with its 95\% CI.
\begin{table}[!htb]
  \centering
  \small
  \setlength{\tabcolsep}{4pt}
  \begin{tabular}{l cc cc}
    \toprule
    & \multicolumn{2}{c}{\textbf{AAE}} & \multicolumn{2}{c}{\textbf{Singlish}} \\
    \cmidrule(lr){2-3} \cmidrule(lr){4-5}
    \textbf{Model, after DPO} & request & default & request & default \\
    \midrule
    OLMo-2-7B & \cellcolor{cCmpA!26}\res{dv.comp_corr.l2_aae.comp} & \res{comp.l2.dpo.aae.default_ratio} & \res{dv.comp_corr.l2_sge.comp} & \res{comp.l2.dpo.sge.default_ratio} \\
    T\"ulu-3-8B & \cellcolor{cCmpA!27}\res{dv.comp_corr.l4_aae.comp} & \res{comp.l4.dpo.aae.default_ratio} & \res{dv.comp_corr.l4_sge.comp} & \res{comp.l4.dpo.sge.default_ratio} \\
    OLMo-3-7B & \cellcolor{cCmpA!21}\res{dv.comp_corr.l3_aae.comp} & \res{comp.l3.dpo.aae.default_ratio} & \res{dv.comp_corr.l3_sge.comp} & \res{comp.l3.dpo.sge.default_ratio} \\
    \bottomrule
  \end{tabular}
  \caption{\textbf{Asked to, all three families produce AAE features, two at more than half their authentic rate; by default, almost none appear.} Feature competence after DPO; request rates are scaled by the audited detector precision.}
  \label{tab:comp}
\end{table}

\begin{table*}[t]
  \centering
  \tabstyle
  \begin{tabular}{l ccc cc}
    \toprule
    & \multicolumn{3}{c}{\textbf{Prestige and register cues}} & \multicolumn{2}{c}{\textbf{Stigmatized varieties}} \\
    \cmidrule(lr){2-4} \cmidrule(lr){5-6}
    \textbf{Update} & \vty{INF} & \vty{BRE} & \vty{AUS} & \vty{AAE} & \vty{SGE} \\
    \midrule
    \multicolumn{6}{@{}l}{\textit{DPO vs.\ SFT}} \\
    OLMo-2-1B & \makecell{\res{e4.l1.dpo_vs_sft.inf.mb}\\[-1pt]{\scriptsize\rci{e4.l1.dpo_vs_sft.inf.mb}}} & \makecell{\res{e4.l1.dpo_vs_sft.bre.mb}\\[-1pt]{\scriptsize\rci{e4.l1.dpo_vs_sft.bre.mb}}} & \makecell{\res{e4.l1.dpo_vs_sft.aus.mb}\\[-1pt]{\scriptsize\rci{e4.l1.dpo_vs_sft.aus.mb}}} & \makecell{\res{e4.l1.dpo_vs_sft.aae.mb}\\[-1pt]{\scriptsize\rci{e4.l1.dpo_vs_sft.aae.mb}}} & \makecell{\res{e4.l1.dpo_vs_sft.sge.mb}\\[-1pt]{\scriptsize\rci{e4.l1.dpo_vs_sft.sge.mb}}} \\
    OLMo-2-7B & \makecell{\res{e4.l2.dpo_vs_sft.inf.mb}\\[-1pt]{\scriptsize\rci{e4.l2.dpo_vs_sft.inf.mb}}} & \makecell{\res{e4.l2.dpo_vs_sft.bre.mb}\\[-1pt]{\scriptsize\rci{e4.l2.dpo_vs_sft.bre.mb}}} & \makecell{\res{e4.l2.dpo_vs_sft.aus.mb}\\[-1pt]{\scriptsize\rci{e4.l2.dpo_vs_sft.aus.mb}}} & \makecell{\res{e4.l2.dpo_vs_sft.aae.mb}\\[-1pt]{\scriptsize\rci{e4.l2.dpo_vs_sft.aae.mb}}} & \makecell{\res{e4.l2.dpo_vs_sft.sge.mb}\\[-1pt]{\scriptsize\rci{e4.l2.dpo_vs_sft.sge.mb}}} \\
    OLMo-2-13B & \makecell{\res{e4.olmo2_13b.dpo_vs_sft.inf.mb}\\[-1pt]{\scriptsize\rci{e4.olmo2_13b.dpo_vs_sft.inf.mb}}} & \makecell{\res{e4.olmo2_13b.dpo_vs_sft.bre.mb}\\[-1pt]{\scriptsize\rci{e4.olmo2_13b.dpo_vs_sft.bre.mb}}} & \makecell{\res{e4.olmo2_13b.dpo_vs_sft.aus.mb}\\[-1pt]{\scriptsize\rci{e4.olmo2_13b.dpo_vs_sft.aus.mb}}} & \makecell{\res{e4.olmo2_13b.dpo_vs_sft.aae.mb}\\[-1pt]{\scriptsize\rci{e4.olmo2_13b.dpo_vs_sft.aae.mb}}} & \makecell{\res{e4.olmo2_13b.dpo_vs_sft.sge.mb}\\[-1pt]{\scriptsize\rci{e4.olmo2_13b.dpo_vs_sft.sge.mb}}} \\
    OLMo-3-7B & \makecell{\res{e4.l3.dpo_vs_sft.inf.mb}\\[-1pt]{\scriptsize\rci{e4.l3.dpo_vs_sft.inf.mb}}} & \makecell{\res{e4.l3.dpo_vs_sft.bre.mb}\\[-1pt]{\scriptsize\rci{e4.l3.dpo_vs_sft.bre.mb}}} & \makecell{\res{e4.l3.dpo_vs_sft.aus.mb}\\[-1pt]{\scriptsize\rci{e4.l3.dpo_vs_sft.aus.mb}}} & \makecell{\res{e4.l3.dpo_vs_sft.aae.mb}\\[-1pt]{\scriptsize\rci{e4.l3.dpo_vs_sft.aae.mb}}} & \makecell{\res{e4.l3.dpo_vs_sft.sge.mb}\\[-1pt]{\scriptsize\rci{e4.l3.dpo_vs_sft.sge.mb}}} \\
    OLMo-2-32B & \makecell{\res{e4.olmo2_32b.dpo_vs_sft.inf.mb}\\[-1pt]{\scriptsize\rci{e4.olmo2_32b.dpo_vs_sft.inf.mb}}} & \makecell{\res{e4.olmo2_32b.dpo_vs_sft.bre.mb}\\[-1pt]{\scriptsize\rci{e4.olmo2_32b.dpo_vs_sft.bre.mb}}} & \makecell{\res{e4.olmo2_32b.dpo_vs_sft.aus.mb}\\[-1pt]{\scriptsize\rci{e4.olmo2_32b.dpo_vs_sft.aus.mb}}} & \cellcolor{cCmpA!27}\makecell{\res{e4.olmo2_32b.dpo_vs_sft.aae.mb}\\[-1pt]{\scriptsize\rci{e4.olmo2_32b.dpo_vs_sft.aae.mb}}} & \makecell{\res{e4.olmo2_32b.dpo_vs_sft.sge.mb}\\[-1pt]{\scriptsize\rci{e4.olmo2_32b.dpo_vs_sft.sge.mb}}} \\
    T\"ulu-3-8B & \makecell{\res{e4.l4.dpo_vs_sft.inf.mb}\\[-1pt]{\scriptsize\rci{e4.l4.dpo_vs_sft.inf.mb}}} & \makecell{\res{e4.l4.dpo_vs_sft.bre.mb}\\[-1pt]{\scriptsize\rci{e4.l4.dpo_vs_sft.bre.mb}}} & \makecell{\res{e4.l4.dpo_vs_sft.aus.mb}\\[-1pt]{\scriptsize\rci{e4.l4.dpo_vs_sft.aus.mb}}} & \makecell{\res{e4.l4.dpo_vs_sft.aae.mb}\\[-1pt]{\scriptsize\rci{e4.l4.dpo_vs_sft.aae.mb}}} & \makecell{\res{e4.l4.dpo_vs_sft.sge.mb}\\[-1pt]{\scriptsize\rci{e4.l4.dpo_vs_sft.sge.mb}}} \\
    \multicolumn{6}{@{}l}{\textit{SFT vs.\ base}} \\
    OLMo-2-1B & \makecell{\res{e4.l1.sft_vs_base.inf.mb}\\[-1pt]{\scriptsize\rci{e4.l1.sft_vs_base.inf.mb}}} & \makecell{\res{e4.l1.sft_vs_base.bre.mb}\\[-1pt]{\scriptsize\rci{e4.l1.sft_vs_base.bre.mb}}} & \makecell{\res{e4.l1.sft_vs_base.aus.mb}\\[-1pt]{\scriptsize\rci{e4.l1.sft_vs_base.aus.mb}}} & \makecell{\res{e4.l1.sft_vs_base.aae.mb}\\[-1pt]{\scriptsize\rci{e4.l1.sft_vs_base.aae.mb}}} & \makecell{\res{e4.l1.sft_vs_base.sge.mb}\\[-1pt]{\scriptsize\rci{e4.l1.sft_vs_base.sge.mb}}} \\
    OLMo-2-7B & \makecell{\res{e4.l2.sft_vs_base.inf.mb}\\[-1pt]{\scriptsize\rci{e4.l2.sft_vs_base.inf.mb}}} & \makecell{\res{e4.l2.sft_vs_base.bre.mb}\\[-1pt]{\scriptsize\rci{e4.l2.sft_vs_base.bre.mb}}} & \makecell{\res{e4.l2.sft_vs_base.aus.mb}\\[-1pt]{\scriptsize\rci{e4.l2.sft_vs_base.aus.mb}}} & \makecell{\res{e4.l2.sft_vs_base.aae.mb}\\[-1pt]{\scriptsize\rci{e4.l2.sft_vs_base.aae.mb}}} & \makecell{\res{e4.l2.sft_vs_base.sge.mb}\\[-1pt]{\scriptsize\rci{e4.l2.sft_vs_base.sge.mb}}} \\
    \multicolumn{6}{@{}l}{\textit{RLVR (Instruct) vs.\ DPO}} \\
    OLMo-2-7B & \makecell{\res{e4.l2.instruct_vs_dpo.inf.mb}\\[-1pt]{\scriptsize\rci{e4.l2.instruct_vs_dpo.inf.mb}}} & \makecell{\res{e4.l2.instruct_vs_dpo.bre.mb}\\[-1pt]{\scriptsize\rci{e4.l2.instruct_vs_dpo.bre.mb}}} & \makecell{\res{e4.l2.instruct_vs_dpo.aus.mb}\\[-1pt]{\scriptsize\rci{e4.l2.instruct_vs_dpo.aus.mb}}} & \makecell{\res{e4.l2.instruct_vs_dpo.aae.mb}\\[-1pt]{\scriptsize\rci{e4.l2.instruct_vs_dpo.aae.mb}}} & \makecell{\res{e4.l2.instruct_vs_dpo.sge.mb}\\[-1pt]{\scriptsize\rci{e4.l2.instruct_vs_dpo.sge.mb}}} \\
    \multicolumn{6}{@{}l}{\textit{Instruct vs.\ base}} \\
    Mistral-7B-v0.3 & \rpm{e4b.mistral_7b.instruct_vs_base.inf.mb} & \rpm{e4b.mistral_7b.instruct_vs_base.bre.mb} & \rpm{e4b.mistral_7b.instruct_vs_base.aus.mb} & \rpm{e4b.mistral_7b.instruct_vs_base.aae.mb} & \rpm{e4b.mistral_7b.instruct_vs_base.sge.mb} \\
    Qwen3-8B & \rpm{e4b.qwen3_8b.instruct_vs_base.inf.mb} & \rpm{e4b.qwen3_8b.instruct_vs_base.bre.mb} & \rpm{e4b.qwen3_8b.instruct_vs_base.aus.mb} & \cellcolor{cCmpA!21}\rpm{e4b.qwen3_8b.instruct_vs_base.aae.mb} & \rpm{e4b.qwen3_8b.instruct_vs_base.sge.mb} \\
    \bottomrule
  \end{tabular}
  \caption{\textbf{The DPO update disfavours matching AAE on \res{dv.n_mb_aae_neg} of \res{dv.n_lineages} lineages; on OLMo-2-32B it favours matching every variety.} Matching bonus per update and variety (nats per token), with 95\% bootstrap CIs over requests.}
  \label{tab:mechanism}
\end{table*}

\begin{figure*}[t]
  \centering
  \includegraphics[width=0.8\textwidth]{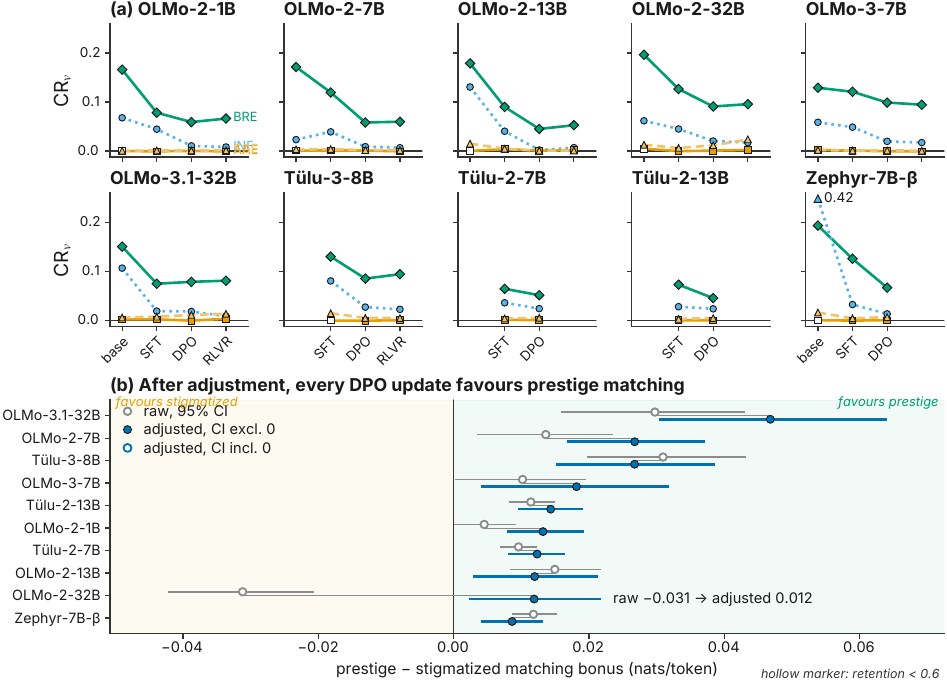}
  \caption{\textbf{Replies lose convergence through post-training, while the update still selects.} (a) Convergence ratio per cue and stage, by lineage. (b) Prestige-minus-stigmatized contrast of each DPO update, raw and adjusted for baseline likelihood and length, with 95\% CIs.}
  \label{fig:heatmap_full}
\end{figure*}

\begin{figure*}[t]
  \centering
  \includegraphics[width=\textwidth]{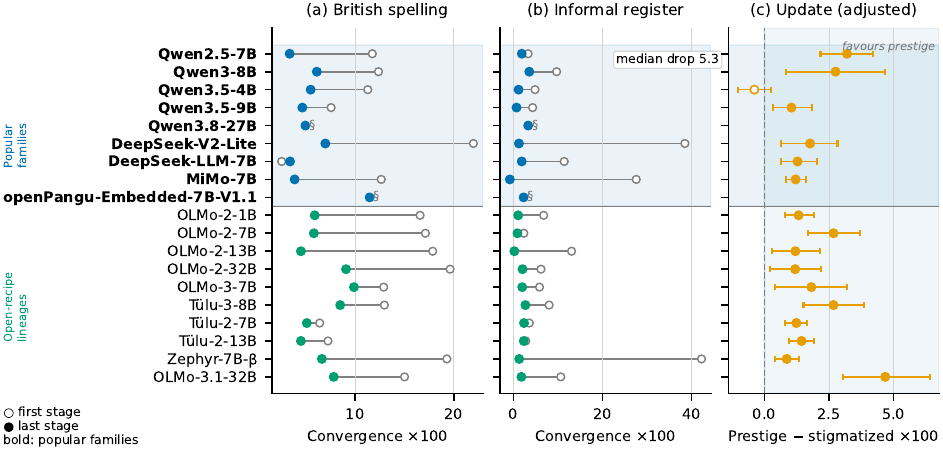}
  \caption{\textbf{Post-training lowers register convergence in popular and open-recipe models alike; the adjusted update favours prestige matching in all but one.} First ($\circ$) and last ($\bullet$) stage per model; \S: chat-only release; (c): adjusted update with 95\% CI.}
  \label{fig:models_full}
\end{figure*}

\textbf{Prior accommodation measures miss the variety.} Function-word measures, namely language style matching \citep{niederhoffer2002linguistic,ireland2010language} and the convergence measure of \citet{blevins-etal-2026-language}, change by less than .04 between varieties and stages on our replies (\Cref{tab:measures}). They capture style, not the features that distinguish varieties, so they cannot show the standardization of \Cref{sec:replies}.
\begin{table*}[t]
  \centering
  \tabstyle
  \setlength{\tabcolsep}{2.4pt}
  \begin{tabular}{@{}l cc cc cc@{}}
    \toprule
    & \multicolumn{2}{c}{\vty{AAE}} & \multicolumn{2}{c}{\vty{INF}} & \multicolumn{2}{c}{\vty{SGE}} \\
    \cmidrule(lr){2-3} \cmidrule(lr){4-5} \cmidrule(lr){6-7}
    \textbf{Measure} & base & DPO & base & DPO & base & DPO \\
    \midrule
    LSM, novel spans & \resT{e3.l2.base.aae.lsm_novel.delta}\res{e3.l2.base.aae.lsm_novel.mark} & \resT{e3.l2.dpo.aae.lsm_novel.delta}\res{e3.l2.dpo.aae.lsm_novel.mark} & \resT{e3.l2.base.inf.lsm_novel.delta}\res{e3.l2.base.inf.lsm_novel.mark} & \resT{e3.l2.dpo.inf.lsm_novel.delta}\res{e3.l2.dpo.inf.lsm_novel.mark} & \resT{e3.l2.base.sge.lsm_novel.delta}\res{e3.l2.base.sge.lsm_novel.mark} & \resT{e3.l2.dpo.sge.lsm_novel.delta}\res{e3.l2.dpo.sge.lsm_novel.mark} \\
    Convergence \citep{blevins-etal-2026-language} & \resT{e3.l2.base.aae.blevins_conv_novel.delta}\res{e3.l2.base.aae.blevins_conv_novel.mark} & \resT{e3.l2.dpo.aae.blevins_conv_novel.delta}\res{e3.l2.dpo.aae.blevins_conv_novel.mark} & \resT{e3.l2.base.inf.blevins_conv_novel.delta}\res{e3.l2.base.inf.blevins_conv_novel.mark} & \resT{e3.l2.dpo.inf.blevins_conv_novel.delta}\res{e3.l2.dpo.inf.blevins_conv_novel.mark} & \resT{e3.l2.base.sge.blevins_conv_novel.delta}\res{e3.l2.base.sge.blevins_conv_novel.mark} & \resT{e3.l2.dpo.sge.blevins_conv_novel.delta}\res{e3.l2.dpo.sge.blevins_conv_novel.mark} \\
    Copy rate & \resT{e3.l2.base.aae.copy_rate.delta}\res{e3.l2.base.aae.copy_rate.mark} & \cellcolor{cCmpA!2}\resT{e3.l2.dpo.aae.copy_rate.delta}\res{e3.l2.dpo.aae.copy_rate.mark} & \resT{e3.l2.base.inf.copy_rate.delta}\res{e3.l2.base.inf.copy_rate.mark} & \resT{e3.l2.dpo.inf.copy_rate.delta}\res{e3.l2.dpo.inf.copy_rate.mark} & \resT{e3.l2.base.sge.copy_rate.delta}\res{e3.l2.base.sge.copy_rate.mark} & \resT{e3.l2.dpo.sge.copy_rate.delta}\res{e3.l2.dpo.sge.copy_rate.mark} \\
    \midrule
    \textbf{CR (ours)} & \resT{e1.l2.base.aae.cr} & \cellcolor{cCmpA!27}\resT{e1.l2.dpo.aae.cr} & \resT{e1.l2.base.inf.cr} & \resT{e1.l2.dpo.inf.cr} & \resT{e1.l2.base.sge.cr_rate} & \resT{e1.l2.dpo.sge.cr_rate} \\
    \bottomrule
  \end{tabular}
  \caption{\textbf{Function-word style measures barely move between varieties and stages.} Difference to the Standard-English twin on novel spans, OLMo-2-7B; $^{*}$: 95\% CI excludes 0.}
  \label{tab:measures}
\end{table*}

\textbf{Instructions, personas and copying do not explain the pattern.} An instruction to ``match my style'', which names no variety, and a community persona raise convergence to stigmatized varieties only slightly, the novel-span filter changes little because copy rates are low, and base-model convergence depends on the framing (\Cref{tab:controls}).
\begin{table}[!htb]
  \centering
  \tabstyle
  \setlength{\tabcolsep}{2.6pt}
  \begin{tabular}{l cccc}
    \toprule
    & \textbf{Prestige} & \multicolumn{3}{c}{\textbf{Stigmatized varieties}} \\
    \cmidrule(lr){3-5}
    \textbf{Condition} & \vty{BRE} & \vty{AAE} & \vty{SGE} \\
    \midrule
    \multicolumn{4}{@{}l}{\textit{After DPO}} \\
    Default & \resT{e1.l2.dpo.bre.cr_rate} & \cellcolor{cCmpA!3}\resT{e1.l2.dpo.aae.cr} & \resT{e1.l2.dpo.sge.cr_rate} \\
    ``Match my style'' & \resT{e2.l2.dpo.style.bre.cr_rate} & \cellcolor{cCmpA!5}\resT{e2.l2.dpo.style.aae.cr} & \resT{e2.l2.dpo.style.sge.cr_rate} \\
    Community persona & \resT{e2.l2.dpo.persona.bre.cr_rate} & \cellcolor{cCmpA!11}\resT{e2.l2.dpo.persona.aae.cr} & \resT{e2.l2.dpo.persona.sge.cr_rate} \\
    No novel-span filter & \resT{e1.l2.dpo.bre.cr_rate_tot} & \cellcolor{cCmpA!3}\resT{e1.l2.dpo.aae.cr_tot} & \resT{e1.l2.dpo.sge.cr_rate_tot} \\
    \multicolumn{4}{@{}l}{\textit{Base model, by framing}} \\
    Raw continuation & \resT{e1.l2.base_raw.bre.cr_rate} & \cellcolor{cCmpA!27}\resT{e1.l2.base_raw.aae.cr} & \resT{e1.l2.base_raw.sge.cr_rate} \\
    Chat template & \resT{e1.l2.base_f1.bre.cr_rate} & \cellcolor{cCmpA!21}\resT{e1.l2.base_f1.aae.cr} & \resT{e1.l2.base_f1.sge.cr_rate} \\
    Transcript scaffold & \resT{e1.l2.base.bre.cr_rate} & \cellcolor{cCmpA!3}\resT{e1.l2.base.aae.cr} & \resT{e1.l2.base.sge.cr_rate} \\
    \bottomrule
  \end{tabular}
  \caption{\textbf{Instructions and personas raise convergence to stigmatized varieties only slightly.} Convergence ratio of OLMo-2-7B under each condition and base framing; informal register and Nigerian Pidgin: \Cref{tab:mainfull}.}
  \label{tab:controls}
\end{table}

\paragraph{Authentic twins.}\label{app:authentic}
We also ran \res{dv.auth.n_lin_all} lineages (OLMo-2 at 1B, 7B and 13B, OLMo-3-7B and T\"ulu-3-8B) on real dialect text paired with Standard-English versions: \res{dv.auth.n_auth_gro} AAE tweets with human Standard-English translations \citep{groenwold-etal-2020-investigating}, each with at least one retained AAE feature, and \res{dv.auth.n_auth_endive_aae} AAE and \res{dv.auth.n_auth_endive_sge} Singlish items of EnDive \citep{gupta-etal-2025-endive}, translated by an LLM from Standard-English benchmark prompts. Harness, detectors and estimators are those of the main runs. A pair is excluded if either reply refuses (a phrase rule on the first 300 characters); for OLMo-2-7B after DPO this removes \res{dv.auth.ref_pair_dpo}\% of the tweet pairs, and the replies to the Standard-English tweets refuse more often (\res{dv.auth.ref_sae_dpo}\%) than those to the AAE tweets (\res{dv.auth.ref_dia_dpo}\%). After post-training, convergence to the human-written AAE is near zero, at most \res{dv.auth.post_max} at any post-trained stage (\Cref{tab:authentic}), as for constructed twins. Before it, convergence depends on the framing, again as for constructed twins: it is near zero under the transcript scaffold and clearly positive under raw continuation and the chat template. Our pre-declared test of standardization on human-written AAE therefore meets its first condition (near-zero convergence after DPO) on all \res{dv.auth.n_lin_all} lineages and its second (below the base model under every framing) only on OLMo-2-13B, so both hold on \res{dv.auth.k_both} of the \res{dv.auth.n_lin} lineages with a base model; T\"ulu-3-8B's base is gated. We use no recogniser-selected WildChat requests as dialect input (\Cref{app:dpo}).
\begin{table*}[t]
  \centering
  \footnotesize
  \setlength{\tabcolsep}{6pt}
  \begin{tabular}{l ccc}
    \toprule
    \textbf{Stage} & \textbf{Human-written AAE} & \textbf{EnDive AAE} & \textbf{EnDive Singlish} \\
    \midrule
    \multicolumn{4}{@{}l}{\textit{OLMo-2-1B}} \\
    base, transcript & \cellcolor{cCmpA!4}\textcolor{gray}{0.2 {\scriptsize[0, 0.4]}} & \textcolor{gray}{\ensuremath{-}0.4 {\scriptsize[\ensuremath{-}1.1, 0.1]}} & 2.1 {\scriptsize[1.0, 3.5]} \\
    base, raw continuation & \cellcolor{cCmpA!16}3.6 {\scriptsize[2.0, 5.6]} & 4.0 {\scriptsize[2.5, 5.6]} & 8.2 {\scriptsize[6.3, 10.6]} \\
    base, chat template & \cellcolor{cCmpA!22}8.0 {\scriptsize[4.7, 11.7]} & 10.0 {\scriptsize[6.7, 13.4]} & 12.9 {\scriptsize[10.4, 15.7]} \\
    SFT & \cellcolor{cCmpA!5}\textcolor{gray}{0.4 {\scriptsize[\ensuremath{-}0.2, 1.6]}} & \textcolor{gray}{0.2 {\scriptsize[\ensuremath{-}0.1, 0.6]}} & 2.9 {\scriptsize[1.8, 4.2]} \\
    DPO & \cellcolor{cCmpA!7}0.7 {\scriptsize[0.2, 1.4]} & \textcolor{gray}{0.3 {\scriptsize[\ensuremath{-}<0.1, 0.6]}} & 1.8 {\scriptsize[1.1, 2.5]} \\
    RLVR & \cellcolor{cCmpA!9}1.1 {\scriptsize[0.4, 2.0]} & 0.4 {\scriptsize[0.1, 0.7]} & 2.4 {\scriptsize[1.6, 3.3]} \\
    \multicolumn{4}{@{}l}{\textit{OLMo-2-7B}} \\
    base, transcript & \cellcolor{cCmpA!4}\textcolor{gray}{0.3 {\scriptsize[0, 0.7]}} & \textcolor{gray}{0.5 {\scriptsize[\ensuremath{-}0.4, 1.8]}} & 4.1 {\scriptsize[1.7, 8.3]} \\
    base, raw continuation & \cellcolor{cCmpA!22}8.0 {\scriptsize[1.0, 17.2]} & 5.4 {\scriptsize[3.6, 7.4]} & 7.7 {\scriptsize[6.4, 9.1]} \\
    base, chat template & \cellcolor{cCmpA!27}15.0 {\scriptsize[10.1, 20.6]} & 6.6 {\scriptsize[4.5, 9.0]} & 10.7 {\scriptsize[9.0, 12.6]} \\
    SFT & \cellcolor{cCmpA!7}0.8 {\scriptsize[<0.1, 2.1]} & \textcolor{gray}{\ensuremath{-}0.1 {\scriptsize[\ensuremath{-}0.6, 0.4]}} & 1.3 {\scriptsize[0.7, 2.2]} \\
    DPO & \cellcolor{cCmpA!8}0.9 {\scriptsize[0.2, 1.9]} & 0.2 {\scriptsize[<0.1, 0.5]} & 0.8 {\scriptsize[0.4, 1.2]} \\
    RLVR & \cellcolor{cCmpA!6}0.6 {\scriptsize[0.1, 1.2]} & \textcolor{gray}{0.2 {\scriptsize[\ensuremath{-}<0.1, 0.6]}} & 0.8 {\scriptsize[0.5, 1.3]} \\
    \multicolumn{4}{@{}l}{\textit{OLMo-2-13B}} \\
    base, transcript & \cellcolor{cCmpA!9}\textcolor{gray}{1.2 {\scriptsize[0, 3.1]}} & 1.8 {\scriptsize[0.7, 3.3]} & 2.0 {\scriptsize[1.2, 2.9]} \\
    base, raw continuation & \cellcolor{cCmpA!13}2.3 {\scriptsize[0.6, 4.3]} & 6.0 {\scriptsize[4.4, 7.7]} & 6.3 {\scriptsize[5.2, 7.5]} \\
    base, chat template & \cellcolor{cCmpA!20}6.2 {\scriptsize[4.4, 8.2]} & 4.6 {\scriptsize[3.3, 6.0]} & 9.0 {\scriptsize[7.6, 10.6]} \\
    SFT & \cellcolor{cCmpA!7}\textcolor{gray}{0.8 {\scriptsize[0, 2.0]}} & \textcolor{gray}{0.5 {\scriptsize[\ensuremath{-}<0.1, 1.1]}} & 1.1 {\scriptsize[0.5, 1.9]} \\
    DPO & \cellcolor{cCmpA!8}0.9 {\scriptsize[0.3, 1.5]} & 0.4 {\scriptsize[0.1, 0.7]} & 1.2 {\scriptsize[0.4, 2.5]} \\
    RLVR & \cellcolor{cCmpA!7}0.8 {\scriptsize[0.1, 1.6]} & \textcolor{gray}{0.4 {\scriptsize[\ensuremath{-}<0.1, 0.9]}} & 0.7 {\scriptsize[0.3, 1.1]} \\
    \multicolumn{4}{@{}l}{\textit{OLMo-3-7B}} \\
    base, transcript & \cellcolor{cCmpA!3}\textcolor{gray}{\ensuremath{-}<0.1 {\scriptsize[\ensuremath{-}<0.1, 0]}} & \textcolor{gray}{0.2 {\scriptsize[\ensuremath{-}<0.1, 0.5]}} & 1.3 {\scriptsize[0.7, 2.0]} \\
    base, raw continuation & \cellcolor{cCmpA!17}4.4 {\scriptsize[1.9, 7.3]} & 1.9 {\scriptsize[1.1, 2.7]} & 3.3 {\scriptsize[2.4, 4.3]} \\
    base, chat template & \cellcolor{cCmpA!10}1.3 {\scriptsize[0.5, 2.1]} & 0.8 {\scriptsize[0.4, 1.3]} & 1.0 {\scriptsize[0.6, 1.5]} \\
    SFT & \cellcolor{cCmpA!9}1.2 {\scriptsize[0.5, 1.9]} & \textcolor{gray}{0.1 {\scriptsize[\ensuremath{-}0.1, 0.3]}} & 1.3 {\scriptsize[0.6, 2.0]} \\
    DPO & \cellcolor{cCmpA!11}1.7 {\scriptsize[1.0, 2.6]} & 0.3 {\scriptsize[0.1, 0.6]} & 1.1 {\scriptsize[0.6, 1.7]} \\
    RLVR & \cellcolor{cCmpA!7}0.7 {\scriptsize[0.3, 1.2]} & \textcolor{gray}{0.1 {\scriptsize[\ensuremath{-}<0.1, 0.3]}} & 1.0 {\scriptsize[0.5, 1.7]} \\
    \multicolumn{4}{@{}l}{\textit{T\"ulu-3-8B$^{\dagger}$}} \\
    SFT & \cellcolor{cCmpA!8}0.9 {\scriptsize[<0.1, 2.5]} & \textcolor{gray}{0.7 {\scriptsize[\ensuremath{-}0.2, 1.7]}} & 6.0 {\scriptsize[4.4, 8.0]} \\
    DPO & \cellcolor{cCmpA!10}1.3 {\scriptsize[0.3, 2.8]} & \textcolor{gray}{0.2 {\scriptsize[\ensuremath{-}0.1, 0.4]}} & 1.1 {\scriptsize[0.3, 2.4]} \\
    \bottomrule
  \end{tabular}
  \caption{\textbf{After post-training, replies barely converge to human-written AAE either.} Convergence ratio $\times$100 on AAE tweets with human Standard-English translations and on LLM-translated twins (EnDive), with 95\% CIs; grey: CI includes 0; pairs with a refused reply excluded.}
  \label{tab:authentic}
\end{table*}

\newpage
\subsubsection{Case Study}
\label{app:case}
The running example of \Cref{sec:intro} (request \texttt{v1s002}), answered by OLMo-2-1B after SFT and after DPO (sampling seed 0, first sample; replies trimmed to 70 words, markdown symbols removed). Both stages answer the AAE request and its Standard-English twin in Standard English; neither reproduces \emph{keep}, \emph{been} or \emph{y'all}.

\begin{promptbox}{Request (AAE)}
I am not sure if my new job can afford the theater tickets I usually buy for my sister, who keep dragging me to the movies. My boss says the budget been canceled this month. Do y'all have any tips on how to handle this situation?
\end{promptbox}

\begin{promptbox}{Request (Standard English)}
I am not sure if my new job can afford the theater tickets I usually buy for my sister, who keeps dragging me to the movies. My boss says the budget has been canceled this month. Do you have any tips on how to handle this situation?
\end{promptbox}

\begin{promptbox}{OLMo-2-1B SFT, reply to the AAE request}
I'm sorry to hear about your financial situation, but I'm here to help! Here are some tips for handling the situation: 1. Talk to your sister about your budget and the financial situation. Explain that you can't afford the theater tickets and ask if there's a way you can still go together. She may be able to let you pay for a ticket, or you could both save up for \ldots
\end{promptbox}
\begin{promptbox}{OLMo-2-1B SFT, reply to the Standard-English request}
It seems like you are in a difficult situation. Here are a few tips that might help you handle this: 1. Communicate with your boss: The first step in any difficult situation is to talk to your boss. Explain to them why you need to buy the theater tickets and what they can do to accommodate you. If the budget has been canceled, you may be able to negotiate a \ldots
\end{promptbox}
\begin{promptbox}{OLMo-2-1B DPO, reply to the AAE request}
Handling a situation where you're concerned about your budget but need to continue supporting your sister's movie-going habits presents a delicate balance. Here are some steps you could consider to navigate this situation: 1. Open Communication - Talk to Your Sister: Schedule a private conversation where you can express your concerns about the budget without making her feel judged. Let her know how much you appreciate her company and her \ldots
\end{promptbox}
\begin{promptbox}{OLMo-2-1B DPO, reply to the Standard-English request}
Handling the situation of needing to adjust your movie outings due to a canceled budget can be a bit tricky, especially if you're trying to keep your sister happy. Here are some tips to navigate this situation effectively: 1. Communicate Openly with Your Sister Express your concerns to your sister honestly and kindly. Let her know that the recent budget cancellations make it difficult for you to afford movie outings \ldots
\end{promptbox}

\subsection{Preference-Data Audit and DPO Re-runs}
\label{app:dpo}
This section audits the released preference data and details the DPO re-runs.
\begin{table*}[t]
  \centering
  \tabstyle
  \setlength{\tabcolsep}{2.5pt}
  \begin{tabular}{l c ccc c}
    \toprule
    & & \multicolumn{2}{c}{\textbf{Held-out CR}} & \textbf{Trained} & \\
    \cmidrule(lr){3-4}
    \textbf{Model} & \textbf{Runs} & features\hup & varieties\hup & features & $\Split$\hdown \\
    \midrule
    Re-run, dose 0 & \res{e5.d0.n_seeds} & \cellcolor{cCmpA!3}\rcell{e5.d0.crheld} & \rcell{e5.d0.crg} & \rcell{e5.d0.ftrain} & \rcell{e5.d0.split} \\
    \midrule
    + 1k matched pairs & \res{e5.aa1k.n_seeds} & \cellcolor{cCmpA!3}\rcell{e5.aa1k.crheld} & \rcell{e5.aa1k.crg} & \rcell{e5.aa1k.ftrain} & \rcell{e5.aa1k.split} \\
    + 5k matched pairs & \res{e5.aa5k.n_seeds} & \cellcolor{cCmpA!3}\rcell{e5.aa5k.crheld} & \rcell{e5.aa5k.crg} & \rcell{e5.aa5k.ftrain} & \rcell{e5.aa5k.split} \\
    + 20k matched pairs & \res{e5.aa20k.n_seeds} & \cellcolor{cCmpA!5}\rcell{e5.aa20k.crheld} & \rcell{e5.aa20k.crg} & \rcell{e5.aa20k.ftrain} & \rcell{e5.aa20k.split} \\
    \hirow
    + 60k matched pairs & \res{e5.aa60k.n_seeds} & \cellcolor{cCmpA!27}\rcell{e5.aa60k.crheld} & \rcell{e5.aa60k.crg} & \rcell{e5.aa60k.ftrain} & \rcell{e5.aa60k.split} \\
    \midrule
    Control 5k & \res{e5.ctl5k.n_seeds} & \cellcolor{cCmpA!3}\rcell{e5.ctl5k.crheld} & \rcell{e5.ctl5k.crg} & \rcell{e5.ctl5k.ftrain} & \rcell{e5.ctl5k.split} \\
    Control 20k & \res{e5.ctl20k.n_seeds} & \cellcolor{cCmpA!4}\rcell{e5.ctl20k.crheld} & \rcell{e5.ctl20k.crg} & \rcell{e5.ctl20k.ftrain} & \rcell{e5.ctl20k.split} \\
    Control 60k & \res{e5.ctl60k.n_seeds} & \cellcolor{cCmpA!5}\rcell{e5.ctl60k.crheld} & \rcell{e5.ctl60k.crg} & \rcell{e5.ctl60k.ftrain} & \rcell{e5.ctl60k.split} \\
    Anti-arm 1k & \res{e5.anti1k.n_seeds} & \cellcolor{cCmpA!2}\rcell{e5.anti1k.crheld} & \rcell{e5.anti1k.crg} & \rcell{e5.anti1k.ftrain} & \rcell{e5.anti1k.split} \\
    Anti-arm 5k & \res{e5.anti5k.n_seeds} & \cellcolor{cCmpA!2}\rcell{e5.anti5k.crheld} & \rcell{e5.anti5k.crg} & \rcell{e5.anti5k.ftrain} & \rcell{e5.anti5k.split} \\
    \bottomrule
  \end{tabular}
  \caption{\textbf{Accommodation-consistent preference pairs raise convergence only at the largest dose tested.} OLMo-2-1B DPO re-runs by added pairs: convergence on held-out and trained features and the split; mean\std{std} over training runs (Runs); IFEval in \Cref{sec:exp:h5}.}
  \label{tab:mitigation}
\end{table*}

\paragraph{Preference-data audit.}
We ran the retained-feature recognisers over the prompts of the released preference mixes of OLMo-2-1B, OLMo-2-7B and T\"ulu-3-8B. A prompt counts as dialect-bearing if it contains at least one realization of a retained morphosyntactic feature of AAE or Singlish. The shares are \res{dv.pref.olmo2_1b}\%, \res{dv.pref.olmo2_7b}\% and \res{dv.pref.tulu3}\%; on the human-written Dolly-15k instructions, a proxy for the false-positive floor on plain English, the same recognisers fire on \res{dv.pref.floor}\% of prompts. On \res{dv.wild.n} first user turns of WildChat-1M \citep{zhao2024wildchat} (English, non-toxic, not redacted, 20--300 words, no code or URLs), the AAE recognisers fire on \res{dv.wild.aae}\% and the recognisers of Singlish particles and words on \res{dv.wild.sge}\%. In a blind check of \res{audit.auth.flagged.n} sampled hits, the expert panel confirmed \res{audit.auth.aaen.k} of \res{audit.auth.aaen.n} AAE hits (\res{audit.auth.by_feature.aae_ain_t.k} of \res{audit.auth.by_feature.aae_ain_t.n} for \emph{ain't}, \res{audit.auth.by_feature.aae_negative_concord.k} of \res{audit.auth.by_feature.aae_negative_concord.n} for multiple negation) and \res{audit.auth.sgen.k} of \res{audit.auth.sgen.n} Singlish hits, and found the feature in none of \res{audit.auth.ctrl.n} unflagged controls. The rejected hits are mostly templated creative-writing prompts, code and names. On chat text, false positives thus far exceed the Dolly floor, which understates them, and the shares above overstate real dialect use (AAE in WildChat: about \res{audit.auth.wild_aae_corr.est}\%, 95\% CI \res{audit.auth.wild_aae_corr.lo}--\res{audit.auth.wild_aae_corr.hi}\%). We therefore use no recogniser-selected WildChat requests as dialect input. The shares are not exhaustive: the recognisers cover only retained features, and in a check of eight sampled hits by an LLM agent, at most two were dialectal, the others being answer templates (``or not'') or tokenization artefacts (``gon na'').
The mix used for OLMo-2-32B reuses the 7B mix's prompts with completions from the 32B family; its dialect-bearing share is \res{dv.pref.olmo2_32b}\% (7B: \res{dv.pref.olmo2_7b}\%). On AAE-bearing prompts, chosen and rejected replies differ in AAE density by \res{dv.prefwin.32b.diff} (CI [\res{dv.prefwin.32b.lo}, \res{dv.prefwin.32b.hi}]; \res{dv.prefwin.32b.tie}\% ties) against \res{dv.prefwin.7b.diff} (CI [\res{dv.prefwin.7b.lo}, \res{dv.prefwin.7b.hi}]; \res{dv.prefwin.7b.tie}\% ties) for 7B. These are correlational descriptions of the data, not tests of what the update learns.

\paragraph{Re-run construction.}
We initialize from \texttt{allenai/OLMo-2-0425-1B-SFT} and train with the released DPO recipe in open-instruct (length-normalized loss, $\beta=5$, learning rate $2.5\times10^{-6}$, one epoch, 128 pairs per step) on a fixed, seeded 120k-pair subset of the released 1B preference mix; this is dose 0. Added pairs come from a disjoint pool of first-turn requests (Dolly-15k, OASST1). The prompt is rendered in AAE or Singlish using only the training split of the features. The chosen reply is the SFT model's reply to the standard prompt, rendered in the same variety, and the rejected reply is the same reply unrendered, so the two differ only in matching the user. Matched controls use the same prompts with two standard replies, and the anti-arm swaps chosen and rejected.
The split of the features into training and held-out features was fixed by hash before any run. Outcomes are the rate-based convergence ratio on the held-out features of AAE and Singlish and on the held-out varieties Nigerian Pidgin and Appalachian English, the latter sharing Multi-VALUE rules with AAE. Doses 0, 1k, 5k and 20k were pre-registered; 60k was added at the gate (\Cref{app:outcomes}).

\subsection{Extended Discussion}
\label{app:discussion}
This section extends the discussion of \Cref{sec:discussion}.

\paragraph{Outputs show half the picture.}
An audit that inspects replies finds the same Standard-English answer for every user and cannot tell whether training treats varieties alike. The update diagnostic shows that it does not: it moves the policy toward matching some users and away from matching others. This preference does not show in the default output, and content-matched reply pairs read it from public checkpoints without retraining. We recommend that model releases report it next to output audits when consecutive checkpoints are public.

\paragraph{The standard default arises with little dialect-specific data.}
Convergence falls at the preference stage on data in which fewer than 1\% of prompts carry a detectable dialect feature, and the ability to produce AAE features on request survives. The default is thus consistent with the form that general post-training data favour, in line with the mostly stylistic shifts that alignment tuning produces \citep{lin2024unlocking}. Deployed chatbots that answer Nigerian Pidgin in kind (\Cref{sec:exp:h1}) show that standardizing a stigmatized variety is not inevitable after post-training. Accommodation-consistent preference pairs raise convergence, though without calibration to the user (\Cref{sec:exp:h5}), and some capacity to produce the features remains accessible on request.

\paragraph{Prestige differences persist in the update.}
CAT predicts convergence toward prestige. After alignment, the replies of our models show little convergence of any kind, but after adjustment the updates of \res{dv.txt.adj} lineages still favour matching prestige spelling over stigmatized varieties, and the ordering of behaviour by prestige is already present at the base checkpoints. The ordering is consistent with prestige-directed convergence but does not identify prestige as its cause. Raw and adjusted contrasts answer different questions: OLMo-2-32B's update up-weights unlikely replies in general, so its contrast is \res{e4i.dpo_vs_sft.per_lineage.olmo2_32b.raw} on the observed pairs and \res{e4a.olmo2_32b.adj} after adjustment, and diagnostics of an update should report both. Aligned chat models are thus interlocutors whose surface behaviour is uniform while their learned preferences are not. Differences in reply formality between vernacular users and equally informal Standard-English users are small (\Cref{app:outcomes:register}).

\subsection{Further Results Details}
\label{app:results_more}
Details of the results in \Cref{sec:replies,sec:updates}.

\paragraph{Contribution 1 in numbers.} The preference stage lowers convergence to informal register on \res{dv.txt.inf_dpo} lineages and to British spelling on \res{dv.txt.bre_dpo}, and deployed chatbots standardize AAE too. After adjusting for baseline likelihood and length, the pooled contrast is \res{e4i.dpo_vs_sft.adjusted.coef} nats per token, and it is positive in \res{e4a.breadth_adj_pos} of \res{e4a.breadth_n} further updates.

\paragraph{Contribution 2 in numbers.} The gain appears in all \res{dv.e560.n} DPO re-runs at \res{dv.e5dose.share_60k}\% of the pairs, but not at the lower doses tested.

\paragraph{Contribution 3 in numbers.} On OLMo-2-1B, CAP raises held-out convergence by \res{dv.capres.b1.cap_crheld} over matched controls; one-sided pairs from the same builder gain \res{dv.capabl.mx_crheld} but insert \res{dv.capabl.mx_ins} features per 100 words against \res{dv.capabl.cap_ins} for CAP.

\paragraph{OLMo-2-7B in detail.} Under the chat template that the later stages use, the base model reproduces \res{e1.l2.base_f1.inf.cr} of a user's informality (\res{e1.l2.base.inf.cr} in the transcript framing of \Cref{fig:heatmap}), after SFT \res{e1.l2.sft.inf.cr}, and after DPO \res{e1.l2.dpo.inf.cr}. After DPO, AAE detector hits are as rare in replies to AAE requests (density \res{e1.l2.dpo.aae.d_cond}) as in replies to their Standard-English twins (\res{e1.l2.dpo.aae.d_sae}; \Cref{sec:validity}).

\paragraph{Stage attribution.} The preference stage, which keeps the chat template fixed, lowers convergence to informal register on \res{dv.txt.inf_dpo} lineages and to British spelling on \res{dv.txt.bre_dpo}; on OLMo-3.1-32B, spelling convergence stays level (\res{e1.olmo31_32b.sft.bre.cr_rate} after SFT, \res{e1.olmo31_32b.dpo.bre.cr_rate} after DPO). The step from base to SFT also changes how the model is prompted, so we count a decline only where it holds under all three base framings: for spelling on \res{dv.n_bre_sft_falls_allframes} of the \res{dv.n_lineages_allframes} lineages run in all three framings, and for informal register on \res{dv.n_inf_sft_falls_allframes} (\Cref{app:details}; OLMo-2-7B in \Cref{tab:controls}).
The split between prestige and stigmatized cues falls from the first released stage to DPO on \res{dv.n_split_falls} of \res{dv.n_lineages} lineages, under the rate-based ratio as well (\res{dv.n_split_rate_falls}).

\paragraph{Dialect grammar after SFT.} After SFT both are near zero under either ratio (\Cref{tab:mainfull}). Post-trained replies to human-written AAE tweets, paired with human Standard-English translations, are near zero too (OLMo-2-7B after DPO: \res{au.l2.dpo.auth_gro.aae.cr}; \Cref{app:authentic}).

\paragraph{Scale.} OLMo-2-32B, one of two 32B lineages with every stage released, reproduces British spelling at \res{e1.olmo2_32b.base.bre.cr_rate} at base and \res{e1.olmo2_32b.dpo.bre.cr_rate} after DPO, and Qwen2.5-72B at \res{e1.qwen25_72b.base.bre.cr_rate} in its base release and \res{e1.qwen25_72b.instruct.bre.cr_rate} in its instruct release (\Cref{tab:mainfull}).

\paragraph{Popular families.} The exceptions to the spelling decline are \res{dv.fam.bre_rise_names}, whose British-spelling convergence starts low and rises slightly. Both Qwen3.5 models follow the pattern.

\paragraph{openPangu.} openPangu-Embedded-7B-V1.1, a chat release without an open base, standardizes like the open models: informal register \res{e1o.openpangu_embedded_7b_v11.chat_enprompt.inf.cr}, AAE \res{e1o.openpangu_embedded_7b_v11.chat_enprompt.aae.cr} and Nigerian Pidgin \res{e1o.openpangu_embedded_7b_v11.chat_enprompt.ngp.cr_rate} (\Cref{tab:mainfull}).

\paragraph{Competence details.} Singlish features are realized at \res{dv.comp_corr.l2_sge.comp} on OLMo-2-7B. The rates are corrected by a blind human audit: as pre-declared, each feature's detector rate is multiplied by the detector precision on audited requested replies before capping at the authentic rate (the panel confirmed \res{audit.v4.b.primary.precision.k} of \res{audit.v4.b.primary.precision.n} detector hits in requested replies; \Cref{app:detectors}).

\paragraph{Adjusted contrast.} We compare the matching bonus of the two prestige cues with that of the two stigmatized varieties for the same request, adjusting for how likely each matched reply already was under the SFT model and for its length.

\paragraph{Reading a negative bonus.} For the sister example, a negative bonus means that the update makes \emph{who keep dragging you} less likely against \emph{who keeps dragging you} for an AAE user than for a Standard-English user.

\paragraph{OLMo-2-32B.} The raw contrast describes the update on the observed reply pairs; the adjusted one compares varieties at the same SFT likelihood and length. Unadjusted, the DPO update of OLMo-2-32B favours matching every variety, Singlish most (\res{e4.olmo2_32b.dpo_vs_sft.sge.mb}), so its raw contrast reverses (\res{e4i.dpo_vs_sft.per_lineage.olmo2_32b.raw}). Its update up-weights replies that the SFT model found unlikely far more than any other (slope \res{e4a.olmo2_32b.ll_slope} per SD of log-likelihood, against at most \res{e4a.max_abs_ll_slope_others} in magnitude elsewhere), and stigmatized renderings are less likely. Adjusted, its contrast is positive (\res{e4a.olmo2_32b.adj}, CI [\res{e4a.olmo2_32b.adj_lo}, \res{e4a.olmo2_32b.adj_hi}]), and within OLMo-2 the adjusted contrast shows no trend with size (\Cref{tab:sizes}). The other 32B lineage, OLMo-3.1-32B, does not reverse (raw contrast \res{e4i.dpo_vs_sft.per_lineage.olmo31_32b.raw}), and its update favours unlikely replies far less (slope \res{e4a.olmo31_32b.ll_slope}), so the effect is not a matter of size alone. OLMo-2-32B's preference mix reuses the 7B prompts, with the same dialect-bearing share (\res{dv.pref.olmo2_32b}\%) and the same near-tie between chosen and rejected replies (\Cref{app:dpo}).
\begin{table*}[t]
  \centering
  \footnotesize
  \setlength{\tabcolsep}{3pt}
  \begin{tabular}{l l cccccc}
    \toprule
    \textbf{Size} & \textbf{Release} & $\Delta$ raw & $\Delta$ adjusted & likelihood slope & $\MB{AAE}$ & $\MB{SGE}$ & $\MB{BRE}$ \\
    \midrule
    OLMo-2-1B & 0425 & 0.4 {\scriptsize[<0.1, 0.9]} & \cellcolor{cCmpA!18}1.3 {\scriptsize[0.8, 1.9]} & \ensuremath{-}0.5 & \ensuremath{-}0.9 {\scriptsize[\ensuremath{-}1.2, \ensuremath{-}0.6]} & 0.5 {\scriptsize[\ensuremath{-}0.1, 1.1]} & 0.4 {\scriptsize[0.1, 0.7]} \\
    OLMo-2-7B & 1124 & 1.4 {\scriptsize[0.3, 2.4]} & \cellcolor{cCmpA!27}2.7 {\scriptsize[1.7, 3.7]} & \ensuremath{-}0.9 & \ensuremath{-}1.8 {\scriptsize[\ensuremath{-}2.4, \ensuremath{-}1.2]} & \ensuremath{-}0.7 {\scriptsize[\ensuremath{-}1.7, 0.1]} & 0.2 {\scriptsize[\ensuremath{-}0.5, 0.9]} \\
    OLMo-2-13B & 1124 & 1.5 {\scriptsize[0.8, 2.2]} & \cellcolor{cCmpA!17}1.2 {\scriptsize[0.3, 2.1]} & 0.3 & \ensuremath{-}2.1 {\scriptsize[\ensuremath{-}2.6, \ensuremath{-}1.6]} & \ensuremath{-}1.5 {\scriptsize[\ensuremath{-}2.3, \ensuremath{-}0.8]} & \ensuremath{-}0.2 {\scriptsize[\ensuremath{-}0.7, 0.2]} \\
    OLMo-2-32B & 0325 & \ensuremath{-}3.1 {\scriptsize[\ensuremath{-}4.2, \ensuremath{-}2.1]} & \cellcolor{cCmpA!17}1.2 {\scriptsize[0.2, 2.2]} & \ensuremath{-}3.4 & 2.0 {\scriptsize[1.5, 2.5]} & 7.1 {\scriptsize[5.7, 8.5]} & 1.7 {\scriptsize[1.2, 2.4]} \\
    \bottomrule
  \end{tabular}
  \caption{\textbf{After adjustment, the OLMo-2 update favours prestige spelling at every size.} DPO update, $\times$100, 95\% CIs; slope: change per SD of SFT log-likelihood.}
  \label{tab:sizes}
\end{table*}

\paragraph{SFT and RLVR updates.} From base to SFT the raw contrast is positive on OLMo-2-1B, OLMo-2-7B and OLMo-3.1-32B (\res{e4i.sft_vs_base.per_lineage.l1.raw}, \res{e4i.sft_vs_base.per_lineage.l2.raw} and \res{e4i.sft_vs_base.per_lineage.olmo31_32b.raw}, all with CIs above zero), and larger than at DPO on the two OLMo-2 lineages; on Zephyr-7B-$\beta$ it appears only after adjustment (raw \res{e4i.sft_vs_base.per_lineage.zephyr_7b.raw}, adjusted \res{e4a.sftbase.zephyr_7b.adj}). The RLVR stage barely moves it (OLMo-2-7B, AAE: \res{e4.l2.instruct_vs_dpo.aae.mb}).

\paragraph{Popular families.} Over all of post-training, these families' bonus for matching AAE is slightly positive, but matching prestige spelling gains more (\Cref{tab:main}). Qwen3.5-9B's update favours prestige spelling after adjustment (\res{e4a.breadth.qwen35_9b_instruct_vs_base.adj}) although its raw contrast is negative.

\paragraph{Dialect-bearing prompts.} For comparison, \res{dv.pref.floor}\% of human-written instructions trigger the same recognisers (\Cref{app:dpo}).

\paragraph{Per-run trained features at 60k.} Pooled over AAE and Singlish, only the first run over-produces the trained features, at \res{dv.e560.ftrain_s0} times the user's own rate (others: \res{dv.e560.ftrain_s1}, \res{dv.e560.ftrain_s2} and \res{dv.e560.ftrain_s3}).

\paragraph{Reverse pairs.} Reverse pairs, which prefer the standard reply, leave convergence unchanged (\Cref{app:outcomes}).

\paragraph{Ratio versus calibration.} A higher ratio means more features in replies to dialect requests than to their Standard-English twins, not replies at the user's density.

\paragraph{Sensitivity of the two findings.} Changing one analysis choice at a time leaves both findings in place (\Cref{tab:sensitivity} in \Cref{app:results}). The raw contrast is weaker than the adjusted one, because OLMo-2-32B's update up-weights unlikely replies in general, but its point estimate is positive on \res{dv.n_int_pos} of \res{dv.n_int} lineages. A stratified bootstrap, replies of 1{,}024 tokens and a fresh set of sampling seeds change no conclusion, and the human-audit correction lowers the requested AAE rate only slightly. The step from base to SFT also changes how the model is prompted, so there we count a decline only where it holds under all three framings.
\begin{table}[!htb]
  \centering
  \scriptsize
  \setlength{\tabcolsep}{2pt}
  \begin{tabular}{@{}p{0.2\columnwidth} p{0.27\columnwidth} p{0.49\columnwidth}@{}}
    \toprule
    \rowcolor{gray!10} \textbf{Choice} & \textbf{Alternative} & \textbf{Result} \\
    \midrule
    \multicolumn{3}{@{}l}{\textit{Selective update (\Cref{sec:updates})}} \\
    Update estimator & raw paired contrast & pooled \res{e4i.dpo_vs_sft.raw_mean_over_lineages} [\res{e4i.dpo_vs_sft.raw_lo}, \res{e4i.dpo_vs_sft.raw_hi}], positive on \res{dv.n_int_pos} of \res{dv.n_int}; adjusted \res{e4i.dpo_vs_sft.adjusted.coef}, on \res{dv.txt.adj} \\
    Bootstrap & stratified by task type & \res{dv.txt.strat} adjusted CIs above 0, \res{e4a.strat.min_width_ratio}--\res{e4a.strat.max_width_ratio}$\times$ as wide \\
    \addlinespace[0.5mm]
    \multicolumn{3}{@{}l}{\textit{Standardized replies (\Cref{sec:replies})}} \\
    Base framing & chat template; raw continuation & base-to-SFT decline under all three framings: spelling \res{dv.n_bre_sft_falls_allframes} of \res{dv.n_lineages_allframes}, register \res{dv.n_inf_sft_falls_allframes} of \res{dv.n_lineages_allframes}; DPO keeps the template \\
    Reply length & 1{,}024 instead of 400 tokens & convergence changes by at most \res{dv.cap.max_abs_d_used} \\
    Sampling seeds & a fresh set, \res{dv.s5.n_lin} lineages & within \res{dv.s5.max_abs_d}; DPO declines on \res{dv.txt.s5_bre} \\
    Detector precision & human-audit correction & requested AAE rate \res{comp.l2.dpo.aae.comp} $\to$ \res{dv.comp_corr.l2_aae.comp} (OLMo-2-7B); default false hits cancel (\res{rh.aligned.aae.aae_hits} vs.\ \res{rh.aligned.sae.aae_hits}) \\
    \bottomrule
  \end{tabular}
  \caption{\textbf{The findings hold under other analysis choices.} Each row changes one choice of the main analysis.}
  \label{tab:sensitivity}
\end{table}

\section{Pre-registration and Analysis Details}
The theory and hypotheses we pre-registered, every prediction with its outcome, and the statistical procedures.

\subsection{Theory and Pre-Registered Hypotheses}
\label{app:theory}\label{sec:theory}
This section states the theory we test and the hypotheses we pre-registered.
\begin{table*}[t]
  \centering
  \small
  \setlength{\tabcolsep}{3pt}
  \renewcommand{\arraystretch}{1.1}
  \begin{tabularx}{\textwidth}{@{}>{\raggedright\arraybackslash}p{4.2cm} >{\raggedright\arraybackslash}X c@{}}
    \toprule
    \rowcolor{gray!10} \textbf{CAT construct} & \textbf{Measurement on model replies} & \textbf{Hyp.} \\
    \midrule
    Convergence / maintenance / divergence & $\CR{v}>0$ / $\approx 0$ / $<0$ on novel spans; $\dF{v}>\dF{INF}$ & \Cref{hyp:select} \\
    Upward asymmetry & split index $\Split$; $\mathrm{CR}$ ordered by pre-registered prestige; lexical \vs morphosyntactic & \Cref{hyp:select} \\
    Where it is learned & $\Split$ by stage (base, SFT, DPO, RLVR) within a lineage & \Cref{hyp:locus} \\
    Psychological \vs linguistic & competence $\COMP{v}$ on explicit request \vs spontaneous $\CR{v}$ & \Cref{hyp:choice} \\
    Motivation (approval) & matching bonus $\MB{v}$ in the update between released stages & \Cref{hyp:mech} \\
    Learned, hence changeable & dose--response of $\Split$ in DPO re-runs & \Cref{hyp:cause} \\
    Attitude $\leftrightarrow$ behaviour & $\CR{v}$ \vs matched-guise stereotype score (not run) & \hyp{6} \\
    Measurement validity & agreement with gold and ratings; copy and length sensitivity & \hyp{7} \\
    \bottomrule
  \end{tabularx}
  \caption{\textbf{Every pre-registered CAT construct has a measurable counterpart on model replies.} The CAT construct, the quantity we measure (\Cref{sec:method}), and the hypothesis it tests.}
  \label{tab:operationalization}
\end{table*}


\paragraph{Communication Accommodation Theory.}
CAT holds that speakers adjust their speech to their interlocutors for social reasons \citep{giles1991accommodation,giles2016communication}.
\emph{Convergence} reduces linguistic differences to seek approval or affiliation; \emph{divergence} accentuates them to mark distinctiveness or status; and \emph{maintenance} keeps one's own variety.
Three tenets make the theory testable on language models \citep{dragojevic2016accommodative}.
\circled{1} Convergence is \emph{asymmetric} and flows upward, toward the higher-status variety.
\circled{2} \emph{Over-accommodation}, meaning caricatured or patronizing convergence, is perceived as demeaning.
\circled{3} \emph{Psychological} accommodation (intent) differs from \emph{linguistic} accommodation (the forms produced), so a speaker may be able to converge and choose not to.
Mechanistic accounts instead derive convergence from largely automatic priming \citep{pickering2004toward}, though lexical entrainment can be partner-specific \citep{brennan1996conceptual}. A pretrained model continuing a user turn is such a priming device.
The two accounts therefore make different predictions for language models: priming predicts convergence to every variety, while CAT predicts convergence that depends on the social value of the variety. We use CAT only as a source of such predictions; we make no claim about model intentions.

\paragraph{Standard-language ideology supplies the ordering.}
Standard-language ideology holds that one variety is correct and that the others are errors \citep{milroy2001language,milroy2012authority}. Institutions reproduce it by rewarding the standard \citep{lippigreen2012english}. Its stigma falls on racialized and postcolonial speakers even when they use standard forms \citep{rosa2017unsettling}, and matched-guise experiments measure it as covert evaluation from speech alone \citep{lambert1960evaluational,hofmann2024ai}.
\citet{smith2026standard} argue that generative AI reproduces this ideology; we measure it.
Under the ideology, upward convergence has a predictable shape. Models should converge to the standard and to prestige cues, including an informal but standard register. They should maintain or diverge from stigmatized morphosyntax, which the ideology classes as ``wrong grammar''. And they should keep the competence to produce it.

\noindent\emph{Is the standard-English default of chat models learned by imitating standard responses, or installed by preferring them?}

\begin{claimbox}
CAT separates what a speaker produces from what the speaker is disposed to do. We therefore read accommodation on two channels at every training stage: the \emph{reply} (does it converge toward the user's variety?) and the \emph{update} (does the training step favour converging?), and we ask whether either channel is selective by the prestige of the user's variety.
\end{claimbox}

\paragraph{From theory to measurement.}
\Cref{tab:operationalization} maps each construct to one quantity on replies, defined in \Cref{sec:method}.
The \emph{convergence ratio} $\CR{v}$ is the share of the user's variety-feature density, over the matched Standard American English (\vty{SAE}) request, that the reply reproduces on novel spans. Values above, near and below zero mean convergence, maintenance and divergence. A reply that is more formal to a $v$-speaker than to an equally informal \vty{SAE} speaker ($\dF{v}>\dF{INF}$, with $\dF{v}$ the same difference of means over the part-of-speech formality score of \citet{heylighen2002variation}) also diverges.
Selectivity is the \emph{split index} $\Split$: mean CR over prestige cues minus mean CR over stigmatized varieties, with tiers pre-registered from attitude studies. Competence $\COMP{v}$ separates psychological from linguistic accommodation.

\noindent We pre-registered five primary and two secondary predictions with falsifiers (\Cref{app:prereg}).

\begin{hypothesis}[Locus]\label{hyp:locus}
The split grows more from SFT to DPO than from base to SFT, and reinforcement learning with verifiable rewards (RLVR) does not reverse it, on at least 2 of 3 fully open lineages.
\emph{Falsified if} SFT carries $\ge$80\% of the final split, if the split grows equally at every stage, or if there is no split.
\end{hypothesis}

\noindent The falsifiers are pre-registered alternatives, reported as the finding if they win: ($\alpha$) the locus is imitation at SFT; ($\beta$) the split accumulates across stages, as performance gaps do \citep{elle2026dialect}; ($\gamma$) all informal input is normalized uniformly; and ($\delta$) no stage, not even the base model under any framing, converges to dialect grammar, while register convergence persists. Under $\delta$, selectivity belongs to pretrained chat behaviour, and the locus question moves to competence (\Cref{hyp:choice}) and reward (\Cref{hyp:mech}).
None is read before the validity checks pass: per-feature false positives $\le$2\% on replies to \vty{SAE} prompts at every stage, at least 3 opportunities per reply, and at least 60\% retention (\Cref{sec:method}).

\begin{hypothesis}[Selectivity]\label{hyp:select}
At aligned stages $\CR{INF},\CR{BRE}>0$ and $\CR{v}<\CR{INF}$ for stigmatized $v$, with $\dF{v}>\dF{INF}$, lexis above morphosyntax, and CR ordered by prestige.
\emph{Falsified if} $\rho(\CR{v},\mathrm{tier}_v)<0.3$ in most aligned models, or if stigmatized varieties and \vty{INF} fall equally.
\end{hypothesis}

\begin{hypothesis}[Can but won't]\label{hyp:choice}
On explicit request, aligned models realize stigmatized features at $\ge$50\% of authentic rates for $\ge$70\% of features, and the gap $\COMP{v}-\CR{v}$ widens across stages.
\emph{Falsified if} competence is below 50\% for most features and varieties.
\end{hypothesis}

\begin{hypothesis}[Mechanism]\label{hyp:mech}
At identical content, the DPO update diagnostic and released reward models give a matching bonus $\MB{v}>0$ to prestige cues and $\MB{v}\le 0$ to stigmatized varieties.
\emph{Falsified if} $\MB{v}\ge 0$ for stigmatized varieties, or if there is no prestige ordering.
\end{hypothesis}

\begin{hypothesis}[Cause]\label{hyp:cause}
Re-running DPO from the SFT checkpoint with a dose of accommodation-consistent pairs raises stigmatized convergence monotonically in the dose, at matched helpfulness.
\emph{Falsified if} Spearman over doses is $<0.8$, or if IFEval drops by more than 2 points.
\end{hypothesis}

\noindent Two secondary predictions concern validity and attitudes (\Cref{sec:validity,app:outcomes}): that per-variety convergence tracks covert matched-guise prejudice, or dissociates from it across stages (\hyp{6}), and that CR beats prior accommodation measures on validity (\hyp{7}).

\noindent The rival account, \emph{helpfulness normalization} (all non-standard input standardized equally), is the null of \Cref{hyp:select}. All pre-registered outcomes, including those that did not hold, are reported in \Cref{app:outcomes}.

\subsection{Pre-registered Predictions and Outcomes}
\label{app:outcomes}
We report every prediction frozen in the pre-registration (\Cref{app:prereg}) with its outcome, including those that did not hold (\Cref{tab:outcomes}). Numbers are from the same result files as the main text (local sync of 2026-10-01; OLMo-2-7B unless stated).
\begin{table*}[t]
  \centering
  \small
  \setlength{\tabcolsep}{3pt}
  \renewcommand{\arraystretch}{1.1}
  \begin{tabularx}{\textwidth}{@{}>{\raggedright\arraybackslash}p{2.6cm} >{\raggedright\arraybackslash}X >{\raggedright\arraybackslash}p{2.3cm}@{}}
    \toprule
    \textbf{Prediction} & \textbf{Evidence} & \textbf{Outcome} \\
    \midrule
    \hyp{1} Locus: the prestige--stigma split grows more at SFT$\to$DPO than at base$\to$SFT &
    $\Split$ falls instead: base \res{e1.l2.base.split}, SFT \res{e1.l2.sft.split}, DPO \res{e1.l2.dpo.split}; it falls on \res{dv.n_split_falls} of \res{dv.n_lineages} lineages (rate-based version: \res{dv.n_split_rate_falls}), with the larger step at SFT on \res{dv.n_sft_larger_prereg} of the \res{dv.n_lineages_base} lineages with a released base model (rate-based: \res{dv.n_sft_larger_rate}). Closest pre-registered alternative: $\gamma$ (uniform decline) with an SFT-weighted locus ($\alpha$) &
    not supported \\
    $\delta$ No dialect convergence at any stage, including base &
    Base raw continuation mirrors AAE grammar at \res{e1.l2.base_raw.aae.cr} \rci{e1.l2.base_raw.aae.cr} and Singlish lexis at \res{e1.l2.base_raw.sge.cr_lex} \rci{e1.l2.base_raw.sge.cr_lex}, both above 0 &
    not met \\
    \hyp{2} Selectivity in convergence: $\CR{v}<\CR{INF}$ for stigmatized $v$ at aligned stages &
    Informal-register convergence also collapses (DPO \res{e1.l2.dpo.inf.cr} \rci{e1.l2.dpo.inf.cr}), so the contrast with stigmatized varieties (CR $\approx 0$) is near zero, not negative. Prestige ordering: $\rho=\res{e1.l2.base.rho}$ at base and \res{e1.l2.dpo.rho} after DPO (bar $\geq 0.6$). Formality channel, with the validated ranker: more formal in \res{dv.dfr.n_div_sft} of \res{dv.dfr.n_cells_sft} cells after SFT and less formal in \res{dv.dfr.n_opp_dpo} of \res{dv.dfr.n_cells_dpo} after DPO (\Cref{app:outcomes:register}) &
    not supported (convergence or register) \\
    \hyp{3} Competence $\geq$50\% of authentic rates for $\geq$70\% of features &
    Human-corrected COMP (\Cref{app:detectors}), AAE: OLMo-2-7B DPO \res{dv.comp_corr.l2_aae.comp} (detector \res{comp.l2.dpo.aae.comp}); T\"ulu-3-8B \res{dv.comp_corr.l4_aae.comp}; OLMo-3-7B \res{dv.comp_corr.l3_aae.comp}; Singlish \res{dv.comp_corr.l2_sge.comp}. Retained AAE features at $\geq$50\% of authentic rates: \res{dv.h3.l2_aae.k_ret_corr}, \res{dv.h3.l4_aae.k_ret_corr} and \res{dv.h3.l3_aae.k_ret_corr} of \res{dv.h3.l2_aae.n_ret} (detector rate: \res{dv.h3.l2_aae.k_ret}, \res{dv.h3.l4_aae.k_ret}, \res{dv.h3.l3_aae.k_ret}); Singlish at most \res{dv.h3.l3_sge.k_ret} of \res{dv.h3.l3_sge.n_ret}. Falsifier (a majority of features below 50\% in a majority of varieties) not met &
    partly: the $\geq$70\% criterion holds at the detector rate on OLMo-2-7B only, at the corrected rate on none \\
    \hyp{4} Matching diagnostic $>0$ for prestige cues, $\le 0$ for stigmatized &
    AAE negative on \res{dv.n_mb_aae_neg} of \res{dv.n_lineages}; prestige spelling not penalized on \res{dv.n_mb_prestige_spared}; positive for prestige only on some lineages (e.g.\ T\"ulu-3-8B \res{e4.l4.dpo_vs_sft.bre.mb} \rci{e4.l4.dpo_vs_sft.bre.mb}). Pre-registered mixed model (P4: $\MB{} \sim \mathrm{tier} + (1|\mathrm{request}) + (1|\mathrm{variety})$): tier effect positive on \res{dv.n_mixed_tier_pos} of \res{dv.n_lineages} lineages, CI above zero on \res{dv.n_mixed_tier_sig}; pooled \res{e4m.dpo_vs_sft.pooled.tier} (CI [\res{e4m.dpo_vs_sft.pooled.tier_lo}, \res{e4m.dpo_vs_sft.pooled.tier_hi}]), with the variety component weakly identified by four varieties. Paired within-request contrast (added after pre-registration): at equal baseline likelihood positive on \res{e4a.n_adj_pos} of \res{e4a.n_lin} lineages; unadjusted on \res{dv.n_int_pos} of \res{dv.n_int}, OLMo-2-32B reversing (\Cref{tab:update,tab:sizes}). Released reward models: not run &
    supported for the stigmatized side; the prestige ordering in the paired test only \\
    \hyp{5} DPO re-run: monotone dose--response at matched helpfulness & Pre-registered doses: held-out features, treatment minus matched control, \res{e5t.crheld.effect_5k_T_minus_C.mean} at 5k and \res{e5t.crheld.effect_20k_T_minus_C.mean} at 20k (CI [\res{e5t.crheld.effect_20k_T_minus_C.lo}, \res{e5t.crheld.effect_20k_T_minus_C.hi}]); dose slope \res{e5t.crheld.interaction_slope_T_minus_C_per10k.mean} per 10k (one-sided $p=\res{e5t.crheld.interaction_slope_T_minus_C_per10k.p_one_sided}$). The split's slope (\res{e5t.split.interaction_slope_T_minus_C_per10k.mean} per 10k) is driven by the control arm (control minus dose 0 at 20k: \res{e5t.split.exploratory_20k_C_minus_d0.mean}). Added dose (gated on uptake at 20k, \res{e5t.ftrain.effect_20k_T_minus_C.mean} on trained features, but the order of check and launch is not verified; \res{dv.e560.n} training runs): at 60k, held-out features \res{e5t.crheld.effect_60k_T_minus_C.mean} on average (CI [\res{e5t.crheld.effect_60k_T_minus_C.lo}, \res{e5t.crheld.effect_60k_T_minus_C.hi}]; runs \res{dv.e560.eff_s0}, \res{dv.e560.eff_s1}, \res{dv.e560.eff_s2} and \res{dv.e560.eff_s3}, with CIs [\res{dv.e560.lo_s0}, \res{dv.e560.hi_s0}], [\res{dv.e560.lo_s1}, \res{dv.e560.hi_s1}], [\res{dv.e560.lo_s2}, \res{dv.e560.hi_s2}] and [\res{dv.e560.lo_s3}, \res{dv.e560.hi_s3}]), held-out varieties \res{e5t.crg.effect_60k_T_minus_C.mean}, trained features \res{dv.e560.ftrain_s0}, \res{dv.e560.ftrain_s1}, \res{dv.e560.ftrain_s2} and \res{dv.e560.ftrain_s3} times the user's rate (over-production in the first run only); slope over all four doses \res{e5t.crheld.interaction_slope4_T_minus_C_per10k.mean} per 10k. Reverse pairs: no change. IFEval (prompt-level strict): \res{dv.ifeval.d0.strict} at dose 0, \res{dv.ifeval.t20k.strict} at 20k, \res{dv.ifeval.t60k.strict} at 60k (mean of \res{dv.ifeval.t60k.n} runs: \res{dv.ifeval.t60k.strict_s0}, \res{dv.ifeval.t60k.strict_s1}, \res{dv.ifeval.t60k.strict_s2}, \res{dv.ifeval.t60k.strict_s3}; matched controls \res{dv.ifeval.c60k.strict}), a drop of \res{dv.ifeval.drop60} points, within the pre-registered 2-point bound for the arm mean, though one run drops \res{dv.ifeval.maxdrop60_run} & not supported at the pre-registered doses; an effect at the added dose only \\
    \hyp{6} Convergence tracks covert prejudice & matched-guise runs not performed; compute went to the stage-resolved lineages & not run \\
    \hyp{7} CR beats prior measures on validity criteria & Prior style measures re-run on our replies change by less than .04 between varieties and stages (\Cref{tab:measures}); the pre-registered comparison on gold density and reference ratings was not run, because no reference ratings were collected & partly tested \\
    G0 validity gate: retention $\geq$60\% &
    passes for register, AAE (aligned stages) and Appalachian English; fails for opportunity-based spelling (DPO retention \res{e1.l2.dpo.bre.retention}) and Singlish grammar, which are therefore read with the rate-based ratio &
    partial \\
    Preference-data regime & recogniser hits below the pre-registered threshold even if every hit is dialectal (\Cref{app:dpo}) & Regime B \\
    \bottomrule
  \end{tabularx}
  \caption{\textbf{Outcomes of all pre-registered predictions.} Evidence keys as in the main text; 95\% bootstrap CIs over prompts in brackets. ``Not supported'' and ``not met'' predictions are reported in full here and summarized in one sentence in the main text.}
  \label{tab:outcomes}
\end{table*}

\subsubsection{Register Divergence Is Not Confirmed}
\label{app:outcomes:register}
CAT predicts that divergence can show in register as well as in variety features. We compared the formality of replies to each variety with replies to an equally informal Standard-English user ($\dF{v}-\dF{INF}$; paired sign-flip test, Holm-corrected), scoring formality with the GYAFC-trained ranker that the protocol named, run after labelling and validated against the human audit (\res{audit.r.trimmed.k} of \res{audit.r.trimmed.n}; \Cref{app:detectors}).
The pre-declared test, more formal replies in at least a quarter of the stigmatized-variety cells after SFT and none significantly less formal, fails: \res{dv.dfr.n_div_sft} of \res{dv.dfr.n_cells_sft} cells are more formal after SFT and \res{dv.dfr.n_opp_sft} less formal. After DPO, \res{dv.dfr.n_div_dpo} of \res{dv.dfr.n_cells_dpo} cells are more formal and \res{dv.dfr.n_opp_dpo} less formal (OLMo-2-7B, AAE: \res{g2r.l2.dpo.aae.dF_v_minus_dF_inf}, CI [\res{g2r.l2.dpo.aae.lo}, \res{g2r.l2.dpo.aae.hi}]), so replies to vernacular users are, if anything, slightly less formal; the differences are small on the ranker's 0--1 scale. In base models the difference is positive in \res{dv.dfr.n_base_div_pos} of \res{dv.dfr.n_base_cells} cells.
The part-of-speech score of \citet{heylighen2002variation}, which failed its human validation, had suggested divergence after SFT (\res{dv.n_df_div_sft} of \res{dv.n_df_cells_sft} cells); the validated ranker does not confirm it.

\subsection{Statistical Procedures}
\label{app:stats}
All intervals are percentile 95\% CIs from a cluster bootstrap over the \res{design.n_seeds} requests (2{,}000 resamples, fixed random seed), not stratified by task type as pre-registered (\Cref{app:prereg}). Table $\pm$ values are standard deviations over the three sampling seeds, or over training runs for the DPO re-runs. Contrasts between stages of the split resample requests jointly across stages.
Paired contrasts between conditions use one-sided sign-flip permutation tests over requests (10{,}000 permutations); the register tests are Holm-corrected over the varieties tested within each model and stage (\Cref{app:outcomes:register}).
The update diagnostic is analysed in two ways. (i) The pre-registered mixed model $\MB{} \sim \mathrm{tier} + (1|\mathrm{request}) + (1|\mathrm{variety})$ (registered with requests called seeds), with tiers AAE = Singlish $<$ Australian $<$ British, is fitted by REML per lineage and pooled with lineage fixed effects. (ii) The paired within-request contrast of prestige minus stigmatized matching bonuses (\vty{BRE}, \vty{AUS} minus \vty{AAE}, \vty{SGE}) is adjusted by OLS, $\MB{} = \beta_0 + \beta_1\,\mathrm{prestige} + \beta_2\,z(\ell_{\mathrm{SFT}}) + \beta_3\,z(\log n_{\mathrm{tok}}) + \varepsilon$, one row per request and cue, with $\ell_{\mathrm{SFT}}$ the matched reply's mean log-likelihood under the SFT model and $z(\cdot)$ standardization within lineage (lineage fixed effects when pooled), with a two-level bootstrap over lineages and requests. We added (ii) after the pre-registration.
For the DPO re-runs, treatment-minus-control and treatment-minus-dose-0 effects use the request bootstrap on run means, and dose slopes are fitted to the treatment-minus-control differences. Human-audit proportions use Wilson intervals; Cohen's $\kappa$ and Gwet's AC1 use a bootstrap over items.

\subsubsection{Pre-registration}
\label{app:prereg}
We froze the pre-registration on 2026-10-01 at 02:51 HKT, before any main-run output existed; an outage delayed the planned freeze of Sep 30, 22:00. It fixes the prestige tiers with their sources, the feature inventories (in two steps, the second frozen before first use), the hypotheses with their falsifiers and alternatives (\Cref{app:theory}), the validity checks, the primary estimands and the exclusions. The frozen file (content SHA-256 \texttt{a3cb680d\ldots}) and its deviation log are released.
Deviations after the freeze are as follows.
\begin{itemize}[leftmargin=*,itemsep=0pt,topsep=2pt]
\item The second-step feature lists were added before first use.
\item There are 287 requests instead of 300.
\item Detector bugs were fixed before any main run.
\item The human audit was scored against the formality score actually used, and it included two Singlish features outside its specification (\Cref{app:detectors}).
\item The LLM reference-label pass was dropped.
\item The paired test of the update diagnostic was added.
\item A 60k dose was added to the DPO re-runs (\Cref{app:outcomes}).
\item The bootstrap resamples requests without the pre-registered stratification by task type. Resampling within task type leaves \res{dv.txt.strat} adjusted contrasts of \Cref{sec:updates} above zero, with intervals \res{e4a.strat.min_width_ratio}--\res{e4a.strat.max_width_ratio} times as wide.
\end{itemize}

\section{Related Work}
Prior work on accommodation in language models, dialect bias and post-training.

\subsection{Related Work}
\label{app:related}\label{sec:related}
This section reviews prior work on accommodation in language models, dialect bias and post-training.


\paragraph{Accommodation in language models.}
Accommodation is measured as style matching in word use \citep{niederhoffer2002linguistic,ireland2010language}, as the coordination of style markers \citep{danescu2011mark,danescu2012echoes}, or as baseline-corrected lexical and syntactic alignment \citep{nenkova-etal-2008-high,doyle2016robust,duran2019align}; these measures are sensitive to low-level confounds such as utterance length \citep{xu-etal-2018-much,litterer-etal-2025-coordinating}.
Language models converge strongly when completing human dialogues, and instruction-tuned models converge less \citep{blevins-etal-2026-language}; deployed chat models over-converge to their users' style in the wild \citep{blevins2026accommodation}, although convergence in human--ChatGPT dialogue declines over turns \citep{mitiaeva2026talking}.
LLM replies match their partners' style less than humans do \citep{durandard-etal-2025-language,durandard-etal-2025-llms} and entrain less than therapists \citep{kian-etal-2025-using}; alignment also depends on the interlocutor, falling below human levels with children though not with adults \citep{french-etal-2024-aligning} and better for higher socioeconomic groups \citep{bassignana-etal-2026-large}.
Models also favour tokens seen in the context, even irrelevant ones \citep{niu-etal-2025-llama,kukreja-etal-2026-better}, which a convergence measure must discount as copying.
Unlike these works, which measure style, we measure the features that distinguish \emph{varieties}---spelling, register and dialect grammar---at every released post-training stage, and we separate what the model does from what its update favours.

\paragraph{Homogenization or directed standardization?}
Post-training is associated with reduced output diversity \citep{kirk2024understanding,padmakumar2024does,west2025base,gude-etal-2026-aligned}, which has been traced to typicality bias in preference data \citep{zhang2025verbalized} and located in particular post-training stages \citep{karouzos2026where,li2026narrative}. On the superficial-alignment view, models learn the format of replies from little fine-tuning data \citep{zhou2023lima}, and alignment tuning shifts mostly stylistic tokens \citep{lin2024unlocking}.
Direction matters as much as variance: AI writing suggestions pull Indian writers toward Western styles \citep{agarwal2025ai}, LLM rewriting erases South Asian English markers \citep{bharati-etal-2026-american}, and chat models answer speakers of other varieties in a ``standard'' variety \citep{fleisig-etal-2024-linguistic}.
Homogenization predicts that replies vary less; we test a directed shift instead, in which replies to every informal cue move toward one standard at both fine-tuning stages, in line with the superficial-alignment view, while the update is not variety-neutral.

\paragraph{Reward models, preference data and dialect.}
Reward models disprefer African American Language and steer conversations toward White Mainstream English \citep{mire-etal-2025-rejected}; across the pipeline, dialect gaps accumulate and reward-model preferences between dialect versions of a problem change sign across tasks \citep{elle2026dialect}; preference tuning creates disparities between English dialects \citep{ryan-etal-2024-unintended}; and reward models are weaker outside English \citep{gureja-etal-2025-rewardbench}.
Preference models also reward surface form: length \citep{singhal2024long}, lists, bold text and emojis \citep{zhang-etal-2025-lists}, and human and LLM judges rate fluent answers above factually correct ones \citep{wu-aji-2025-style}; preference-tuned models also defer to users' views \citep{sharma2024towards}.
Unlike these studies, which score fixed texts with a reward model or judge, we read the \emph{policy's own} update on content-matched replies that differ only in whether they match the user's variety.

\paragraph{Likelihood-ratio diagnostics of post-training.}
Direct preference optimization makes the log-ratio between policy and reference an implicit reward \citep{rafailov2023direct}, interpretable at the token level \citep{rafailov2024r}, and reward benchmarks evaluate it alongside explicit reward models \citep{lambert-etal-2025-rewardbench}. As a reward model it generalizes less well than an explicit one \citep{lin-etal-2024-limited}, and preference-tuned models often fail to rank preferred responses above rejected ones \citep{chen2024preference}.
Differences between a tuned and an untuned model's log-probabilities also isolate what fine-tuning adds and can transfer it across scales \citep{mitchell2024emulator,liu2024tuning}.
We use the ratio descriptively, as an \emph{update diagnostic}: comparing it across content-matched replies shows which reply forms a training stage up- or down-weights given the user's variety. We do not treat it as an identified reward, and we read the SFT-versus-base ratio as a likelihood shift only.

\paragraph{Standards, ideology and who is affected.}
Standard-language ideology treats one variety as correct and others as errors \citep{milroy2001language,lippigreen2012english}; sociolinguistics distinguishes monocentric from pluricentric languages, whose national varieties have their own norms \citep{clyne1992pluricentric}, and postcolonial Englishes can develop endonormative standards \citep{schneider2003dynamics}. Recent work argues that generative AI reproduces this ideology \citep{smith2026standard,ugwuanyi2026ai}.
The fairness literature documents its costs: covert dialect prejudice revealed by matched-guise probing \citep{hofmann2024ai} and by paired dialect--standard comparisons \citep{bui-etal-2025-large}, which human-feedback alignment conceals rather than removes \citep{hofmann2024ai} and side-by-side comparison amplifies \citep{kondapally2026side}; dialect-dependent safety behaviour \citep{haq2026dialect}; harms that must be named against a normative standard \citep{blodgett-etal-2020-language}; and adaptation labour shifted onto speakers \citep{cunningham-etal-2024-understanding}.
Speakers want control over when a variety is used: Black American participants prefer Mainstream U.S.\ English in formal settings and are more open to AAE in informal ones \citep{sandoval-etal-2025-llm}, and preferences differ between text and spoken AAE chatbots \citep{finch-etal-2025-finding}.
Our results are consistent with, but do not by themselves establish, a monocentric standard in post-training. They give sociolinguists a test of accommodation theory with machine interlocutors, alignment researchers a diagnostic of what preference data teach beyond content, and auditors a way to separate what a model does from what its training favours.

\FloatBarrier

\end{document}